\documentclass[11pt,a4paper]{article}
\usepackage[utf8]{inputenc}
\usepackage[T1]{fontenc}

\usepackage[english]{babel}
\usepackage[hidelinks]{hyperref}
\addto\extrasenglish{%
}
  
\usepackage[english]{isodate}
\usepackage[paper=a4paper]{geometry}
\newgeometry{top=3.5cm,bottom=2.5cm,right=2.5cm,left=2.5cm}
\usepackage{graphicx}
\usepackage{comment}
\usepackage{fancyhdr}
\usepackage{framed}
\usepackage{lastpage}
\usepackage{tabularx}
\usepackage[table]{xcolor}
\usepackage{enumitem}
\usepackage{mdwlist}
\usepackage{placeins}
\usepackage{amsmath}
\usepackage{xcolor}
\usepackage{listings}
\usepackage{subcaption}
\usepackage{natbib}
\usepackage{wrapfig}

\usepackage[short]{optidef}

\DeclareMathOperator*{\argminB}{argmin}

\usepackage{amsthm}

\usepackage{amssymb}
\newcommand{\R}{\mathbb{R}}
\renewcommand{\r}{\mathbb{R}}
\newcommand{\rr}{$\r$ }
\newcommand{\N}{\mathbb{N}}
\newcommand{\n}{\mathbb{N}}
\usepackage{mathrsfs}
\newcommand{\h}{\mathscr{H}}
\newcommand{\hh}{$\h$ }
\newcommand{\lp}{$\ell^p$ }

\DeclareMathSymbol{\sminus}{\mathbin}{AMSa}{"39}

\usepackage{ifthen}
\newcommand{\ang}[2]{\left\langle \ifthenelse{\equal{#1}{}}{\cdot}{#1},
\ifthenelse{\equal{#2}{}}{\cdot}{#2} \right\rangle}
\newcommand{\angs}[1]{\langle #1, #1 \rangle}
\newcommand{\nystrom}{Nystr{\"o}m }

\newcommand{\todo}[1]{\textcolor{RedText}{\huge\textbf{TODO:}} #1
{\newline\textcolor{RedText}{\rule{\paperwidth}{0.03cm}}} }

\newcommand{\colred}[1]{\textcolor{colr}{\textbf{#1}}}
\newcommand{\colblue}[1]{\textcolor{colb}{\textbf{#1}}}
\newcommand{\colgreen}[1]{\textcolor{colg}{\textbf{#1}}}

\newtheoremstyle{thmstyle}
{\topsep}
{\topsep}
{\itshape}
{}
{\bfseries}
{}
{.5em}
{\thmname{#1}~\thmnumber{#2}\thmnote{ (\bfseries#3)}}
\theoremstyle{thmstyle}
\usepackage{mdframed} 
 \numberwithin{dummy}{section}
\newmdtheoremenv{mydef}[dummy]{Definition}
\newmdtheoremenv{myprop}[dummy]{Proposition}
\newmdtheoremenv{mylemma}[dummy]{Lemma}
\newmdtheoremenv{mytheorem}[dummy]{Theorem}

\numberwithin{equation}{section}

\newcommand{\defref}[1]{Definition~\ref{#1}}
\newcommand{\lemref}[1]{Lemma~\ref{#1}}
\newcommand{\thmref}[1]{Theorem~\ref{#1}}

\definecolor{GreenText}{HTML}{43A047}
\definecolor{RedText}{HTML}{D32F2F}

\definecolor{colorfive}{HTML}{2E7BFF}
\definecolor{colorthree}{HTML}{038314}
\definecolor{colortwo}{HTML}{02AB18}

\definecolor{colr}{HTML}{E53935}
\definecolor{colb}{HTML}{1565C0}
\definecolor{colg}{HTML}{4CAF50}

\definecolor{Bkgr}{HTML}{F1F1F1} 
\definecolor{Text}{HTML}{414141} 
\definecolor{Comment}{HTML}{706161} 
\definecolor{Key}{HTML}{D32F2F} 
\definecolor{Nums}{HTML}{9C27B0} 
\definecolor{Args}{HTML}{8BC34A} 
\definecolor{NewArgs}{HTML}{6EA929} 
\begin{document}


\newcommand{\thtitle}  	 {Efficient Tensor Kernel methods for sparse regression}
\newcommand{\headtitle}  {Efficient Tensor Kernel methods}
\newcommand{\thsubtitle} {}
\newcommand{\name}    	 {Final Thesis}
\newcommand{\versione}{1.0}


\pagenumbering{Alph}
\begin{titlepage}
	\begin{center}
		\includegraphics[trim={0 9.7cm 5.75cm 0},clip,width=0.5\textwidth]{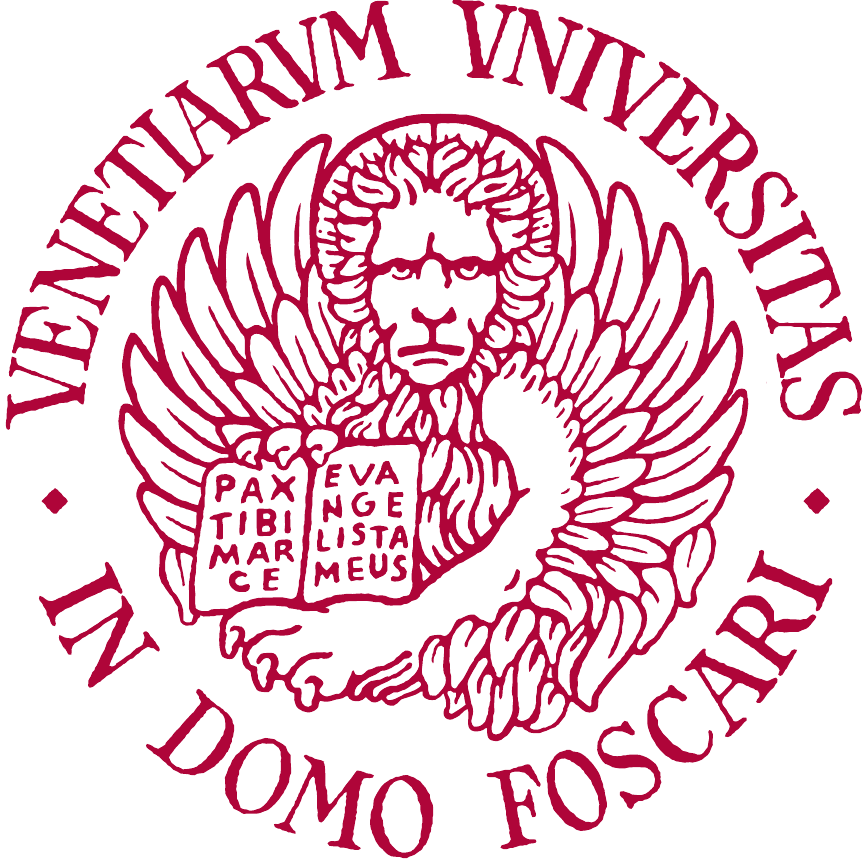}
				
		\vspace{1.5cm}
		\huge
		\textbf{Master's Degree in Computer Science}\\
		\Large {[CM9] Computer Science - D.M. 270/2004}
		
		\vspace{0.8cm}
		\Huge{Final Thesis}
		
    	\vspace{0.8cm}
		\Huge{\textbf{\thtitle}}\\
		\huge{\thsubtitle}
		\vspace{1cm}
		\vfill
	\end{center}
  \begin{raggedright}
    \Large
    \textbf{Supervisors} \\
    \large Ch. Prof. Marcello Pelillo\\
    \large Ch. Prof. Massimiliano Pontil\\
    \vspace{0.5cm}
    \textbf{Assistant Supervisor} \\
    \large Dott. Saverio Salzo\\
    \vspace{0.5cm}
    \Large
    \textbf{Graduand}\\ 
    \large 
    Feliks Hibraj \hfill {\Large \textbf{Academic Year}} \\
    Matriculation Number 854342 \hfill 2019/2020\\
    \vspace{0.5cm}
  \end{raggedright}
\end{titlepage}

\newgeometry{top=2.5cm,bottom=1.5cm,right=3.5cm,left=3.5cm}
\begin{titlepage}
\huge
\begin{center}
\textbf{Abstract}
\end{center}
\vspace{1cm}
\large
Recently, classical kernel methods have been extended by the introduction of suitable tensor kernels so to promote sparsity in the solution of the underlying regression problem. Indeed they solve an \lp -norm regularization problem, with $p=m/(m-1)$ and $m$ even integer, which happens to be close to a lasso problem.
However, a major drawback of the method is that storing tensors generally requires a considerable amount of memory, ultimately limiting its applicability. 
In this work we address this problem by proposing two advances.
First, we directly reduce the memory requirement, by introducing a new and more efficient layout for storing the data. 
Second, we use Nystr\"om-type subsampling approach, which allows for a training phase with a smaller number of data points, so to reduce the computational cost.
Experiments, both on synthetic and real datasets show the effectiveness of the proposed improvements.
Finally, we take care of implementing the code in \textit{C++} so to further speed-up the computation.
\newline
\bigbreak
\noindent
\begin{center}\textbf{Keywords}:\\\medbreak
\textit{Machine Learning, Tensor Kernels, Regularization, Optimization}
\end{center}
\end{titlepage}
\newgeometry{top=3.5cm,bottom=2.5cm,right=2.5cm,left=2.5cm}


\renewcommand{\headheight}{14pt}

\pagestyle{fancy}
\lhead{}
\chead{}
\lhead{\textit{\name}}
\rhead{\textbf{\headtitle}}
\cfoot{}
\renewcommand{\headrulewidth}{0.4pt}
\renewcommand{\footrulewidth}{0.4pt}

\renewcommand{\labelitemi}{$\circ$}
\renewcommand{\labelitemii}{$\bullet$}
\renewcommand{\labelitemiii}{$\diamond$}

\setlist{itemsep=0pt}

\setlength{\parindent}{0cm}


\pagenumbering{gobble}
\renewcommand{\contentsname}{Index}
\tableofcontents
\newpage
\pagenumbering{arabic}


\rfoot{\thepage\ di \pageref{LastPage}}


\newcommand{\defi}[1]{\begin{mydef}{#1}\end{mydef}}
\newcommand{\defin}[2]{\begin{mydef}[#1]{#2}\end{mydef}}
\newcommand{\prop}[1]{\begin{myprop}{#1}\end{myprop}}
\newcommand{\lem}[1]{\begin{mylemma}{#1}\end{mylemma}}
\newcommand{\lemn}[2]{\begin{mylemma}[#1]{#2}\end{mylemma}}
\newcommand{\thm}[1]{\begin{mytheorem}{#1}\end{mytheorem}}
\newcommand{\thmn}[2]{\begin{mytheorem}[#1]{#2}\end{mytheorem}}

\newcommand{\spa}{\vspace*{0.35cm}}

\newcommand{\secone} {Introduction}
\newcommand{\sectwo} {Basic Concepts - Preliminaries}
\newcommand{\secthree} {Classical Kernel Methods}
\newcommand{\secfour} {Extension to Tensor Kernel Methods}
\newcommand{\secfive} {Memory efficient implementation of Tensor Kernels}
\newcommand{\secsix} {\nystrom like strategy for Time Efficiency} 


\section{\secone}
Kernel methods are widely used in different machine learning applications. They are usually formulated as an empirical risk minimization problem, enriched with a regularization term lying in the $\ell^2$ space. 
This is because the inner product structure is necessary to the theory behind common kernel functions (in order to build the corresponding Reproducing Kernel Hilbert Space).
But this restriction does not permit to consider other regularization schemes, such as the sparsity providing one, achievable by making use of the $\ell^1$ norm. Specific kernel functions built on such schemes (with an associated Reproducing Kernel Banach Space) are particularly restrictive and computationally unfeasible. 
However, it was proved that choosing $p\in]1,2[$ arbitrary close to $1$ can be seen as a proxy to the $\ell^1$ case, providing relatively similar sparsity property. Recent findings have proposed tensorial kernel functions to tackle the problem of $\ell^p$ regularization with $p\in]1,2[$ arbitrary close to 1. 
This method is taken into consideration in this thesis.\\

Tensor kernel functions are employed by utilising a tensorial structure storing the values of such defined kernel. Problems arise in storing such structures since they are particularly memory requiring. The purpose of this work is to propose improvements not only to reduce the memory usage by introducing a novel memory layout, but also to improve the overall execution time for solving the optimization problem by making use of Nystr\"om type strategy. \bigbreak

This thesis is organized as follows:\\
In \autoref{preliminaries} we introduce the preliminaries of this field, that is we begin by providing the basics of functional analysis, from the definition of a vector space to that of Banach and Hilbert spaces. Successively introducing kernel functions together with some properties and related relevant theory such as Reproducing Kernel Hilbert Space, Riesz representation theorem and more. \\
In \autoref{sec_classical_kernel_methods} we provide fundamental concepts about classical kernel methods, thus including an introduction to statistical learning theory (discussing loss functions and empirical risk minimization), followed by discussions about regularization together with examples such as Ridge and Lasso regression methods. We conclude the chapter by introducing notable examples of kernel methods, such as kernel ridge regression and support vector machine.\\
Tensor kernel functions are introduced in  \autoref{tensor_section} by analogy to the kernel methods of previous chapter in order to better clarify the theory behind them. Some examples of tensor kernel functions are provided in the end of the chapter.\\
The last two chapters are dedicated to experiments. 
In particular in \autoref{sec_five} we discuss the proposed data layout and the improvement made possible. So we carry out experiments on Memory gain and execution times both on real world and synthetic datasets. 
In \autoref{\secsix} instead, the second improvement is considered, namely the \nystrom like strategy, by means of experiments carried out on large numbers otherwise unfeasible. 
Also in this case, both real world and synthetic datasets are utilised, with an emphasis on analysing the feature selection capability of the algorithm.
\newpage
\newpage
\section{\sectwo}\label{preliminaries}
In this chapter we are first going to see (in \autoref{functional_analysis}) some of the basic concepts regarding vector spaces and how we move into Banach and Hilbert spaces, along with the connection of the latter to $\ell^2$ spaces. Next comes the presentation of kernel functions, some properties and reproducing kernel hilbert spaces (in \autoref{section_rkhs}).
\subsection{Functional Analysis}\label{functional_analysis}

\defin{Vector Space}{A vectors space over $\r$ is a set $V$ endowed with two binary operations $+\colon V\times V\to V$ and 
$\cdot\colon \mathbb{R}\times V\to V$ such that
\begin{itemize}
\item Associative law: $(u+ v) + w = u+ (v+ w)\quad 
\forall u,v,w\in V$
\item Commutative law: $u+ v = v+ u\quad \forall u,v\in V$
\item Identity vector: $\exists 0_V\in V\ s.t.\quad \forall u\in V,\ u+0_V = u$
\item Existence of the inverse: $\forall u\in V,\ \exists (-u) \ s.t.\quad  u+(- u) = 0_V$
\item Identity laws: $1 \cdot u = u, \forall\, u \in V$
\item Distributive laws: \\
\hspace*{1cm}$a\cdot (b\cdot u) = (ab)\cdot u$\\
\hspace*{1cm}$(a+ b)\cdot u = a\cdot u+ b \cdot u$
\end{itemize}
}

Examples of Vector spaces are:
\begin{itemize}
\item $\R$
\item $\R^n$
\item $\R^\R$
\item $\R^X$
\item continuous functions from a metric space $x\rightarrow \mathbb{R}$
\end{itemize}

\defin{Norm}{A norm is a function on a vector space $V$ over $\R$, $\|\cdot \|:V\rightarrow \R$, with the following properties:\\
$\forall u,v\in V, \forall a\in \R$:
\begin{itemize}
\item Non-negative: $\|u\|\geq 0$
\item Strictly-positive: $\|u\| = 0 \Rightarrow u=0$
\item Homogeneous: $\|au\| = |a|\|u\|$
Triangle inequality: $\|u+v\| \leq \|u\| + \|v\|$
\end{itemize}
}

A vector space endowed with a norm is called a \textbf{normed vector space}.
\newpage

\defi{Let $(V, \|\cdot\|)$ be a normed vector space. A \textbf{Cauchy Sequence} is a sequence $(u_n)_{n \in \mathbb{N}}$ such that:
$\forall \epsilon>0,\ \exists N \in \mathbb{N}\ s.t.\ \|u_n-u_m\|<\epsilon,\ \forall n,m > N$.\\ 
A sequence $(u_n)_{n \in \mathbb{N}}$ is said to be \textbf{convergent} in $V$ if there is a point $u\in V$ such that 
$\forall \epsilon>0,\ \exists N\in\mathbb{N}\ s.t.\ 
\|u - u_n\|<\epsilon,\ \forall n>N$. In that case one writes
$\lim_{n \to +\infty} u_n = u$.
}

\defi{A \textbf{complete vector space} $V$ is a vector space equipped with a norm and complete with respect to it, i.e., for every Cauchy sequence $(u_n)_{n \in \mathbb{N}}$ in $V$ there exists an element $u\in V$ such that $\lim_{n \to +\infty} u_n = u$.
}

\defi{A \textbf{Banach Space} is a complete normed vector space $(V, \|\cdot\|)$
}

\defi{A function $T:V\rightarrow W$, with $V$ and $W$ vector spaces over \rr, is a \textbf{bounded linear operator} if:
\begin{displaymath}
T(\alpha_1 u_1 + \alpha_2 u_2) = \alpha_1T(u_1) + \alpha_2T(u_2),\quad 
\forall \alpha_1,\alpha_2 \in \r,\ \forall u_1,u_2\in V
\end{displaymath}
and $\ \exists c > 0\ s.t.$
\begin{equation*}
\|T u\|_W \le c\|u\|_V, \quad \forall u\in V.
\end{equation*}
In such case the \emph{norm of $T$} is
\begin{equation}
    \| T \| = \sup_{\| u \|_V \leq 1} \| T u \|_W.
\end{equation}
Bounded linear operators which are defined from $V$ to \rr are called bounded functionals on $V$. 
The space of the bounded functionals on $V$ is called the dual space of $V$ and denoted by $V^*$, that is,
\begin{equation}
    V^* = \{ \varphi\colon V \to \mathbb{R} : \varphi \text{ is bounded linear functional } \}.
\end{equation}
endowed with the norm $\|\varphi \|= \sup_{\| u \| \leq 1} |\varphi(u)|$.
Finally, the canonical \emph{pairing} between $V$ and $V^*$
is the mapping
\begin{equation}
    \langle \cdot, \cdot\rangle \colon V \times V^* \to \mathbb{R},
    \qquad \langle u, \varphi\rangle = \varphi(u).
\end{equation}
}

\defin{$\ell^p$ space}{\label{lp_space}For $0<p<\infty$, the space of sequences $\ell^p$ is defined as
\begin{equation}
 \ell^p = \left\lbrace \left\lbrace x_i\right\rbrace_{i=0}^\infty\ :\ \sum^\infty_{i=0} |x_i|^p < \infty\right\rbrace
 \end{equation} 
}

\defin{Norm on $\ell^p$}{Given an $\ell^p$ space, we define the norm on $\ell^p$ by
\begin{equation}
\left\| \{x_i\}^\infty_{i=0} \right\|_p = 
\left( \sum^\infty_{i=0} |x_i|^p \right)^{1/p}
\end{equation}
}

\defin{Inner product}{\label{innerproduct} An Inner/Dot/Scalar product on a vector space $H$ over \rr is a map $\langle\cdot,\cdot\rangle : H\times H \rightarrow \r$, satisfying the following:\\
$\forall u,v,w \in H, a\in\mathbb{R}$:
\begin{itemize}
\item Symmetry: $\ang{v}{w} = \ang{w}{v}$
\item Linearity w.r.t. first term: $\ang{u+w}{v} = \ang{u}{v}+\ang{w}{v}$
\item Linearity w.r.t. second term: $\ang{u}{v+w} = \ang{u}{v} + \ang{u}{w}$
\item Associative: $\ang{au}{v} = a\ang{u}{v}$
\item Positive Definite: $\ang{v}{v} > 0\quad \forall v\neq 0$
\end{itemize}
The norm associated to the scalar product $\langle \cdot, \cdot \rangle$ is defined as follows:
$\| u \| = \langle u, u\rangle^{1/2}$.
}

\defi{A \textbf{pre-Hilbert Space} $H$ is a vector space endowed with a scalar product. If the norm associated to the scalar product defines a complete normed space, then $H$ is called a Hilbert space.}

\defin{$\ell^2$ space}{\label{l2_space}The space $\ell^2$ is defined as
\begin{equation}
 \ell^2 = \left\lbrace \left\lbrace x_n\right\rbrace_{n=0}^\infty\ :\ \sum^\infty_{n=0} |x_n|^2 < \infty\right\rbrace
 \end{equation} 
 endowed with the scalar product
 \begin{equation}
     \langle x,y \rangle = \sum_{n=0}^{+\infty} x_n y_n.
 \end{equation}
 This space is a Hilbert space.
}

\prop{
Let $H$ be a (separable) Hilbert space. Then an \emph{orthonormal basis} of $H$ is a sequence $(a_n)_{n \in \mathbb{N}}$ in $H$ such that, $\mathrm{span}\{ a_n : n \in \mathbb{N}\} = H$ and
$\langle a_n, a_m \rangle = \delta_{n,m}$, for every $n,m \in \mathbb{N}$. In such case, for every $u \in H$,
we have
\begin{equation}
\sum_{n = 0}^{+\infty} |\langle u,a_n \rangle|^2<+\infty
\quad\text{and}\quad
u = \sum_{n=0}^{+\infty} \langle u,a_n \rangle a_n.
\end{equation}
Moreover, fore very $u,v \in H$, $\langle u,v \rangle = \sum_{n=0}^{+\infty} \langle u,a_n \rangle\langle v, a_n \rangle$
and 
$\|u\|^2 = \sum_{n=0}^{+\infty} |\langle u,a_n \rangle|^2$.
This establishes an isomorphism between $H$ and $\ell^2$.
}

\newpage
\subsection{Reproducing Kernel Hilbert Spaces}\label{section_rkhs}
In this section we are going to introduce a fundamental aspect of this work, that is kernel functions and Reproducing Kernel Hilbert Spaces. These concepts are the starting point upon which, the definition of Tensor Kernels arises.\\
Definitions are kept in a general form in order to give a wider point of view of the field. 
Main properties are reported. More in depth discussions can be retrieved in \citep{steinwart2008}.
\spa

\defin{Kernel}{\label{kernel} Let $\mathcal{X}$ be a non-empty set, a \textbf{Kernel} is a function $k:\mathcal{X}\times \mathcal{X} \rightarrow \R$ and there exists a Hilbert Space $H$ and a mapping $\Phi: \mathcal{X} \rightarrow H$ s.t. for all $x,x' \in \mathcal{X}:$\\
\begin{equation}
k(x,x') = \langle \Phi(x),\Phi(x')\rangle
\end{equation}\spa
The mapping $\Phi$ is called a \textbf{Feature Map} and $H$ a \textbf{Feature Space} of function $k$.
}\spa
There are no conditions on $\mathcal{X}$ other than being a non-empty set, that is, it can be a set of discrete objects such as documents, strings, nodes of a graph or an entire graph. That means that we do not require an inner product to be defined for elements of $\mathcal{X}$.
\spa

Suppose that $H$ is a separable Hilbert space. Then there exists
an isomorphism $T\colon H \to \ell^2$, meaning that
$T$ is a bounded linear and bijective operator and
$\langle u, v \rangle_{H} = \langle T u, T v \rangle_{\ell^2}$.
Therefore, 
\begin{equation}
k(x,x') = \langle \Phi(x),\Phi(x')\rangle_H 
= \langle T\Phi(x),T\Phi(x')\rangle_{\ell^2}.
\end{equation}
This shows that one can always choose a feature map with values in $\ell^2$.

To build a kernel function from scratch, as the successive theorem proves, we first need the definition of positive definiteness and symmetry.

\spa
\defin{Positive definite}{A function $k:\mathcal{X}\times \mathcal{X} \rightarrow \R$ is said to be \textbf{positive definite} if, for all $n\in \N$, $\alpha_1, ...,\alpha_n \in \R$ and all $x_1, ...,x_n\in \mathcal{X}$
\begin{equation}
\sum_{i=1}^n\sum_{j=1}^n \alpha_i\alpha_jk(x_i,x_j) \geq 0
\end{equation}
}\spa

Let $K=(k(x_i,x_j))_{i,j}$ be a $n\times n$ matrix for $x_1,...,x_n\in \mathcal{X}$. Such matrix is referred to as the \textbf{Gram matrix} of $k$ with respect to $x_1,...x_n$. Similarly to the above definition, we say that matrix $K$ is positive semi-definite if:
\begin{displaymath}
\alpha^\top K\alpha = \sum_{i=1}^n\sum_{j=1}^n \alpha_i\alpha_jK_{ij} \ge 0
\end{displaymath}
holds for all $\alpha_1,...\alpha_n$. This definition is also referred to as the \textit{energy-based} definition, and it doesn't seem trivial how to assure such property. A more immediate definition of positive definite is a matrix having all positive eigenvalues. There are other tests one can carry out to prove positive definiteness for a generic matrix $K$:
\begin{itemize}
\item Check that all eigenvalues $\lambda$ associated to $K$ are positive.
\item Check that all pivots of matrix $K$ are positive.
\item Check that all upper-left matrix determinants are positive.
\item Have a decomposition of $K=A^TA$, with $A$ rectangular having all independent columns.
\end{itemize}

\medbreak
A function $k$ is called \textbf{symmetric} if $k(x,x') = k(x',x)$ for all $x,x'\in \mathcal{X}$.\bigbreak

Kernel functions are symmetric and positive definite.\\
Let $k$ be a kernel function with $\Phi:\mathcal{X}\rightarrow H$ the associated feature map. Since the inner product in \hh is symmetric, then $k$ is symmetric.\\
Furthermore, for $n\in \n,\quad \alpha_i\in\r, \quad x_i\in X, \quad i=1,...n$:
$$
\sum^n_{i=1}\sum^m_{j=1}\alpha_i\alpha_jk(x_i,x_j) = \ang{\sum^n_{i=1}\alpha_i\Phi(x_i)}{\sum^m_{j=1}\alpha_j\Phi(x_j)}_H \geq 0
$$
which shows that $k$ is also positive definite.\bigbreak
The following theorem proves that being symmetric and positive definite are necessary and sufficient conditions for a function $k$ to be a kernel.\spa

\thmn{Symmetric, positive definite functions are kernels}{$\quad$\\ A function $k:\mathcal{X}\times \mathcal{X} \rightarrow \r$ is a kernel function if and only if it satisfies the properties of being symmetric and positive definite.
}

\proof{}To prove the theorem we will first consider a pre-Hilbert space of functions, from which we derive a proper inner product that satisfies the aforementioned properties. Successively by defining a proper feature map to a proper feature space (a Hilbert space), we arrive at the definition of a kernel function as presented in \defref{kernel}. \\
Consider
$$
H_{pre} := \left\lbrace\sum^n_{i=1} \alpha_ik(\cdot,x_i)\ |\ n\in\n,\ \alpha_i\in\r,\ x_i\in X,\ i=1,...n\right\rbrace
$$
and taking two elements:
\begin{align*}
f := \sum^n_{i=1} \alpha_ik(\cdot, x_i) \in H_{pre}\\
g := \sum^m_{j=1} \beta_jk(\cdot, x'_j) \in H_{pre}
\end{align*}
we define the following:
$$
\ang{f}{g}_{H_{pre}} := \sum^n_{i=1}\sum^m_{j=1} \alpha_i\beta_jk(x_i, x_j')
$$
We note that it is bilinear, symmetric and we can write independently from the representation of $f$ or $g$:
\begin{align*}
\ang{f}{g}_H = \sum^m_{j=1} \beta_jf(x'_j)\\
\ang{f}{g}_H = \sum^n_{i=1} \alpha_ig(x_i)
\end{align*}
Since $k$ is positive definite, $\ang{}{}_H$ is also positive, that is $\ang{f}{f}\geq 0$ for all $f \in H_{pre}$. Moreover, it satisfies the Cauchy-Schwarz inequality:
\begin{displaymath}
|\ang{f}{g}|^2 \leq \angs{f}_H \cdot \angs{g}_H \quad f,g \in H_{pre}.
\end{displaymath}
That is important in order to prove that $\ang{}{}$ is an inner product for $H_{pre}$, in particular, to prove the last property of \defref{innerproduct} (Inner Product) we write:
\begin{displaymath}
|f(x)|^2 = |\sum^n_{i=1}\alpha_ik(x,x_i)|^2 = |\ang{f}{k(\cdot,x)}_H|^2
\leq \ang{k(\cdot,x)}{k(\cdot,x)}_H \cdot \angs{f}_H = 0
\end{displaymath}
hence we find $f=0$. So $\ang{}{}$ is an inner product for $H_{pre}$.\medbreak
Let $H$ be a completion of $H_{pre}$ and $I:H_{pre} \rightarrow H$ be the isometric embedding. Then $H$ is a Hilbert space and we have
\begin{equation*}
\Phi(x) = I k(\cdot, x)
\end{equation*}
and
\begin{equation*}
\ang{Ik(\cdot,x)}{Ik(\cdot,x')}_H = \ang{k(\cdot,x)}{k(\cdot,x')}_{H_{pre}} 
= k(x,x') \quad \forall x,x'\in X
\end{equation*}
that is the definition of kernel and $x\mapsto Ik(\cdot,x)$ is the feature map of $k$.\qed

Now we are going to introduce the concept of a Reproducing Kernel Hilbert Space (RKHS), followed by some interesting results.\\
To get to the definition of RKHS, we first have a look at the definition of an evaluation functional.

\spa
\defi{Let \hh be a Hilbert Space of functions from $\mathcal{X}$ to \rr. A \textbf{Dirac evaluation functional} at $x\in \mathcal{X}$ is a functional 
\begin{displaymath}
\delta_x:\h \rightarrow \r\quad s.t.\quad \delta_x(f) = f(x)\quad \forall f\in\h
\end{displaymath}
}
This functional simply evaluates the function $f$ at the point $x$.
\spa

\defi{\label{rkhs} A \textbf{Reproducing Kernel Hilbert Space} (RKHS) is a Hilbert Space \hh of function where all the Dirac evaluation functionals in \hh are bounded and continuous.
}

\spa
Being continuous means that:
\begin{equation}
\forall f\in \h\quad \|\delta_x(f)\| \le c_x\|f\|_\h,
\quad for\ some\ c_x > 0
\end{equation}

In other words, this statement points out that norm convergence implies pointwise convergence.\\

\defref{rkhs} is compact and makes use of evaluation functionals. There are some properties residing behind such definition which are fundamental. In particular we are going to see what a Reproducing Kernel is, which, as the name suggests, is the building block of RKHSs. Indeed, an alternative definition based on Reproducing Kernels can be retrieved.

\spa
\defin{Reproducing Kernel}{\label{reproducing_property} Let \hh be a Hilbert Space of functions from $\mathcal{X}$ to \rr, with $\mathcal{X}\neq0$. A function $k:\mathcal{X}\times \mathcal{X} \rightarrow \r$ is called a \textbf{Reproducing Kernel} of \hh if the following holds:
\begin{itemize}
\item $k_x = k(x,\cdot)\in\h,\quad \forall x\in \mathcal{X}$
\item $f(x) = \ang{f}{k_x}_\h,\quad \forall f\in\h,\ x\in \mathcal{X}\qquad$ \textbf{(Reproducing Property)}
\end{itemize}
}

The following Lemma says that an RKHS is defined by a Hilbert space that has a reproducing kernel. 

\spa
\lemn{Reproducing kernels are kernels}{\label{feature_map}Let \hh be a Hilbert space of functions over $\mathcal{X}$ that has a reproducing kernel k. Then \hh is a RKHS and \hh is also a feature space of k, where the feature map $\Phi:\mathcal{X}\rightarrow \h$ is given by
$$
\Phi(x) = k(\cdot, x),\quad x\in \mathcal{X}.
$$
We call $\Phi$ the \textbf{canonical feature map}.
}

\proof{} The reproducing property builds a link between reproducing kernels and Dirac functionals, in particular:
\begin{equation}
|\delta_x(f)| = |f(x)| = |\ang{f}{k_x}| \leq \|k(x,\cdot)\|_\h\|f\|_\h
\end{equation}
for all $x\in \mathcal{X}, f\in\h$ shows the continuity and boundedness of functionals $\delta_x$.\\
To prove that reproducing kernels are kernels, we consider $f:=k(x',\cdot)$ for a fixed $x'\in \mathcal{X}$ and by taking advantage of the reproducing property again, we write:
\begin{equation}
\ang{\Phi(x')}{\Phi(x)} = \ang{k(\cdot, x')}{k(\cdot, x)} = \ang{f}{k(\cdot,x)}
= f(x) = k(x,x')
\end{equation}
for all $x\in \mathcal{X}$. Which is the definition of a kernel on \hh.
\qed

\spa

We saw the alternative definition of a RKHS, which is based on the existence of a reproducing kernel associated to the Hilbert space under consideration.\\
So we can now explore the implications of it, and in particular, the next theorem we are going to see, states that every RKHS has a unique Reproducing kernel.\\
The proof of that theorem makes use of Riesz representation theorem:

\spa
\thmn{Riesz representation theorem}{ If $\varphi$ is a bounded linear functional on a Hilbert Space \hh\!\!\!, then there is a unique $u\in\h$ such that
\begin{displaymath}
\varphi(f) = \ang{f}{u}_\h \quad \forall f\in\h
\end{displaymath}
}
\spa

In an RKHS, evaluation can be represented as an inner product.

\thmn{Every RKHS has a unique reproducing kernel}{$\quad$\\
If \hh is a RKHS, it has a unique reproducing kernel.
}

\proof{} Assume that $\delta_x\in \h$ is a bounded linear functional. By the Riesz representation theorem we know that there exists an element $u\in\h$ s.t.:
\begin{equation}
\delta_x(f) = \ang{f}{u}_\h, \quad \forall f\in\h
\end{equation}
We define $k(x,x') = u(x'),\ \forall x,x'\in \mathcal{X}$. Now we can write that:
\begin{equation}
k_x = k(\cdot, x) = u \in \h
\end{equation}
which is the first property of a reproducing kernel. For the reproducing property, it is sufficient to note that
\begin{equation}
\ang{f}{k_x}_\h = \delta_x(f) = f(x) \quad \forall f\in\h, x\in \mathcal{X}.
\end{equation}
So we proved that \hh has a Reproducing kernel. To prove uniqueness, assume \hh has two reproducing kernels $k_1$ and $k_2$. So we have
\begin{equation}
\ang{f}{k_1(\cdot, x) - k_2(\cdot, x)}_\h = f(x) - f(x) = 0,\quad \forall f\in\h, x\in \mathcal{X}
\end{equation}
if we take $f = k_1(\cdot, x) - k_2(\cdot, x)$, we obtain:
\begin{equation}
\|k_1(\cdot, x) - k_2(\cdot, x)\|^2_\h = 0, \quad \forall x\in \mathcal{X}
\end{equation}
which is equivalent to saying that $k_1 = k_2$, thus proving the uniqueness.
\qed

\spa
The theory of RKHS has been widely studied, and further discussions can be retrieved from \citep{steinwart2008,berlinet2011} and many others.\\

There exists a more general idea, that is that of a Reproducing Kernel Banach Space (RKBS), which does not rely on an inner product to define a norm. The idea is recent, and less work can be found on it. Worth mentioning is \citep{zhang2009}. The concept of Tensor Kernel arises in non-Hilbertian spaces, since it relies on $\ell^p$-norm, with $p\in]1,2[$.\\


\newpage
\newpage
\section{\secthree}\label{sec_classical_kernel_methods}
In this section we begin by providing the basis and then discuss classical kernel methods.\\
So we are first going to introduce the field of Statistical Learning in \autoref{statistical_learning}, discuss some notable loss functions, take into consideration the problem of overfitting, make a distinction between classification and regression.\\
Secondly, we are going to clarify the role of regularization and create a link to the Representer Theorem, in order to introduce kernel functions practically. We are going to conclude by giving two examples of kernel methods, Kernel Ridge Regression (in \autoref{kernel_ridge}) since it is the starting point for the introduction of Tensor Kernel Methods (topic for the next section) and Support Vector Machine in \autoref{svm}, since it is the classical kernel method par excellence.

\subsection{Statistical Learning Theory}\label{statistical_learning}
In the field of Statistical Learning we are given a set of input data (training set)
\begin{equation}
(x_1,y_1), ..., (x_n,y_n)
\end{equation}
where $x_i$ are the input vectors belonging to the input space  $\mathcal{X}$, also called space of instances, while $y_i$ are responses in the output space $\mathcal{Y}$, label space. Each input $x_i$ has a corresponding known output $y_i$.\\
The learning algorithm is fed with the training set, and we would like it to discover the mapping 
\begin{equation}
f: \mathcal{X} \rightarrow \mathcal{Y}
\end{equation}
which is a functional relationship between input and output spaces. This mapping is called a \textit{classifier/regressor}. We would like it to provide as few errors as possible. Since responses are known, the learning algorithm can evaluate its predictions during the learning process. So we are talking about  supervised learning. \bigbreak

In case of binary responses $y_i := \{-1, +1\}$, each object belongs to one of the two classes. We refer to this problem as a \textit{classification} problem, since the goal is to classify a pattern $x_i$ as one of the two known classes. We make no assumption on $\mathcal{X}$ and $\mathcal{Y}$ spaces. The only assumption is that there exists a joint probability distribution $\rho$ on $\mathcal{X}\times \mathcal{Y}$ and $(x_i, y_i)$ pairs are independently sampled from it. The latter is considered a rather strong assumption, but it is usually justified in most of applications in machine learning. Still, there are cases in which we don't have such assumption.\bigbreak

There is no assumption on $\rho$, which means that it can be any distribution on $\mathcal{X}\times\mathcal{Y}$. This way, Statistical Learning Theory provides a general setting. \\
We can have a non-deterministic behaviour of responses $y_i$, in the sense that they are not necessarily determined by the corresponding $x_i$ input.
This is the case for noisy settings, in which the generation of labels is subject to some noise, making it possible for cases where the given response is actually wrong. \\
Another case is that of overlapping classes. Take as an example the task of predicting the gender of a person, given its characteristics. If we are given the characteristic $x$ of having blonde hairs, it is clear that both gender responses are valid, so we cannot assign a unique label $y$.
\bigbreak

In Statistical Learning Theory, the distribution $\rho$ is fixed, so that it doesn't change over time and there is no assumption on the ordering of the training set. \\
The last point to mention about this field is that the distribution is unknown, so the goal is that of approximating it by making use of the training set. It is clear that a larger dataset allows for a better estimation of the underlying distribution $\rho$.\bigbreak 

\subsubsection{Loss Functions}\label{loss_functions}
As we were saying earlier, we would like the classifier $f$ to make as few errors as possible. But, how do we measure its error? We have to introduce a measure of "goodness" for a classifier. For this reason, we introduce a \textit{loss function} $\ell$ defined as
\begin{equation}
\ell:\mathcal{Y}\times \r \rightarrow \r_+
\end{equation}
which takes in input responses $y_i$ and the prediction of the classifier $f(x_i) = \hat{y_i}$, and returns the cost for that classification. In other words, it computes the error made by predicting $y_i$ as $\hat{y_i}$.\bigbreak

\begin{figure}[!ht]
\centering
\subcaptionbox{0-1 loss}{\includegraphics[width=0.3\textwidth]{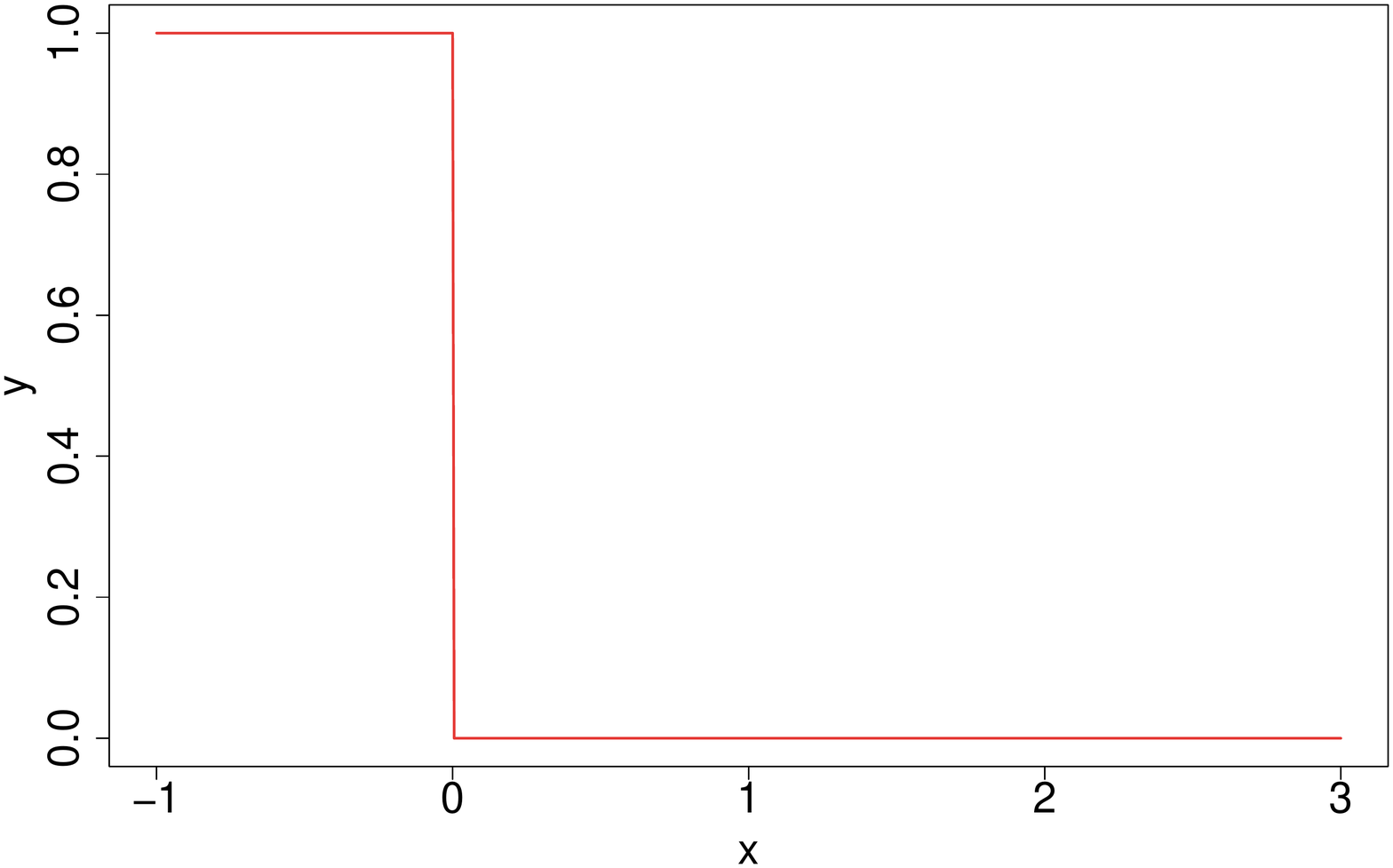}}
\hfill
\subcaptionbox{Hinge loss}{\includegraphics[width=0.3\textwidth]{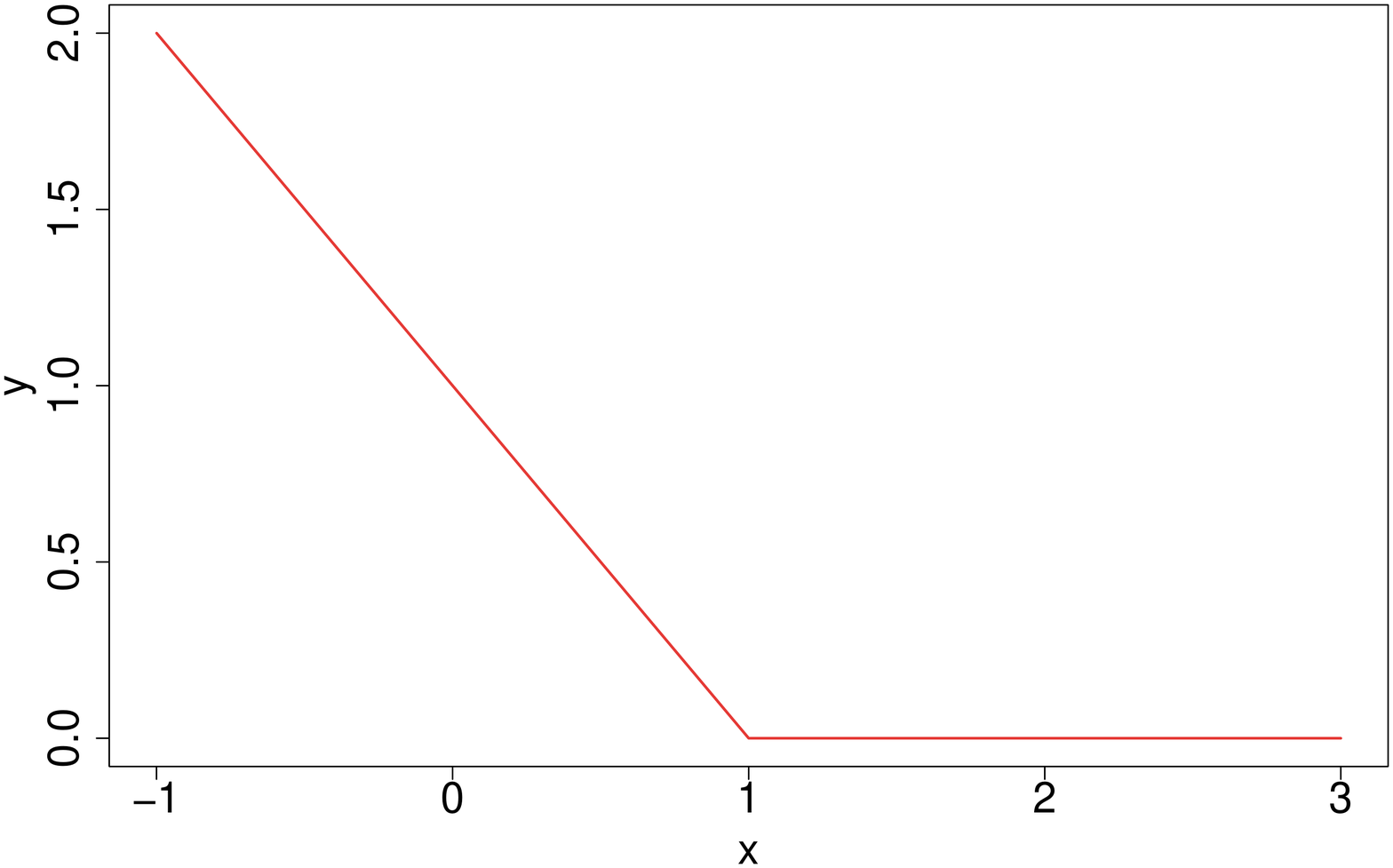}}
\hfill
\subcaptionbox{Logistic loss}{\includegraphics[width=0.3\textwidth]{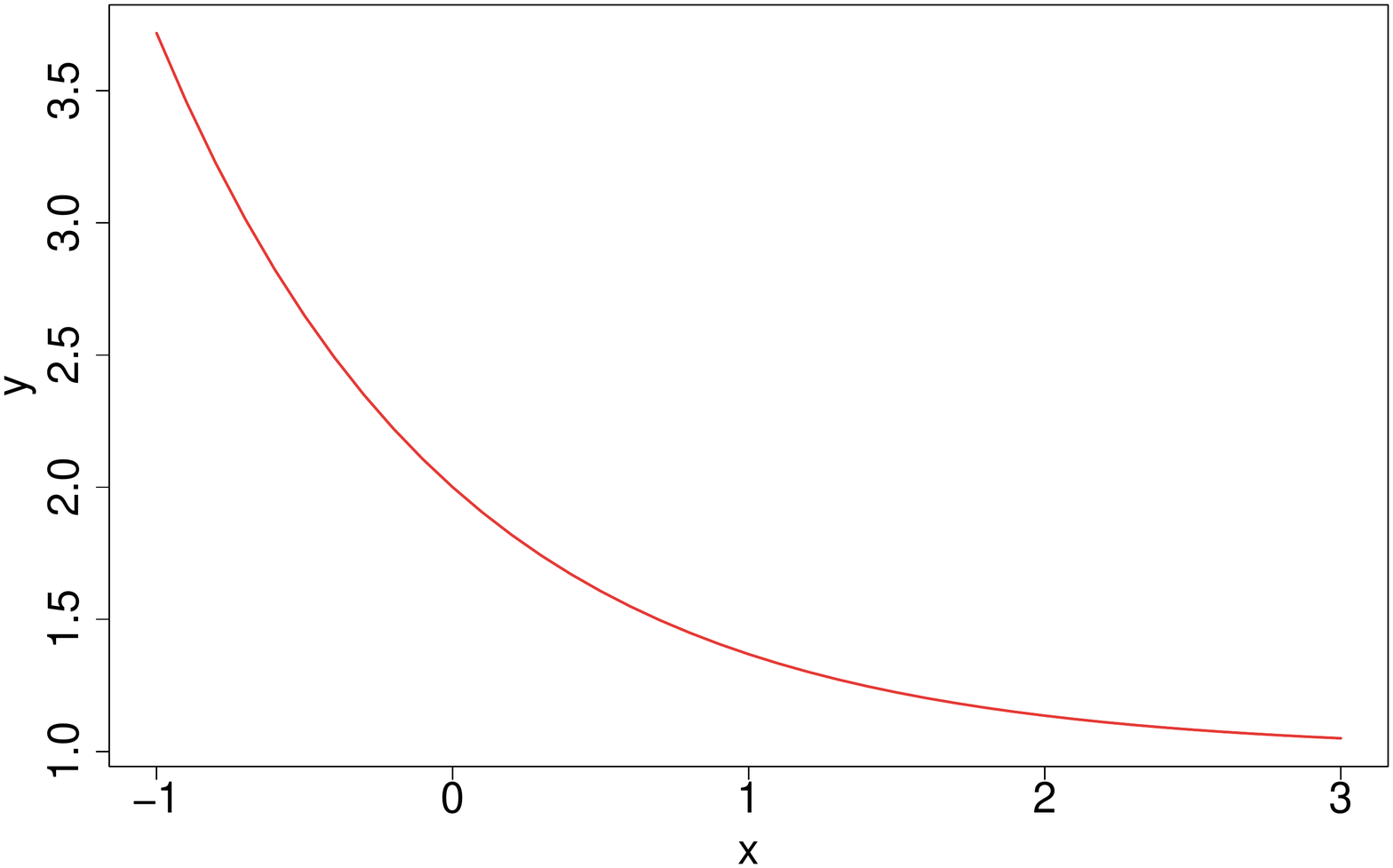}}
\caption{Loss functions for classification problem.}
\label{loss-classification}
\end{figure}

There are plenty of loss functions, in the following we are going to introduce some of them.
The simplest one, is the \textit{0-1 loss} (missclassification error), used for classification problems, defined as
\begin{equation}
\ell_{0-1} (y, f(x)) =
\left\{\begin{array}{ll}
1 & if\ f(x)\neq y \\
0 & otherwise
\end{array}
\right.
\end{equation}
since we have only 2 possible outcomes in classification problems, it is straightforward to think of a binary-type cost for each prediction. That is, if the classifier prediction $f(x)$ is equal to the true value $y$, than there is no cost to pay (value $0$), otherwise, the classifier has made a wrong prediction, thus assign a penalty equal to $1$. This loss function is simple and effective. But, as you can see from \autoref{loss-classification}(a), it has the drawback of not being convex, which is a really valuable property in many applications, along with differentiability.\bigbreak

\begin{figure}[!ht]
\centering
\includegraphics[width=0.5\textwidth]{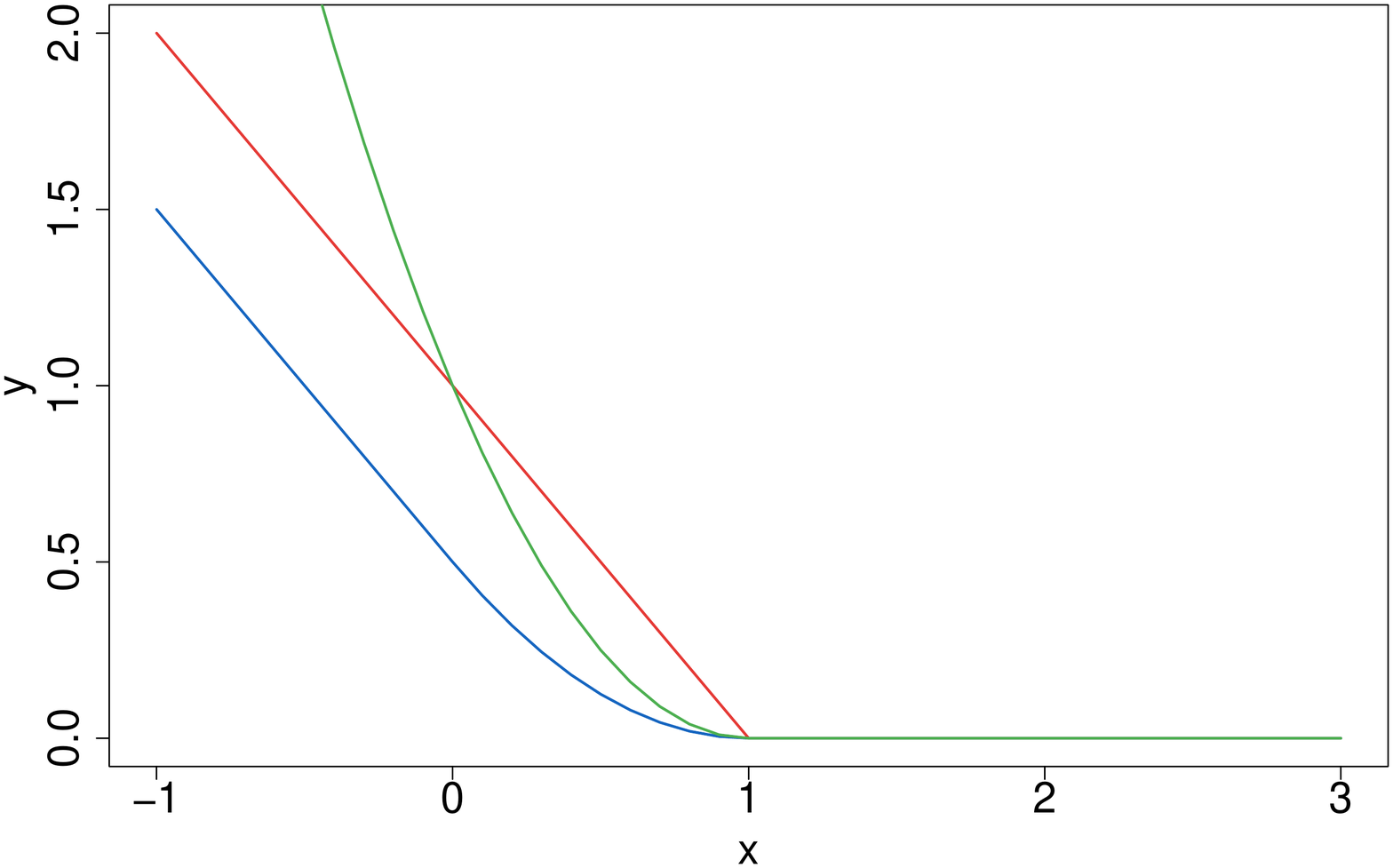}
\caption{Hinge Loss (\colred{red}), Smoothed Hinge Loss (\colblue{blue}), Quadratically smoothed (\colgreen{green}).}
\label{smooth_hinge}
\end{figure}

The \textit{Hinge loss} is a convex (though not differentiable) loss function which is defined as
\begin{equation}
\ell_{hinge}(y, f(x)) = \max\{0,\ 1-yf(x)\}
\end{equation}
Recalling that $y:=\{-1, +1\}$, we note that $yf(x)=1$ for correct estimates, so there is no penalty, while in the case of incorrect predictions, a positive value is returned, accounting for the proper penalty.\\
As you can notice from \autoref{loss-classification}(b), Hinge Loss is not differentiable because of the discontinuity of the derivative at $x=1$. Smoothed versions have been introduced to overcome this problem, for instance in \citep{rennie2005smooth} they define the \textit{smoothed Hinge Loss} 
\begin{equation}
\ell_{smooth-hinge} (z) =
\begin{cases} \tfrac{1}{2}-z & z \le 0,\\\\
\tfrac{1}{2}(1-z)^2 & 0 < z < 1\\\\
0 & z \ge 1
\end{cases}
\end{equation}
where $z$ has been substituted to $yf(x)$ for simplicity. Smoothed Hinge Loss is smooth and its derivative is continuous. There is also a quadratically smoothed one suggested by \citep{zhang2004solving}
\begin{equation}
\ell_{q-smooth} (z) = 
\begin{cases} 
\frac{1}{2\gamma} \max(0, 1-z)^2 & if\ z \ge 1-\gamma, \\
1 - \frac{\gamma}{2} - z 		& otherwise
\end{cases}
\end{equation}
where, again, $z=yf(x)$. Note that for $\gamma \rightarrow 0$, it becomes the Hinge Loss. Graphically, you can see the behaviour of such functions in \autoref{smooth_hinge}.

\bigbreak
Other standard loss functions have been introduced to overcome the lack of differentiability, \textit{Logistic loss} is an example
\begin{equation}
\ell_{log}(y, f(x)) = 1 + e^{-yf(x)},
\end{equation}
which is both convex and differentiable as you can see from \autoref{loss-classification}(c).\bigbreak

A distinction must be made, namely, problems in which the response variable $Y$ can take any real value, are referred to as \textit{Regression Problems}. So we have to consider loss functions that accounts for the amount by which the algorithm mistakes.

\begin{figure}[!ht]
\centering
\subcaptionbox{$L_2$ loss}{\includegraphics[width=0.3\textwidth]{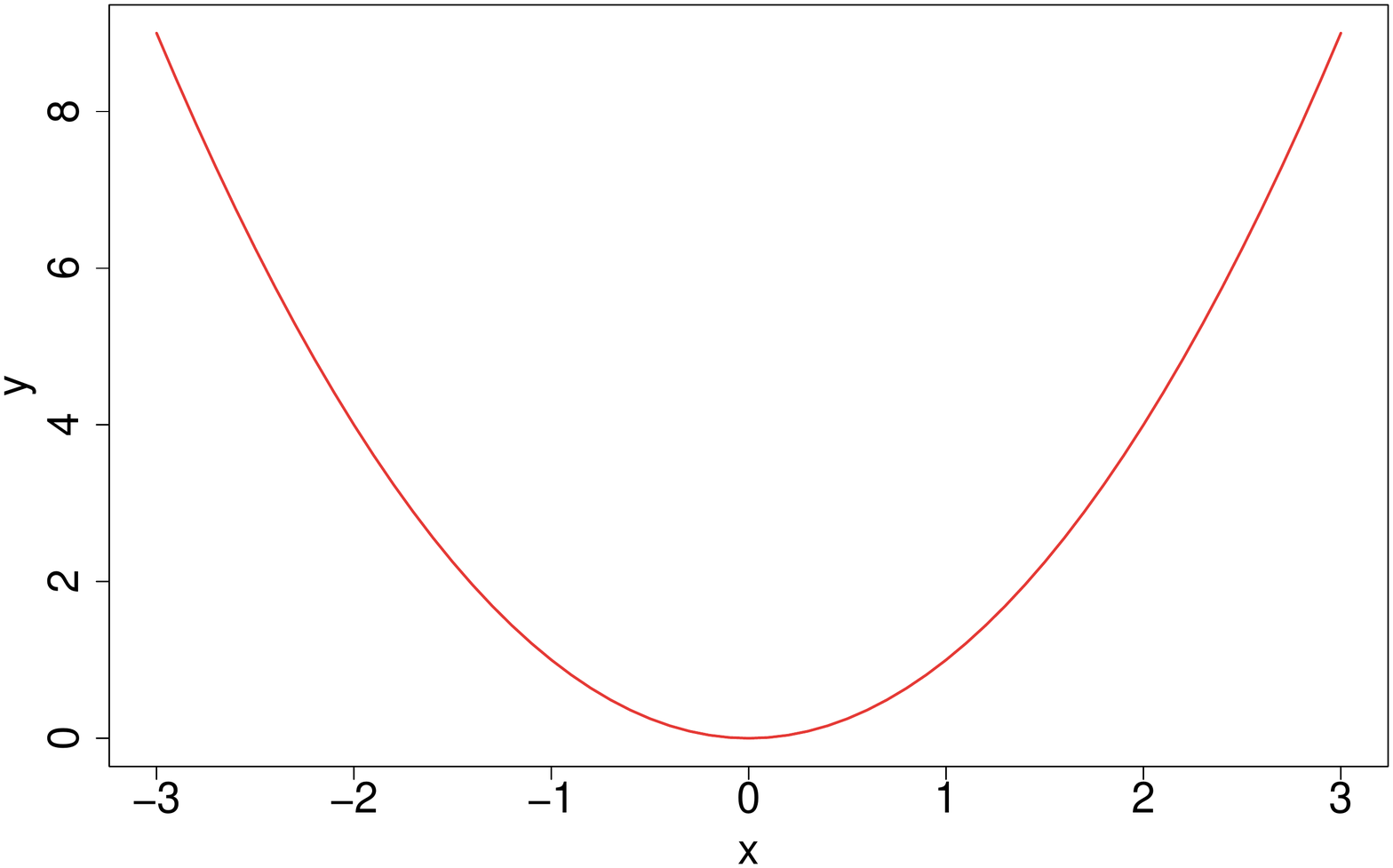}}
\hfill
\subcaptionbox{$L_1$ loss}{\includegraphics[width=0.3\textwidth]{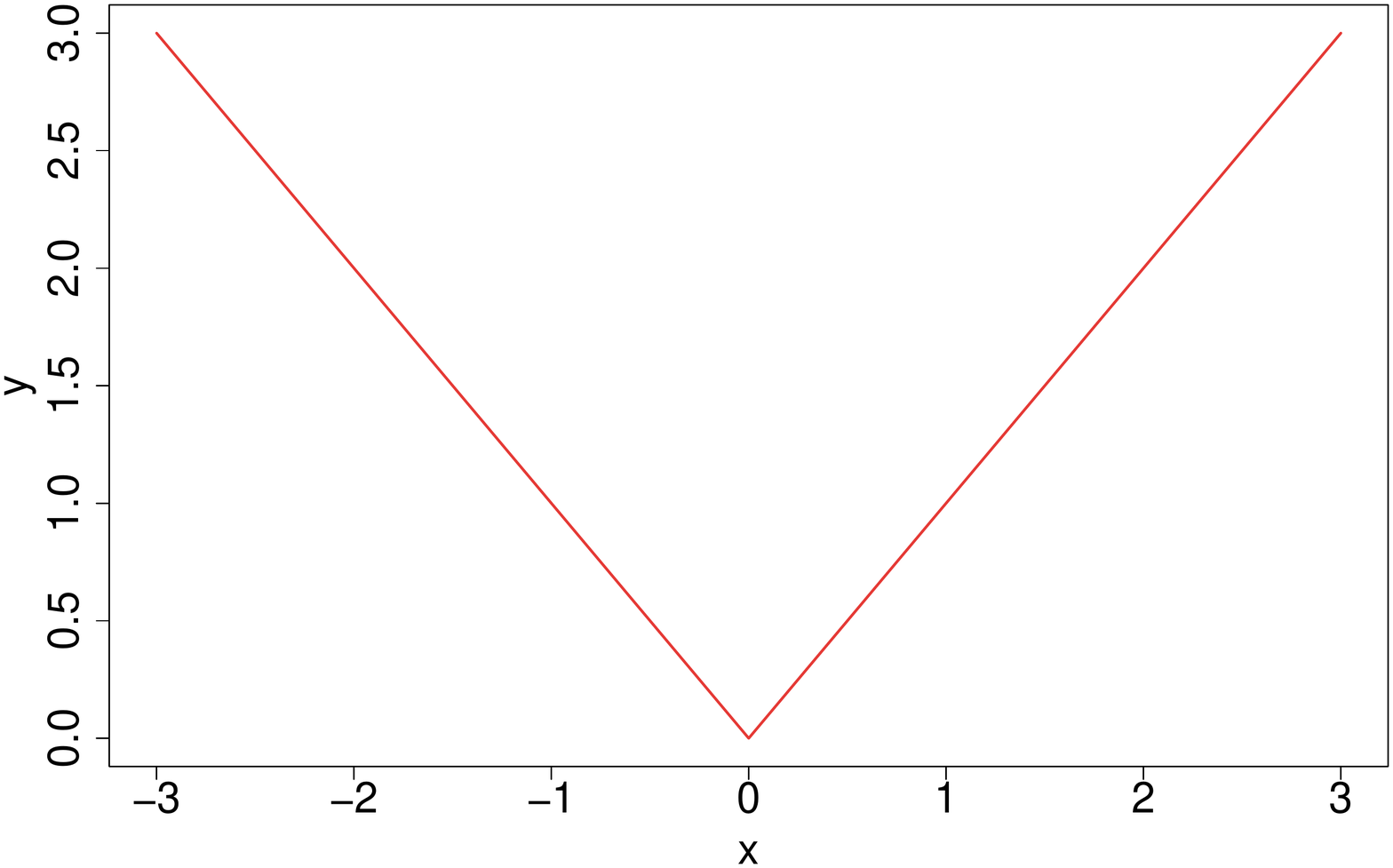}}
\hfill
\subcaptionbox{$\varepsilon$ - insensitive loss ($\varepsilon=1$)}{\includegraphics[width=0.3\textwidth]{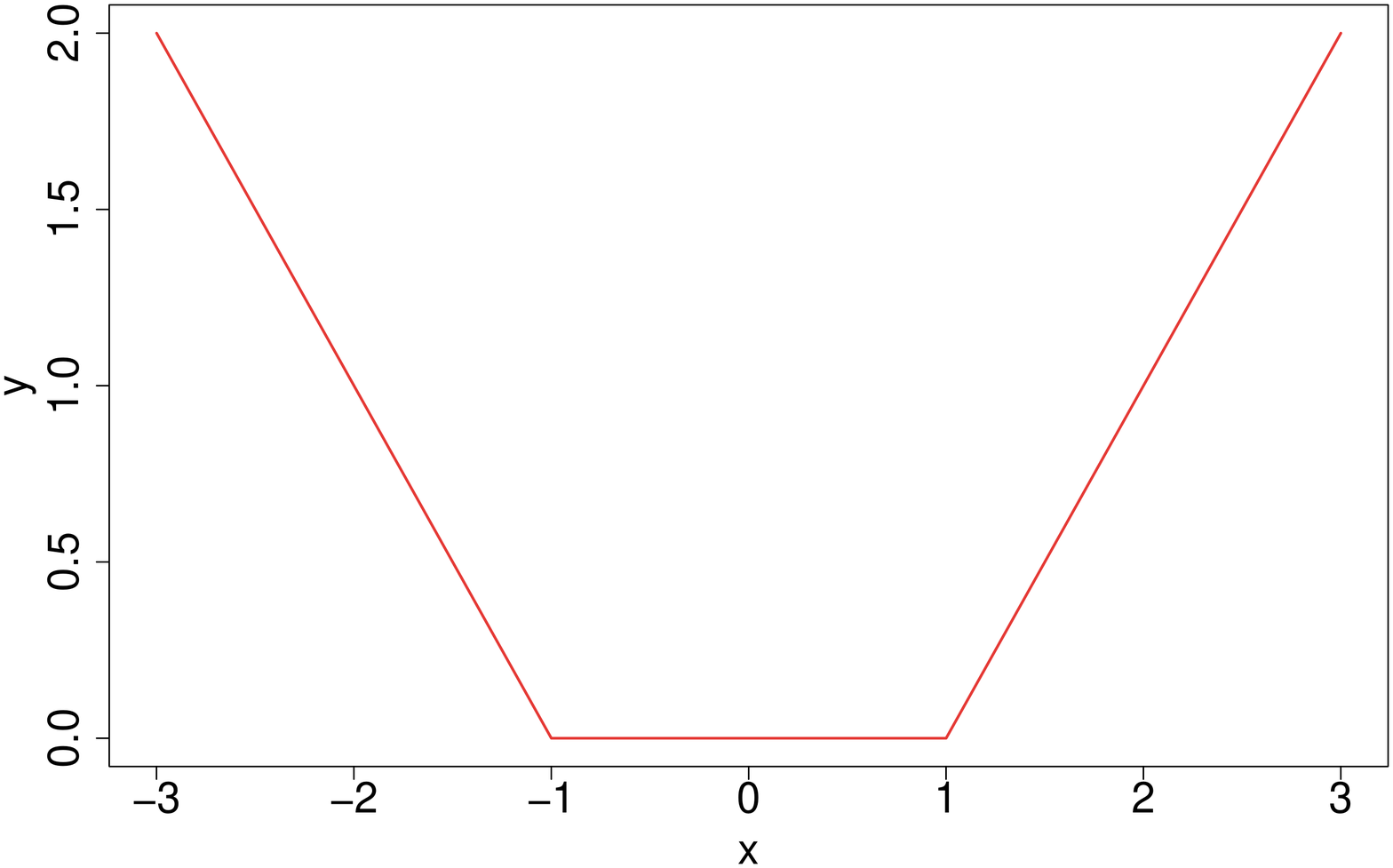}}
\caption{Loss functions for regression problem.}
\label{loss-regression}
\end{figure}

For regression problems, we can use the $L_2$ \textit{loss function}, also known as \textit{Least Square Error}, define as 
\begin{equation}\label{ls}
\ell_{LS}(y, f(x)) = (y-f(x))^2
\end{equation}
as you can see from \autoref{loss-regression}(a) the cost for correct prediction is 0, while a positive or negative difference carries a penalty. In particular, small mistakes are less accentuated. As we move away from the center, the penalty becomes substantial.\bigbreak

The $L_1$ alternative instead, has a linear increment, so we expect the cost to be linearly proportional to the mistake. This loss, also called \textit{Least Absolute Deviation}, is defined as
\begin{equation}
\ell_{LDA}(y, f(x)) = |y-f(x)|
\end{equation}

The $\varepsilon$ - insensitive loss, allows a small margin of error, that is, if the prediction error is smaller than $\varepsilon$, then there is no cost to pay, out of the $\varepsilon$ margin the penalty linearly increments just as the $L_1$ loss. Formally
\begin{equation}
\ell_\varepsilon(y, f(x)) = \max\{0,\ |y-f(x)|-\varepsilon,\}
\end{equation}
graphically, for $\varepsilon=1$, you can see it in \autoref{loss-regression}(c).

\newpage
\subsubsection{Empirical Risk Minimization}
As we said, the loss function measures how different a prediction is from the true outcome. It measures a single prediction, while we might be interested in evaluating the overall behaviour, that is, measuring the "goodness" over possible data generated by the underlying probability $\rho$. So we introduce the concept of \textit{Risk}, which is the expected loss of classifier $f$

\begin{equation}
R(f) = \int_{\mathcal{X}\times \mathcal{Y}}\ell\left(y, f(x)\right)d \rho(x,y) =
\int_\mathcal{X}\int_\mathcal{Y}\ell(y, f(x))d\rho(y|x)d\rho(x)
\end{equation}

it calculates the weighted average over all possible points in the space $\mathcal{X}$ of the loss of those points.\\

Conceptually, it allows us to compare different classifiers, so to choose the one with a smaller risk. The best classifier is the one with the smallest possible risk. \bigbreak
But, which functions $f$ are eligible for our problem? Or rather, what kind of functions can we consider? \\
Let $\mathcal{F}$ be the space of functions mapping $\mathcal{X}$ to $\mathcal{Y}$. We pick our classifier from this space. If we are to consider all possible functions from $\mathcal{X}$ to $\mathcal{Y}$, we would end up with the space:
\begin{equation}
\mathcal{F}_{all} = \{f:\mathcal{X} \rightarrow \mathcal{Y}\}
\end{equation}
In this context, among all classifiers, the optimal one, having the lowest possible risk value is the \textit{Bayes classifier}
\begin{equation}
f_{Bayes}(x) = 
\begin{cases}
1  & if\ P(Y=1 | X = x) \geq 0.5 \\
-1 & otherwise
\end{cases}
\end{equation}
For each point $x$, it looks at the conditional probability of the response being 1 ($Y=1$), given ($X=x$), and if such probability is greater than 0.5, it gives 1 as a response, otherwise, the probability is higher for the other class, so it returns -1. Intuitively this classifier is the best we could imagine, but in practice it is impossible to compute it, since the underlying probability $\rho$ is unknown, as we noted previously. \\

We look for a classifier $f$, which has risk $R(f)$ as close as possible to the risk of the Bayes classifier $R(f_{Bayes})$. But we note that computing the risk of any classifier requires the knowledge of the underlying probability $\rho$, which we do not have. \bigbreak

For this reason, we switch to what we can actually measure, that is the \textit{empirical risk}, determined over the training set. Given a classifier $f$
\begin{equation}
R_{emp}(f) = n^{-1} \sum_{i=1}^n \ell(y_i, f(x_i))
\end{equation}
is the empirical risk, computed as the average of the loss evaluated on the $n$ training set points. \\
We expect our learning algorithm to learn some classifier $f_n$ which minimizes $R_{emp}(f_n)$. However, it might not be a good idea to have an $f_n$ which makes as few errors as possible on the training set. \\
A model trained to perform perfectly on the training set, might be too closely related to that particular set, and would therefore perform poorly on new datas. This is because it would tend to learn the noise as well. 
This problem is known as \textit{Overfitting}.\\
So what we would like, is some simpler classifier, which ignores the useless noise. Hence we restrict the space of functions $\mathcal{F}$ we are considering. \\
The classifier we are looking at, satisfies

\begin{equation}
f_n = \argminB_{f\in\mathcal{F}} R_{emp}(f)
\end{equation}

Back to our problem, we would like the risk of our classifier to be as close as possible to the best possible risk, that of the Bayes classifier.
In other words, we want to minimize
\begin{equation}
R(f_n) - R(f_{Bayes})
\end{equation}
as in \citep{luxburg2011}, we can decompose it in
\begin{equation}
R(f_n) - R(f_{Bayes}) = 
\left(R(f_n)-R(f_\mathcal{F})\right)
+ \left( R(f_\mathcal{F})-R(f_{Bayes}) \right)
\end{equation}

\begin{figure}[!ht]
\centering
\includegraphics[width=0.7\textwidth]{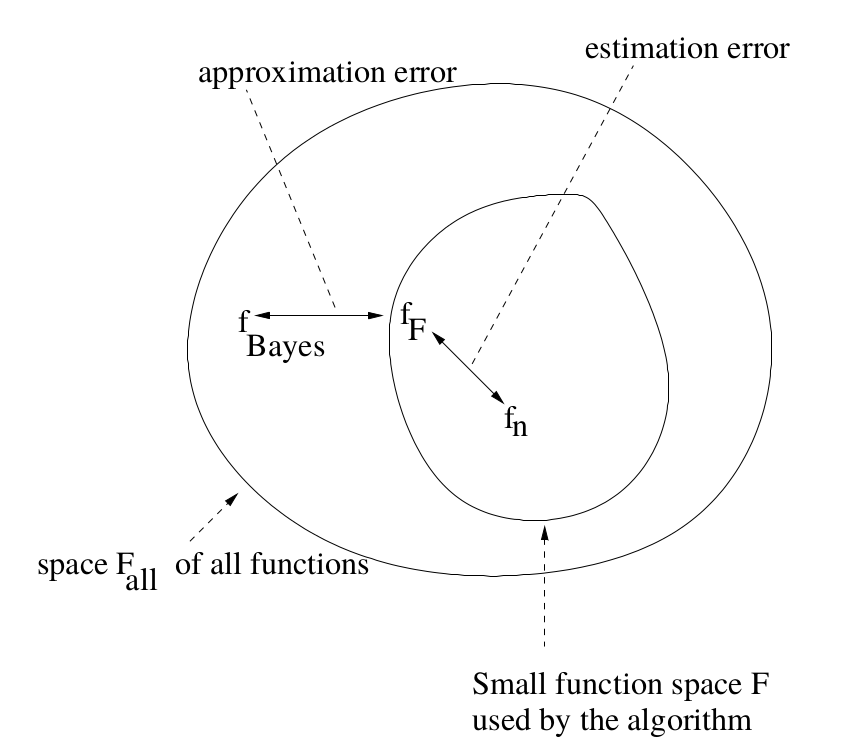}
\caption{Estimation and approximation error. \citep{luxburg2011}}
\label{errors}
\end{figure}

where $R(f_\mathcal{F})$ is the risk associated to the best classifier in the space of functions $\mathcal{F}$ we are considering. Graphically explained in \autoref{errors}. \medbreak
The first term $\left(R(f_n)-R(f_\mathcal{F})\right)$ is called \textit{estimation error} and is responsible for the sampling error. Since we are given only a sample of $X$, it might not be fully representative of the entire space, thus the algorithm might be induced in some error given by the stochasticity of sampling.\medbreak
The second term instead $\left( R(f_\mathcal{F})-R(f_{Bayes}) \right)$, accounts for the error introduced by restricting the overall function space we are considering. As we said before, we do not want our model to be too closely related to the sample we are considering, so we restrict the function space $\mathcal{F}$ to simpler functions. Thus inducing some error related to such reduction, also called \textit{approximation error}.\bigbreak

\begin{figure}[!ht]
\centering
\includegraphics[trim=4cm 4cm 4cm 5cm,clip,width=0.7\textwidth]{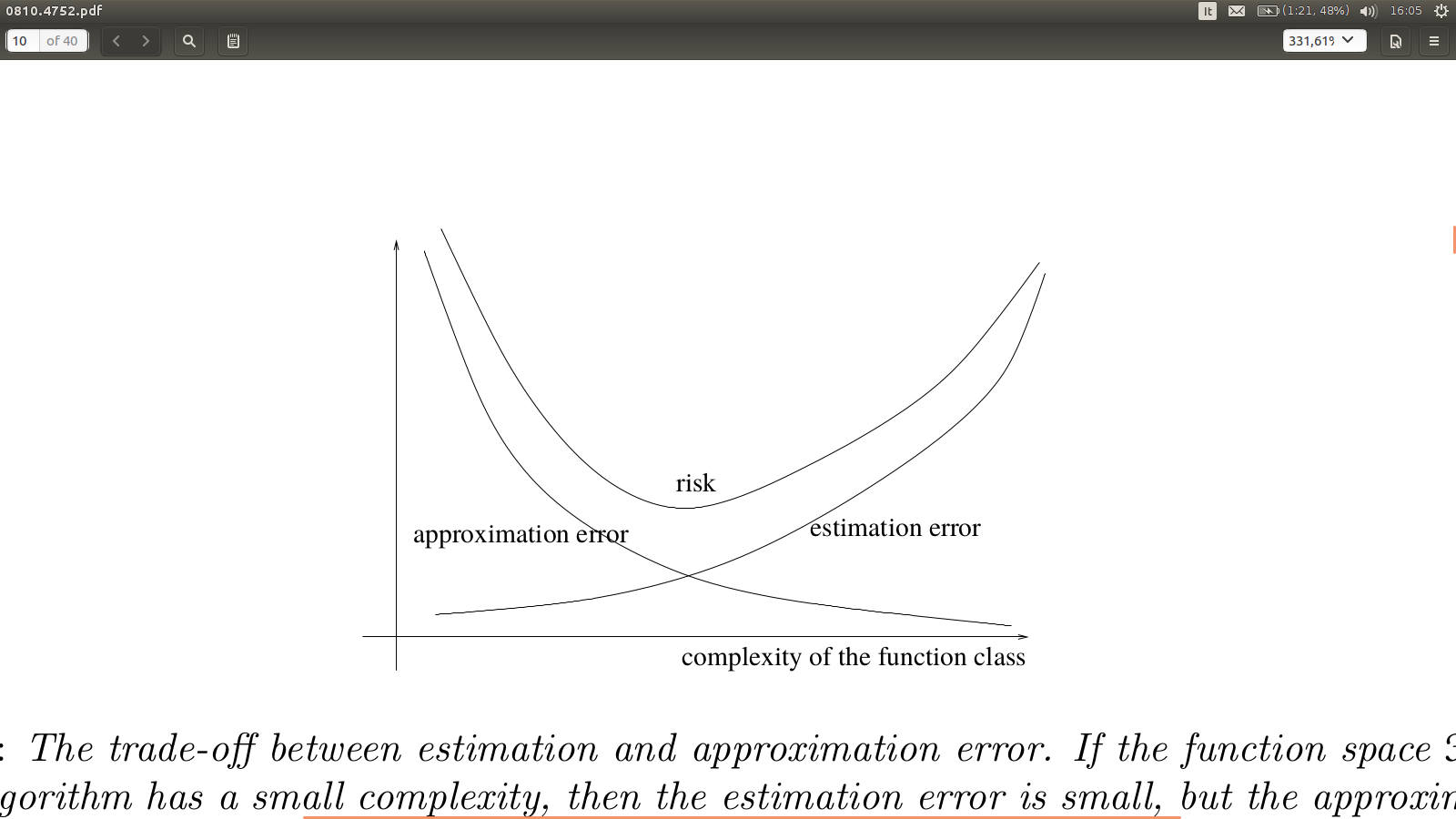}
\caption{The balance between approximation and estimation error, as well as the expected variation of the risk. The minimizer is expected to be in the balanced situation \citep{luxburg2011}.}
\label{balance_errors}
\end{figure}

The choice of the function class is the method used to control the trade-off between estimation and approximation error.\\
By considering large function spaces, we would end up including the Bayes classifier itself, or some classifier behaving similarly, thus resulting in a small approximation error. But this would lead to an increase of the estimation error, since complex functions are prone to overfitting.\\
The other way around, small function spaces would result in small estimation error at the cost of large approximation error. Too small function spaces would additionally cause the opposite effect of overfitting, referred to as \textit{underfitting}. See \autoref{balance_errors}.

A well known example of a classifier which relates well to the empirical risk minimization scheme, is the \textit{Nearest Neighbor Classifier}.\\
Basically, it classifies a new point based on the closest point in the training set. So it assumes the presence of a distance measure, and as soon as a new point is received in input, it measures the distance with all the points in the training set, picks the smallest one and assign its class to the new point. \\
Alternatively, we can consider the closest $k$ points, resulting in the variant called \textit{k-Nearest Neighbor Classifier}.\\
Why is this classifier interesting for our discussions? Because in \citep{cover1967} they proved the following bound:
\begin{equation}
R(f_{Bayes}) \le R(f_{NN}) \le 2R(f_{Bayes})
\end{equation}

which closely relates the risk of the nearest neighbor classifier with that of the Bayes classifier. In particular, it states that it is upper-bounded by twice the optimal classifier risk.
Furthermore, considering the k-NN alternative, in \citep{stone1977} they proved the following theorem
\thmn{Risk of k-NN}{Let $f_n$ be the k-nearest neighbor classifier constructed on n sample points. If $n\rightarrow \infty$ and $k\rightarrow \infty$ such that $k/n\rightarrow 0$, then $R(f_n) \rightarrow R(f_{Bayes})$ for all probability distributions P. That is, the k-nearest neighbor classification rule is universally Bayes-consistent.}

In other words, for growing number of training points, and growing number $k$, where the growth is faster for $n$, such that $k/n\rightarrow 0$, then the risk of $k$-nearest neighbor tends to that of the Bayes classifier.

\subsection{Regularization}
As we said in the previous subsection, we need to find a balanced situation in which we have enough complexity in the functional we are looking for, but not too much, otherwise we end up overfitting. In order to control the power of such functionals, we can restrict the overall function space from which we are looking to. In this case, \textit{Regularization} techniques are used to control the complexity, or the number of parameters used. We add the regularization term to the risk minimization framework discussed previously
\begin{equation}\label{pre_srm}
R_{reg}(f) = R_{emp}(f) + \lambda\Omega(f)
\end{equation}
where $\lambda>0$ is the regularization parameter, controlling the trade-off between the minimization of $R_{emp}(f)$ and the simplicity of the considered functional $f$, enforced by the term $\Omega(f)$.\bigbreak

Recall that $R_{emp}(f)$ is the average loss function over training points. We considered many loss functions, some of which convex. We might want to preserve the convexity property, since it gives the certainty of a single global minimizer. Thus we might prefer using a convex regularization term.\bigbreak

One well-known regularizer is the quadratic term $\frac{1}{2}\|w\|^2$, which is present also in the theory of the well-know Support Vector Machines. This term encourages the sum of the squares of the parameters to be small. Note that we can write $f(x) = \ang{w}{x}\ for\ w\in\r^n$, so the problem becomes 

\begin{equation}\label{risk_reg}
R_{reg}(f) = R_{emp}(f) + \frac{\lambda}{2} \|f\|^2
\end{equation}

We are going to discuss other forms of regularization but for now it is sufficient to take this one in example.\\
Based on notions introduced in \autoref{preliminaries}, as noted in \citep{smola1998}, we say that the feature space can be seen as a Reproducing Kernel Hilbert Space. So we can re-write the risk functional as

\begin{equation}\label{h_risk}
R_{reg}(f) = R_{emp}(f) + \frac{\lambda}{2} \|f\|^2_\h
\end{equation}
 
The explicit form of the minimizer of \autoref{h_risk}, is given by the so called \textit{Representer Theorem}. It is particularly useful in practical problems. Note that \hh is the Reproducing Kernel Hilbert Space associated to the kernel $k$.

\thmn{Representer Theorem}{\label{representer_thm}Denote by $\Omega:[0, +\infty) \rightarrow \r$ a strictly monotonic increasing function, by $\mathcal{X}$ a set, and by $\ell:(\mathcal{X}\times \r)^n\rightarrow\r \cup \{\infty\}$ an arbitrary loss function. Then each minimizer $f\in\h$ of the regularized risk
\begin{equation}
\ell((y_1, f(x_1)), ..., (y_n, f(x_n)) + \Omega(\|f\|_\h)
\end{equation}
admits a representation of the form 
\begin{equation}
f(x) = \sum^n_{i=1} \alpha_ik(x_i, x)
\end{equation}
}
\proof{ For convenience, consider $\bar{\Omega}(\|f\|^2)$ instead of $\Omega(\|f\|)$ without loss of generality, since the quadratic form is strictly monotonic in $[0, \infty)$ if and only if $\Omega$ also satisfies this requirement. \\
We decompose $f\in\h$ into 2 parts, one contained in the span of kernel functions \\
$k(x_1, \cdot), ..., k(x_n, \cdot)$, and the second part contained in the orthogonal complement.
\begin{equation}
f(x) = f_{n}(x) + f_{\perp}(x) = \sum^n_{i=1} \alpha_ik(x_i, x) + f_{\perp}(x)
\end{equation}
where $\alpha_i\in\r$ and $f_{\perp}\in\h$, with $\ang{f_{\perp}}{k(x_i, \cdot)}_\h=0$ for all $i\in \{1, ..., n\}$.\\
Since \hh is a RKHS, it has the reproducing property, as stated in  \defref{reproducing_property}, so we can write $f(x_j)$ as 
\begin{equation}
f(x_j) = \ang{f(\cdot)}{k(x_j, \cdot)} = \sum^n_{i=1} \alpha_ik(x_i, x_j) + \ang{f_{\perp}(\cdot)}{k(x_j, \cdot)}_\h = \sum^n_{i=1}\alpha_ik(x_i, x_j)
\end{equation}
for all $j\in \{1, ..., n\}$.\\
For all $f_{\perp}$,
\begin{equation}
\Omega(\|f\|_\h) = \bar{\Omega} \left( \left\|\sum^n_{i=1} \alpha_ik(x_i, \cdot)\right\|^2_\h + \|f_\perp\|^2_\h \right) \ge 
\bar{\Omega}\left( \left\| \sum^n_{i=1} \alpha_ik(x_i, \cdot) \right\|^2_\h \right)
\end{equation}
Thus, for any $\alpha_i\in\r$, the risk functional considered in this theorem, is minimized for $f_\perp=0$. The same holds for the solution, so the theorem is proved. \\
\qed
}

The original form of this theorem, provided in \citep{kimeldorf1971}, consisted in the particular case in which $\ell$ is the point-wise mean squared loss, and $\Omega(f) = \|f\|_\h^2$. While in the form presented in \thmref{representer_thm}, provided by \citep{smola1998}, they drop the loss function restriction, enlarging the applicability to any strictly monotonic increasing loss function. \bigbreak

The problem we are trying to solve now becomes
\begin{equation}
    \min_{f\in\h}R_{emp}(f) + \frac{\lambda}{2}\|f\|^2_\h
\end{equation}

This problem, with $f\in\h$ which is an infinite-dimensional space, is very hard to deal with. By making use of the Representer Theorem instead, we can solve it by using only \textit{m} particular kernels, by using the training points. This makes \thmref{representer_thm} fundamental. In other words, we can transform a large class of optimization problems into a problem of kernel expansions over the training points.\bigbreak

Minimizing a risk problem under the form of \autoref{pre_srm}, is known as \textit{Structural Risk Minimization}. As we said, we enrich our risk minimization problem by adding a term which penalizes complex models. 

\begin{equation}
\label{risk_comp}
\min_{f\in\h}
\frac 1 n \sum_{i=1}^n \ell(y_i, f(x_i)) + \lambda\Omega(f)
\end{equation}

Our training optimization problem is now composed of the loss term, accounting for how well the model fits the data, and a regularization term, responsible for model complexity.\bigbreak

\subsubsection{$L_2$ regularization and Ridge Regression}
The one we mentioned before, is also known as the $L_2$-regularization, which defines the regularization as a sum of squared elements. Formally
\begin{equation}
\|w\|^2_2 = w_1^2 + w_2^2 + ... + w_d^2
\end{equation}
where we use $w\in\r^d$ to represent $f$, since as we said earlier $f(x)=\ang{w}{x}$.
So each element contributes in increasing the value of \autoref{risk_comp}. For that reason, the algorithm should force some of these parameters to be close to zero, so to not increase the risk and not participate in the final model. Only strongly significant parameters should have large values, all the others should be close to 0.\bigbreak

An example of a regression method relying on $L_2$ regularization is the \textit{Ridge Regression}. It has a two term formulation. The first term is the Least Square Error loss function we described in \autoref{ls}, accounting for the learning part, so to learn the best set of parameters to reduce the error on training set. Just as we said for loss functions. While the second one is the $L_2$ regularization term, responsible for preventing overfitting by forcing irrelevant weights (those corresponding to not significant features) to get close to zero. Ridge regression is defined as

\begin{equation}
\label{ridge}
\min_{w\in\r^d}
\sum^n_{i=1} ( y_i - w^\top x_i)^2 + \lambda\sum^n_{i=1}w_i^2
\end{equation}
where the first term is minimized when the difference $y_i - w^\top x_i$ is minimized, for each training point $i = \{1, ..., n\}$. That is, given the pair $(x_i, y_i)$, the estimate of our algorithm is given by $w^\top x_i$, while the true response is $y_i$. Just as in \autoref{loss_functions}, we want our estimates $f(x_i)$ to be as close as possible to ground truth $y_i$. \\
The second term instead increases the value of \autoref{ridge} for large values of $w$, so that only important features will have corresponding large weights. This term is controlled by the coefficient $\lambda$, which is used to adjust the importance of regularization.\\
The balance of the two terms should give a balanced $w$ vector as a result. The first term is responsible for learning the best set of weights so to minimize the error on the training set, while the second reduces overfitting so to perform well on new data, producing generalization. Note that in the original formulation of ridge regression method, there is an additional parameter to learn, responsible for the intercept. However, if the data are mean-std standardized, there is no need for it, so for clarity we don't consider that parameter.
\bigbreak

\begin{equation}\label{ridge_m}
    \min_{w\in\r^d}\|Xw-y\|^2_2 + \lambda\|w\|^2_2
\end{equation}

Note that $X \in \r^{n\times d}$ is the matrix of $n$ training points, each of which $d$-dimensional, $y\in\r^n$ is the response vector for each of the $n$ points and $w\in\r^d$ the parameter vector.
The solution $w$ to this problem can be computed by taking the partial derivative w.r.t. $w$ and equating it to zero.
\begin{align*}
\frac{\partial}{\partial w} \|Xw-y\|^2_2 + \lambda\|w\|^2_2 &=
 2X^\top\left(y - Xw\right) +2\lambda w
\end{align*}
\begin{align*}
\frac{\partial}{\partial w} \|Xw-y\|^2_2 + \lambda\|w\|^2_2 &= 0 \\
2X^\top\left(y - Xw\right) + 2\lambda w &= 0 \\
2X^\top y - 2X^\top Xw + 2\lambda w &= 0 \\
2X^\top y - 2\left(X^\top X+\lambda I\right)w&= 0 \\
X^\top y &= \left(X^\top X+\lambda I\right)w \\
\left(X^\top X + \lambda I\right)^{-1}\left(X^\top y\right) &= w
\end{align*}
Therefore $w = \left(X^\top X - \lambda I\right)^{-1}\left(X^\top y\right)$. Note that adding the term $\lambda I$ makes that square matrix invertible. In the Linear regression framework, where there is no regularization term, this is not possible, indeed that matrix $\left(X^\top X\right)$ may not be invertible in particular cases. \bigbreak

In order to compute estimates for all training points, we should evaluate 
\begin{equation}
\hat{y} = Xw = X\left(X^\top X + \lambda I\right)^{-1}\left(X^\top y\right)
\end{equation}

\begin{figure}[!ht]
\centering
\includegraphics[trim=15cm 0 0 2cm,clip,width=0.7\textwidth]{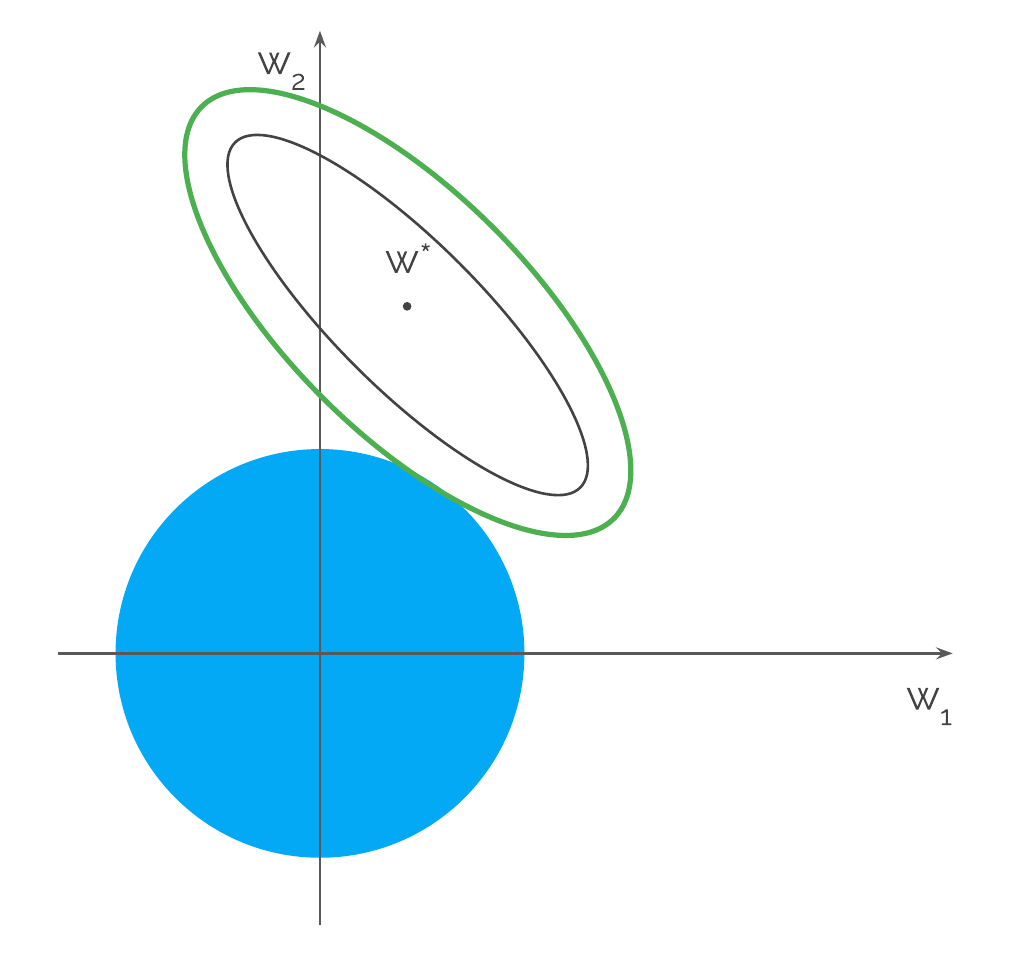}
\caption{Ridge regression objective function in a two-dimensional case. Estimation term is represented by the ellipses around the optimal point $w^*$. Regularization term is represented by the blue ball (unitary $L_2$-ball). The solution is where the two terms intersect each other.}
\label{l2_ridge}
\end{figure}

We can alternatively write \autoref{ridge_m} as 
\begin{equation}
\label{ridge_alt}
\min_{\|w\| \le \alpha}
 \|Xw-y\|^2
\end{equation}
which can be visually seen in  \autoref{l2_ridge}. Here we are considering a two-dimensional case, where $w\in\r^2$. The optimal point $w^*$ is the one that minimizes the term of \autoref{ridge_alt}. The regression term instead (the constrain $\|w\|\le\alpha$), is represented by the blue ball. In particular this is the $L_2$-ball, which contains only points $z$ such that the $L_2$-norm $\|z\|_2\le r$, where $r$ is the radius. As you would expect, all the points for which $\|z\|_2$ is exactly equal to $r$ lie on the border of this ball. The solution lies exactly on the border of the $L_2$ ball in this case, since
the Lagrangian term associated to the constraint ($\lambda\sum^2_{i=1} w_i - \alpha$) is clearly minimized for $\|w\|=\alpha$. We know the solution is
 $w=(X^\top X + \lambda I)^{-1}X^\top y$, for some $\lambda>0$, hence $\alpha = \|(X^\top X + \lambda I)^{-1}X^\top y\|$. \bigbreak 

Since the solution of the estimation term ($w^*$) alone would cause overfitting, then we force the objective function to move away from it by adding the regularization term. Thus the solution lies on the intersection of the 2 terms.

\subsubsection{$L_1$ regularization and Lasso Regression}\label{l1_section}
In alternative to the regularization discussed above, we can utilize the $L_1$ regularization defined as the $L_1$ norm, that is
\begin{equation}
\|w\|_1 = |w_1| + |w_2| + ... + |w_d|
\end{equation}
where again, $w\in\r^d$ represents $f$. Here each parameter $w_i$ adds to the term by means of an absolute value increment. The difference from the previous regularization term consists in the fact that here we are considering absolute values instead of squared values. This difference brings some interesting properties we are going to discuss in what follows.\bigbreak

The regression method utilising $L_1$ regularization is known by the acronym \textit{LASSO Regression}. It stands for \textit{Least Absolute Shrinkage and Selection Operator Regression}. Lasso method not only performs regression under a regularized scheme, but it also operates a feature selection process on the data. \textit{Feature Selection} (also known as variable selection) is the process of selecting a subset of relevant features, distinguishing them from the irrelevant ones. \bigbreak

Just as $L_2$ regularization procedure forces some parameters of $w$ to be reduced if the corresponding features are not significant, so does $L_1$ regularization. The difference is that in the latter case, weights are forced to be exactly $0$ for those irrelevant features, thus promoting sparsity. Here sparsity is intended as the reduced number of features having $w_i > 0$.\bigbreak

\begin{figure}[!ht]
\centering
\includegraphics[width=0.98\textwidth]{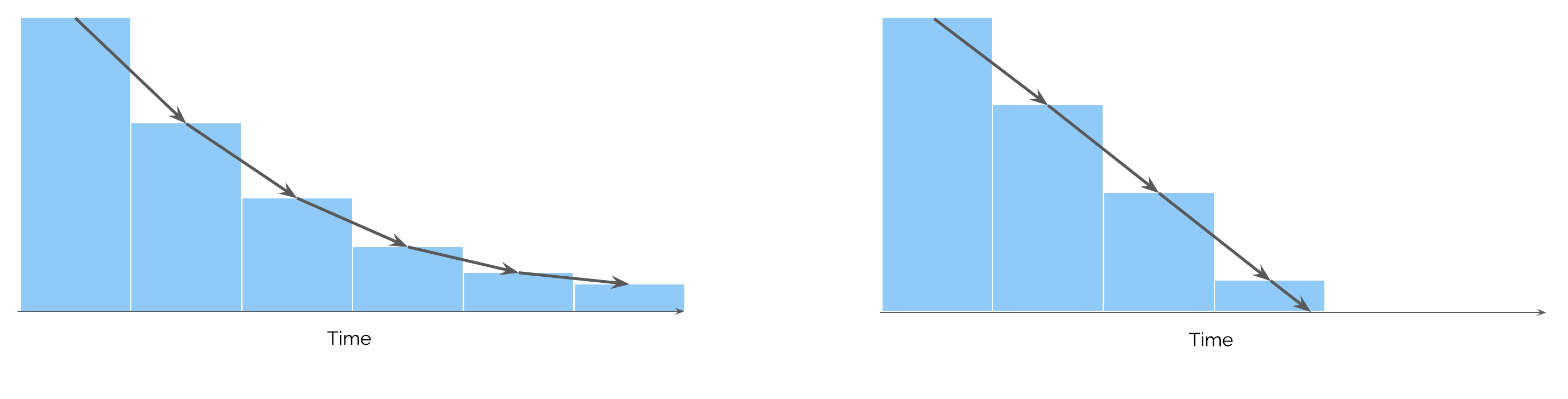}
\caption{Expected situation for convergence changes over iterations (time), for an irrelevant feature. Left is the case for $L_2$-norm regularization, where the decrease is relative to the actual weights. Right is the case for $L_1$-norm regularization where the decrease is constant.}
\label{reg_der}
\end{figure}

Why is it so? An intuition to explain the reason for sparsity relies on structural difference between $L_2$ and $L_1$ norms. If we consider their derivatives, we have that
\begin{itemize}
\item $L_2$ derivative is equal to $2*w$
\item $L_1$ derivative is equal to some constant $c$ not depending on actual weights
\end{itemize}
which means that when we try to minimize functions containing such terms, in the case of $L_2$ we are going to consider values decreasing by some amount relative to the last iterations weight, while in the case of $L_1$ we consider a constant decrease $c$ at each iteration. Hence, the situation for irrelevant features is expected to be the one in \autoref{reg_der}.

As you can see from the figure, in the case of $L_2$-norm, updating the value of $w$ for the next iteration $(i+1)$ results in $w^{(i+1)} = w^{(i)}-2*w^{(i)}$, which is expected never to reach zero. In the $L_1$ case instead the update becomes $w^{(i+1)} = w^{(i)} - c$, which undoubtedly reaches $0$ (subtraction that crosses $0$, is equal to $0$ because of absolute value).\\
Another intuition is going to be added later on.\bigbreak

Lasso regression is formulated as the following problem
\begin{equation}
\min_{w\in\r^d}
\sum^n_{i=1} \left(y_i - w^\top x_i \right)^2 + \lambda\sum^d_{i=1}|w_i|
\label{lasso}
\end{equation}
where the first term is the Least Square Error loss function described in \autoref{ls}, responsible for learning the best set of weights minimizing the error for the training set, just as in Ridge Regression, while the second term prevents overfitting by means of regularization. Writing \autoref{lasso} in matrix notation we end up with

\begin{equation}\label{lasso_m}
\min_{w\in\r^d}
{\|Xw - y\|^2_2 + \lambda\|w\|_1}
\end{equation}
where $X\in\r^{n\times d}$ is the matrix of $d$-dimensional data points, $w\in\r^d$ is the vector of weights to learn and $y\in\r^n$ are the true responses of training points. $\lambda$ is the penalty coefficient determining the amount by which the regularization term affects the objective function. In this case, the second term is referred to as \textit{Lasso Penalty}. \\

\begin{figure}[!ht]
\centering
\includegraphics[width=0.7\textwidth]{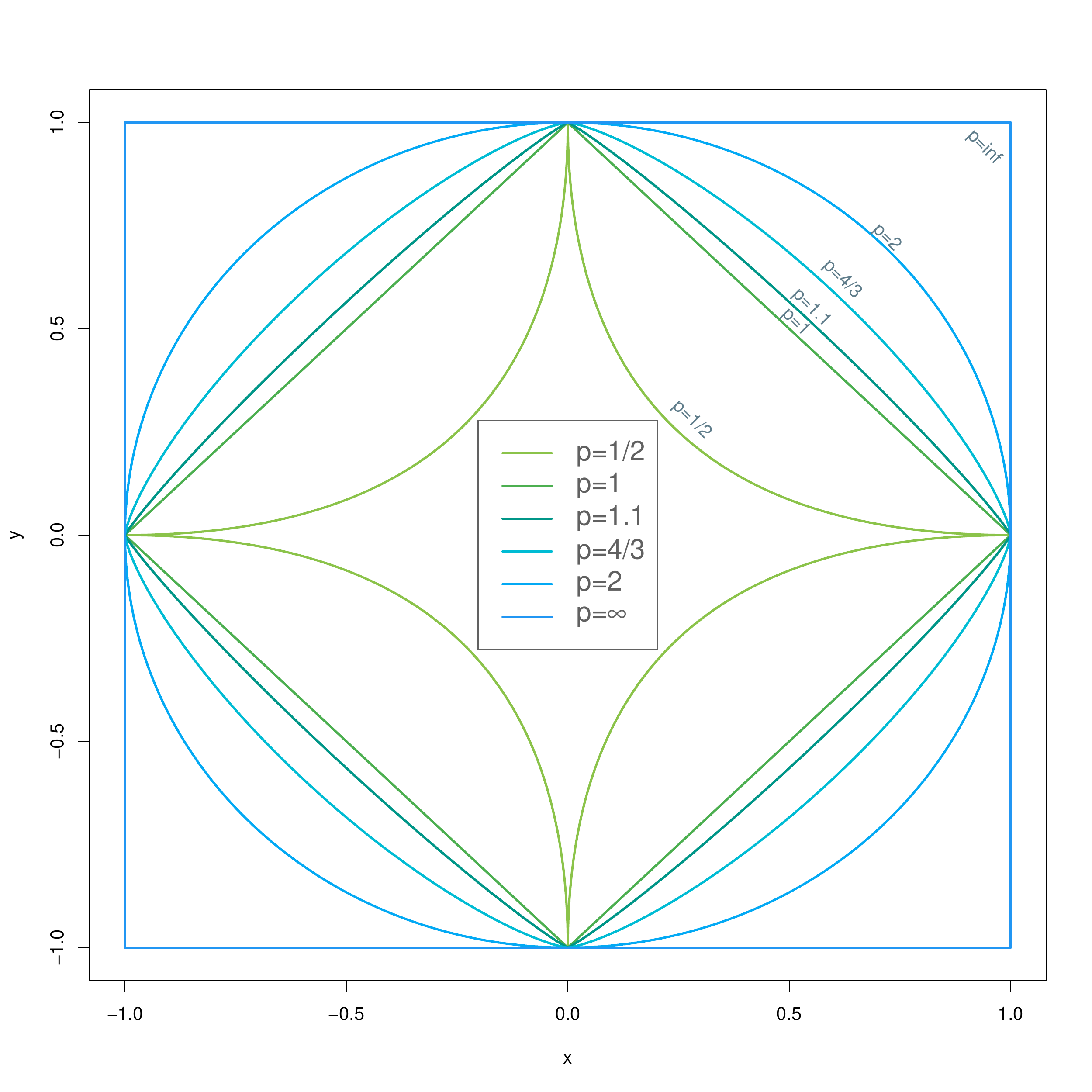}
\caption{Unitary $L_p$-norm balls for different values of $p$. }
\label{pnorm_balls}
\end{figure}

If we were to consider the parallel case of the example in \autoref{l2_ridge}, we should keep in mind that the difference relies on the fact that the $L_1$-ball has a different shape. As we said, $L_2$-ball contains all points $z$ such that $\|z\|_2 \le r$, for some radius $r$. Similarly, the $L_1$-ball contains all the points $z$ such that $\|z\|_1 \le r$, for some radius $r$. This reasoning can be extended to any other $L_p$ case, resulting in balls of different appearance. You can see an example in \autoref{pnorm_balls}.\bigbreak

\begin{figure}[!ht]
\centering
\includegraphics[trim=0 0 15cm 2cm,clip,width=0.7\textwidth]{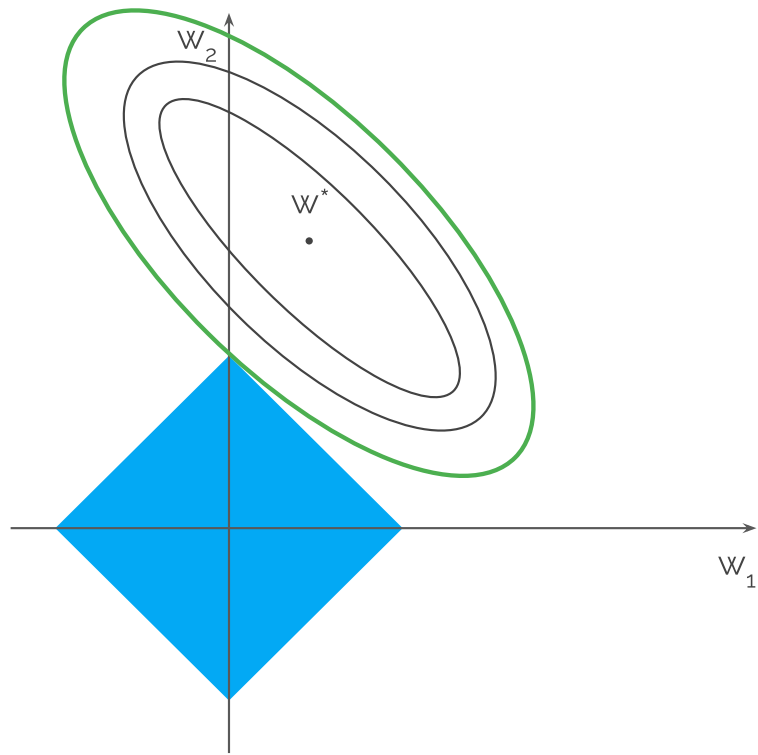}
\caption{Lasso regression objective function in a two-dimensional case. Estimation term is represented by the ellipses around the optimal point $w^*$. Regularization term is represented by the blue ball (unitary $L_1$-ball). The solution is where the two terms intersect each other.}
\label{l1_lasso}
\end{figure}

To visually see the difference of using Lasso regression, look at \autoref{l1_lasso}. We consider a two-dimensional case, so the space of $w$ can be clearly seen. The first term of objective function \autoref{lasso_m} learns the best weights, here noted by $w^*$, while the regularization term imposes a penalty for large values of $w_1$ and $w_2$. In this case, the regularization ball is the one in $L_1$ space, which makes the difference from ridge regression.\\

The minimizer of \autoref{lasso_m} stands in the intersection of the two terms. What is interesting to note is the fact that the intersection for this particular case shown in \autoref{l1_lasso} consists of one of the two entries of $w$ to be exactly equal to $0$, since it stands over the second axis, where $w_2=0$. \\

\begin{figure}[!ht]
\centering
\includegraphics[width=0.48\textwidth]{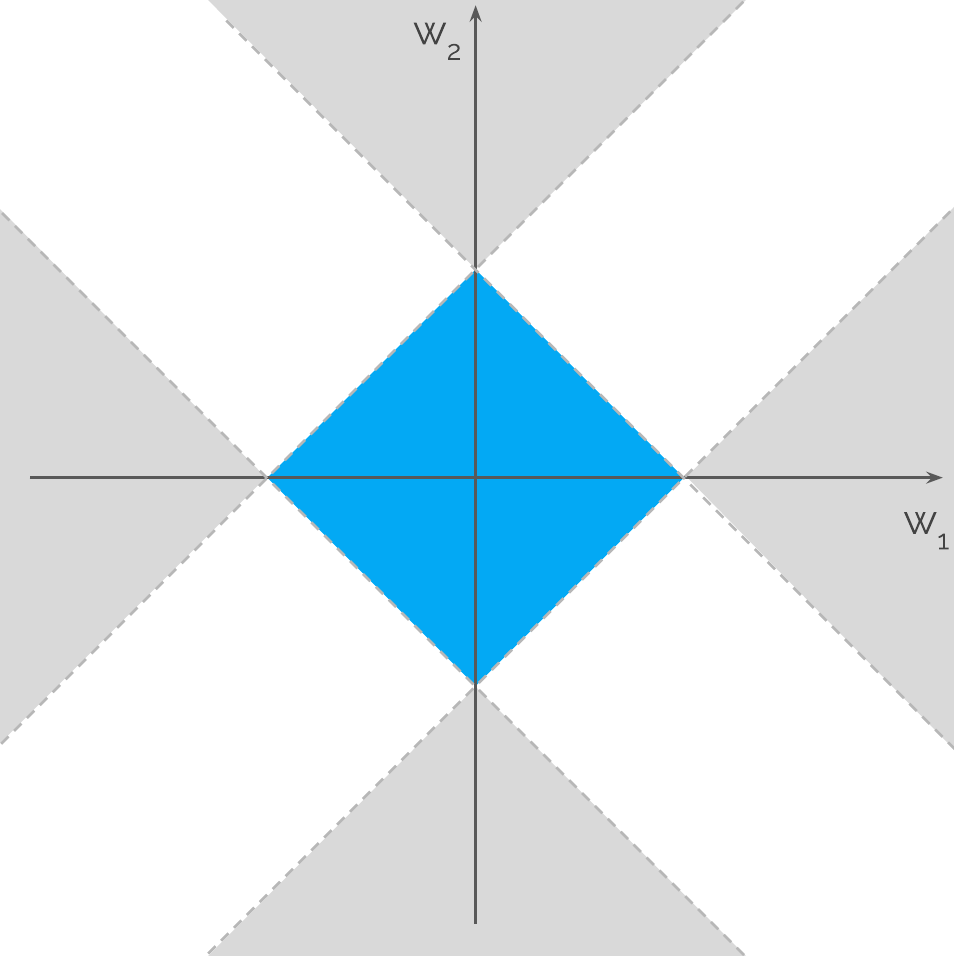}
\caption{Visual insight of the starting area which supposedly gives sparsity whenever $w^*$ sits in there.}
\label{l1_area}
\end{figure}

This is given by the particular shape of the $L_1$-ball, which makes it more likely to touch it over the axis, on its edges. 
To give a better intuition behind this phenomenon, \autoref{l1_area} shows the expected areas in which $w^*$ should be, to provide this sparse effect. Imagine, just as in the previous figure, that starting from some point $w^*$, we enlarge the area in circles around it. Then if we started from the area marked in gray in \autoref{l1_area}, we are going to touch the $L_1$ ball exactly on the edge. While if the starting point is in the rest of the space, this effect is not going to be there.

We can imagine $L_p$ balls, for $p<1$, to be more edgy, thus promoting the likeliness to intersect them exactly over the axes, where edges are. Look at \autoref{pnorm_balls}.

This consideration is easily extendable to higher dimensions, increasing the likeliness for such sparsity. \\

\subsection{Kernel Ridge Regression}\label{kernel_ridge}
Ridge regression is a good starting point to observe how such methods can be kernelized. In what follows we are going to see how we can make use of kernel functions to solve this problem. Note that here we deal with parametric models, that is we consider $w\in\r^d$ a finite number of parameters. After we expose the methodology for this setting we are going to show in \autoref{tensor_section} how we can work with non-parametric models as well, that is infinite-dimensional models having infinite parameters ($w\in\ell^p$).\\
Consider the following formulation of ridge regression:
\begin{equation}\label{classic_primal}
\min_{w\in\r^d}
{\frac{\gamma}{2}\|Xw-y\|^2_2 + \frac{1}{2}\|w\|^2_2}
\end{equation}

We want to switch to the dual formulation and in order to do so we should re-write the problem by considering the substitution $r = Xw-y$, that is the loss function term.
From now on we are going to use the $L_2$ norm, so the sub-index $2$ of the norm symbol is going to be omitted.
\newpage
\begin{align}
\min_{w\in\r^d}
&\ \frac{\gamma}{2}\|r\|^2 + \frac{1}{2}\|w\|^2\\
s.t. &\ r-Xw-y=0 \nonumber
\end{align}

Now, we introduce a vector of $n$ Lagrangian multipliers $(\alpha_1, ...\alpha_n)^\top \in\r^n$ and compute the Lagrangian formulation by multiplying each constraint with the corresponding Lagrangian multiplier. Note that the constraint $r-Xw-y=0$ is the vectorial form for $n$ different constraints, each one for each training point. Note also that here we have no constraints of the form $g(w_i)\le 0$, so the Lagrangian function is going to have one less term. So the Lagrangian function is

\begin{equation}
L(w, r, \alpha) = \frac{\gamma}{2}\|r\|^2 + \frac{1}{2}\|w\|^2 + 
\alpha^\top(r-Xw+y)
\end{equation}

We proceed by computing the partial derivatives w.r.t. $w$ and $r$, and set them to zero.
\begin{align}
0 = \frac{\partial L(w, r, \alpha)}{\partial w} &= 
w - X^\top \alpha
\nonumber \\
w &= X^\top \alpha \label{sub_w}
\end{align}
and
\begin{align}
0 = \frac{\partial L(w,r,\alpha)}{\partial r} &= 
\gamma r + \alpha \nonumber \\
r &= -\frac{\alpha}{\gamma} \label{sub_r}
\end{align}

We operate the above substitutions and re-write the Lagrangian 
\begin{align}
\min_{w,r\in\r^n} L(w, r, \alpha) &= \frac{1}{2}\|r\|^2 + \frac{\gamma}{2}\|w\|^2 + 
\alpha^\top(r-Xw+y)
\end{align}
w.r.t. $\alpha$ obtaining the equivalent form

\begin{align}
\max_{\alpha\in\r^n} L(w(\alpha), r(\alpha), \alpha) &=
	\frac{1}{2}\|\frac{\alpha}{\gamma}\|^2 +
	\frac{1}{2}\|X^\top \alpha\|^2 +
	\alpha^\top(-\frac{\alpha}{\gamma} - XX^\top\alpha + y) \nonumber \\
&=
	\frac{1}{2}\|\frac{\alpha}{\gamma}\|^2 -
	\frac{1}{\gamma}\|\alpha\|^2 +
	\frac{1}{2}\|X^\top \alpha\|^2 -
	\|X^\top\alpha\|^2 + y^\top\alpha \nonumber \\
&=
	-\frac{1}{2\gamma}\|{\alpha}\|^2 -
	\frac{1}{2} \|{X^\top \alpha}\|^2 +
	y^\top\alpha
\end{align}

The above maximization problem is equivalent to the following minimization one (by changing the sign)
\begin{equation}\label{classic_dual}
\min_{\alpha\in\r^n}
{\frac{1}{2}\|X^\top \alpha\|^2 +\frac{1}{2\gamma}\|\alpha\|^2 -y^\top\alpha}
\end{equation}

and the optimality conditions for the primal problem are

\begin{align}
0 =\frac{\partial}{\partial w} 
\frac{\gamma}{2}\|Xw-y\|^2 
+ \frac{1}{2}\|w\|^2
&=
	\gamma X^\top\left(Xw-y\right) + w \nonumber\\
0&= 
	X^\top\left(Xw-y\right) + \frac{w}{\gamma}
\end{align}

while the optimality conditions for the dual problem are

\begin{align}
0= \frac{\partial}{\partial\alpha} \frac{1}{2}\|X^\top \alpha\|^2 +\frac{1}{2\gamma}\|\alpha\|^2 -y^\top\alpha
&=
	X\left(X^\top\alpha\right) + \gamma^{-1}\alpha - y \nonumber\\
0&=
	XX^\top\alpha - y + \gamma^{-1}\alpha
\end{align}

Note that, since we operated the substitution in \autoref{sub_w}, then in order to retrieve $w$ starting from the solution of the dual problem $\alpha$, we should compute
\begin{equation}
w = X^\top\alpha = \sum^n_{i=1} \alpha_ix_i
\end{equation}
which is going to be the unique solution of the problem in \autoref{classic_primal}. This is the starting point for deriving the \textit{representer theorem} we mentioned in \thmref{representer_thm}. That is, if we want to calculate the result for an unseen point $x$, we have to compute $\ang{w}{x}$, but as we just said, $w$ can be written as a linear combination of training points $x_1, ...x_n$, thus resulting in 

\begin{equation}\label{ridge_kernel_test}
\ang{w}{x} = \sum^n_{i=1}\alpha_i\ang{x_i}{x} 
= \sum^n_{i=1}\alpha_ik(x_i, x)
\end{equation}
where $k:\r^d \times \r^d \to \r$ is the \textit{linear kernel function} defined as $k(x_i, x_j) = \ang{x_i}{x_j}$. \\
Furthermore, we can write the objective function in \autoref{classic_dual} in a kernelized manner, since the first term contains an inner product of the form $XX^\top $. Thus we substitute $XX^\top $ with the kernel matrix $K$ (\textit{Gram Matrix}). This matrix is defined as a $n\times n$ matrix of $k$ with respect to $x_1, ..., x_n$, where each entry
\begin{equation}
K_{ij} = k(x_i, x_j).
\end{equation}

So we finally write \autoref{classic_dual} as
\begin{equation}\label{classic_dual_kernel}
\min_{\alpha\in\r^n}
{\frac{1}{2}k(x_i, x_j)\alpha_i\alpha_j +\frac{1}{2\gamma}\|\alpha\|^2 -y^\top\alpha}
\end{equation}
where we simply made the above substitution for $XX^\top $, while $\alpha_i\alpha_j$ comes out of the norm squared.\\

So now we can solve this dual problem, making use of a kernel function, which in this case is the linear kernel, but it can be extended also to a general kernel of the form
\begin{equation}
K(x, x') = \ang{\Phi(x)}{\Phi(x')} = \text{sum}(\Phi(x)\odot\Phi(x'))
\end{equation}
for some nonlinear feature map $\Phi: \r^d \rightarrow \ell^2$ (recall the canonical feature map introduced in \lemref{feature_map}), where $\odot$ is the Hadamard product (element-wise).\bigbreak

We mention some notable kernel functions together with their definition
\begin{itemize}
\item Linear kernel: $k(x, x') = \ang{x}{x'}$
\item Polynomial kernel of degree $s$: $k(x, x') = \ang{x}{x'}^s$
\item Gaussian kernel (also known as Radial Basis Function kernel): $k(x, x') = exp\left(-\frac{\|x-x'\|^2_2}{2\sigma^2} \right) $, 
where $\sigma$ is a free parameter.
\end{itemize}\bigbreak

\subsection{Support Vector Machine}\label{svm}
Support Vector Machine (SVM) is a popular technique used for classification and successively modified for regression and similar problems.  An important characteristic of SVM is that its parameters corresponds to a convex optimization problem, meaning that local solutions are also global.\\ 

The basic principle of SVM is that of learning a classifier of the form
\begin{equation}
f(x) = w^\top x + b
\end{equation}
in a 2 dimensional case the discriminant is a line, where $w$ is the normal to the line and $b$ the bias. $w$ is also know as the \textit{weight vector}. In a multidimensional case, we deal with hyperplanes instead. 

\begin{figure}[!ht]
\centering
\includegraphics[width=0.4\textwidth]{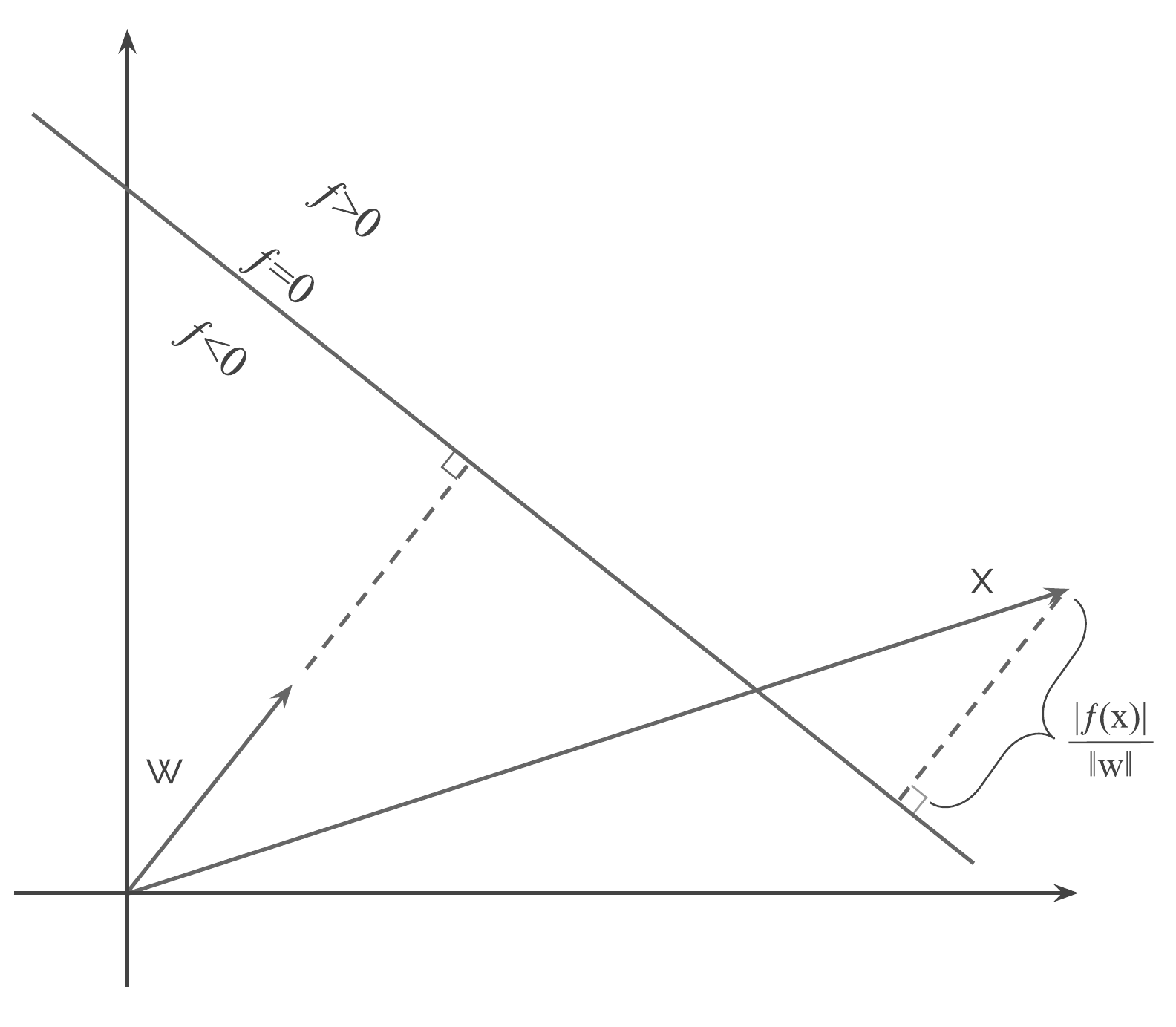}
\caption{Svm parameters shown in a two-dimensional case.}
\label{svm_1}
\end{figure}

Graphically the situation is depicted in \autoref{svm_1}. We can see the discriminant, that is the line we are interested to learn in order to make classification, the $w$ vector, and some point x lying on that space. If we want to project $x$ into $f$ we should make the following reasoning:
\begin{equation}
x = x_\perp + r \frac{w}{\|w\|}
\end{equation}
where we are going to compute the value of $r$. Equivalently
\begin{equation}
x_\perp = x - r \frac{w}{\|w\|}
\end{equation}
Since $x_\perp$ belongs to the plane $f$, then for the definition of $f$ itself we have
\begin{align}
w^\top x_\perp +b &= 0 \\
w^\top \left(x-r\frac{w}{\|w\|} \right) +b &= 0 \\
w^\top x - r \frac{w^\top w}{\|w\|} +b &= 0 \\
w^\top x -r\|w\| +b &= 0 \\
\frac{w^\top x + b}{\|w\|} &= r \\
\frac{f(x)}{\|w\|} &= r
\end{align}

So the distance from some point $x$ to the plane $f$ is equal to $f(x)/\|w\|$. This is important to keep in mind since the main idea of SVM is that of finding the best discriminant function that linearly separates the data. But, among all such classifiers, choose the one that maximizes the margin to the data points. \bigbreak

We normalize $w$ and $b$ such that 
\begin{itemize}
\item $w^\top x+b = +1$ for positive support vectors
\item $w^\top x+b = -1$ for negative support vectors
\end{itemize}
where \textit{support vectors} are those vectors which are the closest to the plane $f$, in particular they are exactly at $f(x) = 1$ and $f(x) = -1$. Keeping this in mind, given a typical statistical learning setting, with training samples $(x_1, y_1), ... , (x_n, y_n)$ with $x_i$ training point in a d-dimensional space and $y_i\in\{+1, -1\}$, then we want is to maximize such margin, so we can write it down as

\begin{align}
\max_{w\in\r^d} &\ {\frac{2}{\|w\|}} \\
s.t. &\ {w^\top x+b \ge +1}\quad{if\ y_i=+1} \nonumber\\
     &\ {w^\top x+b \le -1}\quad{if\ y_i=-1} \nonumber
\end{align}

or, equivalently

\begin{align}
\min_{w\in\r^d} &\ {\frac{1}{2}\|w\|^2} \\
s.t. &\ {y_i(w^\top x_i +b)\ge +1} \quad {for\ i=\{1, ..., n\}} \nonumber
\end{align}

this is a quadratic optimization problem, subject to $n$ linear constraints. To solve this problem we introduce $n$ Lagrangian multipliers $(\alpha_1, ..., \alpha_n)$, each corresponding to a constraint, and write the constraints as $y_i(w^\top x_i +b)-1 \le 0$, thus ending up with the Lagrangian function

\begin{equation}\label{svm_primal}
L(w, b, \alpha) = \frac{1}{2} \|w\|^2 - 
\sum^n_{i=1} \alpha_i[y_i(w^\top x_i +b) -1]
\end{equation}

Calculating the derivatives w.r.t. $w$ and $b$ and equating them to $0$, we obtain

\begin{align}
\frac{\partial L(w,b,\alpha)}{\partial w} &= w - \sum^n_{i=1} \alpha_iy_ix_i \\
w &= \sum^n_{i=1} \alpha_iy_ix_i
\end{align}

\begin{align}
\frac{\partial L(w,b,\alpha)}{\partial b} &= - \sum^n_{i=1} \alpha_iy_i \\
 \sum^n_{i=1} \alpha_iy_i &= 0
\end{align}

Making the corresponding substitutions for $w$ and $b$ to \autoref{svm_primal}, we obtain
\begin{equation}
L(w(\alpha), b(\alpha), \alpha) = 
\sum^n_{i=1} \alpha_i -
\frac{1}{2}\sum^n_{i=1}\sum^n_{j=1} \alpha_i\alpha_jy_iy_jx_ix_j 
\end{equation}

and the optimization problem now becomes
\begin{align}\label{svm_dual}
\max_\alpha 
&\ {L(w(\alpha), b(\alpha), \alpha) = \sum^n_{i=1} \alpha_i -
\frac{1}{2}\sum^n_{i=1}\sum^n_{j=1} \alpha_i\alpha_jy_iy_jx_ix_j} \\
s.t.&\ {\sum^n_{i=1}\alpha_iy_i = 0}\nonumber\\
 &\ {\alpha_i \ge 0}\quad{for\ i=\{1, ..., n\}}\nonumber
\end{align}

Note that even though we added $n$ variables $\alpha$ to the problem, we are effectively reducing its complexity because most of $\alpha_i$ are equal to 0. In particular, only those corresponding to points considered Support Vectors are $\alpha_i \neq 0$. \bigbreak

To conclude, we can write the discriminant function for a given point $x$ as
\begin{equation}
f(x) = w^\top x+b = \sum^n_{i=1} y_i\alpha_ix_i^\top x
\end{equation}

Note that we have an inner product structure appearing both in the learning and testing phases. As we discussed in \autoref{kernel_ridge}, we can make use of such structure to add kernel functions into the game. In particular we can consider the alternative formulation for \autoref{svm_dual}, where we operate the substitution of the inner product by a linear kernel function $k:\r^d\times \r^d\rightarrow \r$ resulting in
\begin{equation}
L(w(\alpha), b(\alpha), \alpha) = \sum^n_{i=1}\alpha_i 
- \frac{1}{2}\sum^n_{i=1}\sum^n_{j=1}\alpha_i\alpha_jy_iy_jk(x_i, x_j)
\end{equation}

and the testing function becomes
\begin{equation}
f(x) = \sum^n_{i=1}y_i\alpha_ik(x_i, x)
\end{equation}

Note that again this is not limited to the linear kernel function.

\newpage
\section{\secfour}\label{tensor_section}
\newcommand{\lone}{$\ell^1$ }

In this chapter we are going to discuss tensor kernel methods and how they arise. We will start by making a link to Kernel Ridge Regression, extend it to the $\ell^p$ case and conclude by showing some tensor kernel functions.
\bigbreak

As we were discussing in the previous chapter, there are different regularization terms we can use to tackle the regression problem. In particular we showed in \autoref{l1_section} how using \lone regularization brings the benefit of sparsity. That is, the resulting vector of weights exhibits the tendency of having many entries equal to zero.\bigbreak

We showed how we can retrieve a kernel representation of the ridge regression problem, but didn't do the same for lasso regression. The reason is that the structure of the problem \autoref{lasso_m} does not admit a kernel representation. We are going to explain why in what follows.\bigbreak

$\ell^2$ regularization relies on the $\ell^2$ norm, which relies in an inner product structure, that is, a Hilbert space. Moving to the \lone norm, as in the case of lasso regression, we lose the inner product structure, indeed we move to a Banach space. In general \lone regularization methods cannot be kernelized. \\
Moreover, defining a representer theorem comes with severe restrictions, which make it unfeasible to use. In other words, using \lone sparsity methods seems unpractical, since a proper kernel function cannot be used as efficiently as in the $\ell^2$ case.\bigbreak

However, it was noted in \citep{koltchinskii2009} that using an \lp space, with $p\in ]1,2[$ arbitrary close to $1$, can be seen as a proxy to the \lone case, in terms of sparsity. Indeed, recalling the discussion carried out for the $\ell^2$ case first, in \autoref{l2_ridge}, and \lone case afterward, in \autoref{l1_lasso}, we noted how the variation on the considered norm changes the situation and promotes sparsity. Furthermore, we observed in \autoref{pnorm_balls} that there are many other \lp norm balls, which are visually similar to the \lone ball. So we can perceive by intuition how using one of the other $p$-norms can possibly promote sparsity as well, at a lower degree.
Indeed this is the case.\bigbreak

Regarding the kernelization aspect, it was shown in \citep{salzo2017} that for certain values of $p\in ]1,2[$, the \lp regularization method can indeed be kernelized. This is possible, provided that a suitable definition of a tensor kernel is introduced. In what follows we are going to formally motivate this.\bigbreak

In analogy with kernel ridge regression as exposed in \autoref{kernel_ridge}, we proceed by first considering the primal problem, then moving to the dual representation and make use of a representer theorem to introduce a suitable kernel version of the dual problem.\bigbreak

The regression problem in the \lp regularization scheme is written as
\begin{equation}\label{tensor_primal}
\min_{w\in\r^d}
{\frac{\gamma}{2}\|Xw-y\|^2 + \frac{1}{p}\|w\|^p_p}
\end{equation}
where $X\in\r^{n\times d}$ is the matrix of training points, $y\in\r^n$ are the corresponding responses and the \lp norm regularization term is used. Note that here we are restricting the analysis to the finite-dimensional case, in analogy with kernel ridge regression, but this procedure can easily be extended to non-parametric models as well. \bigbreak

We proceed by first making the substitution $r = Xw$ 
\begin{align}
\min_{w\in\r^d}&\ 
{\frac{\gamma}{2}\|r-y\|^2 + \frac{1}{p}\|w\|^p_p}{}{}\\
s.t.&\ {r = Xw} \nonumber
\end{align}

So that now we have the constraints written on the form needed to introduce the Lagrangian multipliers $(\alpha_1, ..., \alpha_n)^T \in \r^n$. 
The Lagrangian function is thus
\begin{equation}\label{tensor_lag}
L(w,r, \alpha) = \frac{\gamma}{2}\|r-y\|^2 + \frac{1}{p}\|w\|^p_p 
+ \alpha^\top(r-Xw)
\end{equation}
Before proceeding with the calculations of partial derivatives and the corresponding dual problem, we first introduce $J_p:\r^d\rightarrow\r^d$ and $J_q:\r^d\rightarrow\r^d$ as the gradients of $\frac{1}{p}\|\cdot\|^p_p$ and $\frac{1}{q}\|\cdot\|^q_q$ respectively, where $\frac{1}{p}+\frac{1}{q}=1$. The composition of these two functions $J_p\circ J_q = Id$ and we prove it as follows:\\
Let $x\in\r^n$, then
\begin{equation}
J_p(x) = \left(sign(x_i)|x_i|^{p-1} \right)_{1\le i\le n}
\end{equation}
similarly, given $z\in\r^n$
\begin{equation}
J_q(z) = \left(sign(z_i)|z_i|^{q-1} \right)_{1\le i\le n}.
\end{equation}
The composition
\begin{align}
J_p(J_q(z)) &= sign(J_q(z_i))|J_q(z_i)|^{p-1}\quad &for\ 1\le i\le n \nonumber\\
&= |z_i^{q-1}|^{p-1} &for\ 1\le i \le n \label{jpq_1}
\end{align}
but recalling that $\frac{1}{p}+\frac{1}{p}=1$ we know that $p-1 = \frac{p}{q}$ and $q-1=\frac{q}{p}$ so \autoref{jpq_1} is equivalent to $|z_i|,\ for\ 1\le i\le n$.

\bigbreak
Calculating the partial derivatives \textit{w.r.t.} $w$ and $r$ we obtain
\begin{align}
\frac{\partial L(w,r,\alpha)}{\partial w} &=
J_p(w) - X^\top\alpha \nonumber \\
J_p(w) &= X^\top\alpha \nonumber\\
w &= J_q(X^\top\alpha) \label{tensor_sub_w}
\end{align}
and
\begin{align}
\frac{\partial L(w,r,\alpha)}{\partial r} &=
\gamma(r-y) + \alpha \nonumber\\
&= \gamma r - \gamma y + \alpha \nonumber\\
\gamma r &= \gamma y - \alpha \nonumber\\
r &= y - \gamma^{-1}\alpha \label{tensor_sub_r}
\end{align}

operating substitutions \autoref{tensor_sub_w} and \autoref{tensor_sub_r} into \autoref{tensor_lag} we write its equivalent form w.r.t. $\alpha$ as
\begin{align}
\max_{\alpha\in\r^n} L(w(\alpha), r(\alpha), \alpha)
&= 	\frac{\gamma}{2}\|-\gamma^{-1}\alpha\|^2 
	+ \frac{1}{p}\|J_q(X^\top\alpha)\|^p_p
	+ \alpha^\top\left(y-\gamma^{-1}\alpha-XJ_q(X^\top\alpha) \right) \nonumber\\
&=	\frac{\gamma}{2}\|\frac{\alpha}{\gamma}\|^2
	- \frac{1}{\gamma}\|\alpha\|^2
	+ \alpha^\top y
	+ \frac{1}{p}\|J_q(X^\top\alpha)\|^p_p
	- (X^\top\alpha)^\top(J_q(X^\top\alpha)) \nonumber\\
&=	- \frac{1}{2\gamma}\|\alpha\|^2
	+ \alpha^\top y
	+ \frac{1}{p}\|J_q(X^\top\alpha)\|^p_p
	- (X^\top\alpha)^\top(J_q(X^\top\alpha))\label{tensor_lag_semi}
\end{align}

Now we are going to prove that 
\begin{equation}\label{tensor_sub_jq}
(X^\top\alpha)^\top(J_q(X^\top\alpha))
- \frac{1}{p}\|J_q(X^\top\alpha)\|^p_p
= \frac{1}{q}\|X^\top\alpha\|^q_q 
\end{equation}
Note that we changed the sign for clarity on later usage, since we are going to consider the dual problem.
We start by considering the substitution $u = X^\top\alpha$ with $u\in\r^n$ and re-write \autoref{tensor_sub_jq} more clearly as
\begin{equation}
u^\top J_q(u)
- \frac{1}{p}\|J_q(u)\|^p_p
\end{equation}

then we make use of the definition of $J_q$ to write
\begin{align}
u^\top J_q(u) - \frac{1}{p}\|J_q(u)\|^p_p
&=
\sum_iu_i\ sign(u_i)|u_i|^{q-1} 
- \frac{1}{p} \sum_i |u_i|^{(q-1)p} \nonumber\\
&=
\sum_i |u_i|\ |u_i|^{q-1}
- \frac{1}{p} \sum_i |u_i|^q \label{uu2}\\
&=
\sum_i |u_i|^q 
- \frac{1}{p}\sum_i |u_i|^q \nonumber\\
&= 
\frac{1}{q}\sum_i |u_i|^q \label{uu4}\\
&= 
\frac{1}{q}	\|u\|^q_q \label{uu5}
\end{align}
where in \autoref{uu2} we notice that $u_i\ sign(u_i) = |u_i|$ and by recalling that $(q-1) = \frac{q}{p}$ we know $(q-1)p = q$. To obtain \autoref{uu4} we simply make use of the fact that $1-\frac{1}{p} = \frac{1}{q}$.\bigbreak

Now that we proved \autoref{tensor_sub_jq}, we can write \autoref{tensor_lag_semi} as
\begin{equation}
\max_{\alpha\in\r^n}
{L(w(\alpha), r(\alpha), \alpha)
= 	- \frac{1}{2\gamma}\|\alpha\|^2
	+ \alpha^\top y
	-\frac{1}{q}\|X^\top\alpha\|^q_q}
\end{equation}

which is equivalent to the minimization problem, changed sign (the dual problem)
\begin{equation}\label{dual_tensor}
\min_{\alpha\in\r^n}
{ L(w(\alpha), r(\alpha), \alpha) = \frac{1}{q} \|X^\top\alpha\|^q_q + \frac{1}{2\gamma} \|\alpha\|^2_2
- \ang{y}{\alpha} }
\end{equation}

In analogy with the procedure in \autoref{kernel_ridge} we continue by calculating the optimality conditions for the primal problem first
\begin{align}
\frac{\partial}{\partial w}\left( \frac{\gamma}{2}\|Xw-y\|^2 + 1/p\|w\|^p_p\right)
&= \frac{\partial}{\partial w} \left(\frac{\gamma}{2}\|Xw-y\|^2\right) + J_p(w) \label{jp} \\
&= 2\frac{\gamma}{2}(Xw-y) \frac{\partial}{\partial w}(Xw-y) + J_p(w) \label{jp_2}\\
&= \gamma X^T(Xw-y) + J_p(w) \label{jp_3} \\
&= X^T(Xw-y) + \gamma^{-1}J_p(w) \label{jp_4}
\end{align}
where in \autoref{jp} we make use of
$J_p:\r^d\rightarrow\r^d$, which is the gradient of $\frac{1}{p}\|\cdot\|^p_p$ (it is a duality map), while in \autoref{jp_2} we apply the chain rule on $\|Xw-y\|^2$ and further solve it in \autoref{jp_3}. Note that in \autoref{jp_4} the term $(Xw-y)$ has size $(n\times 1)$, so $X^T(Xw-y)$ has size $(d\times 1)$, thus the operators $-$ and $+$ are to be considered entry-wise.\bigbreak

Similarly the dual problem optimality conditions are calculated as
\begin{align}
\frac{\partial}{\partial \alpha} L(w(\alpha),r(\alpha),\alpha) &=
\frac{\partial}{\partial \alpha} \left(\frac{1}{q} \|X^\top\alpha\|^q_q + \frac{1}{2\gamma} \|\alpha\|^2_2
- \ang{y}{\alpha}\right) \nonumber\\
&= \frac{\partial}{\partial\alpha}     \left(\frac{1}{q}\|X^\top\alpha\|^q_q\right)
+ \frac{2}{2\gamma}\alpha - y  \nonumber\\
&= XJ_q(X^\top\alpha) + \gamma^{-1}\alpha - y \label{jq}
\end{align}
where in \autoref{jq} we make use of $J_q:\r^d\rightarrow\r^d$ as the derivative of $\frac{1}{q}\|\cdot\|^q_q$ and also apply the chain rule, multiplying the term $\partial (X^\top\alpha) / \partial \alpha = X$ to $J_q(X^\top\alpha)$.\bigbreak

Now, if we multiply by $X^\top$ the equation in \autoref{jq} and take into account that $J_p \odot J_q = Id$, it follows that 
\begin{equation}
w = J_q \left( X^\top\alpha \right)
\end{equation}

And in this case the representer theorem becomes 
\begin{equation}\label{tensor_representer}
w = J_q \left( X^\top\alpha \right) = J_q\left(\sum^n_{i=1} \alpha_ix_i \right)
\end{equation}

Note that mapping $J_q$ is the derivative of $1/q|\cdot|^q$, meaning that it is defined as $sign(\cdot)|\cdot|^{q-1}$, applied component-wise. This makes \autoref{tensor_representer} nonlinear in the $\alpha_i$'s. \bigbreak

In order to retrieve a proper kernel representation to solve this problem, as proved in \citep{salzo2016}, we have to make the assumption that $q$ is an even integer and $q\ge 2$. From which follows that 
\begin{equation}
\forall u \in \r^d,\quad J_q(u) = \left( sign(u_j)|u_j|^{q-1} \right) = \left(u_j^{q-1}\right),\quad j=\{1, ..., d\}
\end{equation}
Therefore, as we proceeded for kernel ridge regression, we now define the formula to compute the estimate for a given new point $x$ (that is $\ang{w}{x}$), and we do so by means of a kernel representation, just as in \autoref{ridge_kernel_test}.\bigbreak

So we use the representer theorem in \autoref{tensor_representer}, as a substitute to $w$, and write down the estimation for a new point $x$ as
\begin{equation}\label{tensor_1}
\ang{w}{x} = \sum^d_{j=1}\left(\sum^n_{i=1} \alpha_ix_{i,j}\right)^{q-1} x_j = 
\sum^d_{j=1} \sum^n_{i_1,...i_{q-1}=1} x_{i_1,j}\cdots x_{i_{q-1},j}x_j\alpha_{i_1}\cdots \alpha_{i_{q-1}}
\end{equation}
where we expanded the power of the summation in a multilinear form since $q$ is an integer. \bigbreak

We define the \textit{linear tensor kernel function k} as
\begin{equation}
k : \r^d\times \dots \times \r^d \rightarrow \r,\quad
k(x'_1, \cdots, x'_q) = \sum^d_{j=1} x'_{1,j}\cdots x'_{q,j} 
= \text{sum}(x'_1 \odot \cdots \odot x'_q)
\end{equation}

we apply the above defined kernel function to \autoref{tensor_1} obtaining
\begin{equation}
\ang{w}{x} = \sum^n_{i_1,...i_{q-1}=1} k(x_{i_1}, \cdots , x_{i_{q-1}}, x)\alpha_{i_1}\cdots\alpha_{i_{q-1}}
\end{equation}
thus successfully integrating a kernel function into the problem in hands. This formulation allows us to compute the estimate for a new point by taking into account the solution $\alpha$ of the dual problem and the kernel function only. Moreover, we can write the dual problem in \autoref{dual_tensor} w.r.t. the kernel function in order to completely substitute $k$ to the training points. The new dual problem becomes
\begin{equation}\label{dual_tensor_kernel}
\min_{\alpha\in\r^n}{
\sum^n_{i_1, ..., i_q=1}
\frac{1}{q} k(x_{i_1},\cdots ,x_{i_q}) \alpha_{i_1}\cdots\alpha_{i_q} +
\frac{1}{2\gamma}\|\alpha\|^2_2 -
\ang{y}{\alpha}
}
\end{equation}

Note that the first term is convex since it is equal to $\frac{1}{q}\|X^\top\alpha\|^q_q$, which is convex.
We are dealing with a convex polynomial optimization problem of degree $q$. As we said for the kernel ridge regression case, we can extend the method to general feature maps. In this case, we can define a feature map $\Phi: \r^d \rightarrow \ell^q(\mathbb{K})$, $\Phi(x) = (\phi_k(x))_{k\in\mathbb{K}}$, with $\mathbb{K}$ a countable set. By making use of such feature map, we can consider general \textit{tensor kernels} defined as

\begin{equation}
k(x'_i,\cdots x'_q) = \sum_{k\in\mathbb{K}} \phi_k(x'_1) \cdots \phi_k(x'_q) 
= \text{sum}\left(\Phi(x'_1)\odot \cdots \odot \Phi(x'_q)\right)
\end{equation}

Tensor kernels are symmetric:
\begin{equation}
\forall x_1, ..., x_q \in \r^d,\ and\ every\ permutation\ \sigma\ of\ \{1, ..., q\},\ 
k(x_{\sigma(1)}, ..., x_{\sigma(q)}) = k(x_1, ..., x_q)
\end{equation}
and positive definite:
\begin{equation}
\forall x_1, ..., x_q \in \r^d,\ and\ every\ \alpha\in\r^n,\ 
\sum^n_{i_1, ..., i_q=1} k(x_{i_1}, ..., x_{i_{q}})
\alpha_{i_1} \cdots \alpha_{i_{q}} \ge 0
\end{equation}
It was proved in \citep{salzo2016} that tensor kernels have an associated \textit{reproducing kernel Banach space}. \\
Using a general tensor kernel results in the following representation for the testing phase
\begin{equation}
\ang{w}{\Phi(x)} = \sum^n_{i_1, ..., i_q=1} 
k(x_{i_1}, \cdots, x_{i_{q-1}}, x)\alpha_{i_1}, \cdots, \alpha_{i_q-1}
\end{equation}

Note that it is not needed to know the feature map $\Phi$ explicitly to use a generalized tensor kernel. In particular, we introduce 2 tensor kernels in which this is true:\bigbreak

\textbf{Polynomial tensor kernel of degree $s\in\mathbb{N}, s\ge1$}\\
Defined as
\begin{equation}\label{poly_tensor}
k(x'_1, ..., x'_q) = \left(\sum_{j=1}^d x'_{1,j}, \cdots, x'_{q,j} \right)^s = \left( \text{sum}(x'_1 \odot \cdots \odot x'_q) \right)^s
\end{equation}
describing the space of homogeneous polynomials in $d$ real variables of degree $s$.\bigbreak

\textbf{Exponential tensor kernel}\\
Defined as
\begin{equation}
k(x'_1, ..., x'_q) = \prod^d_{j=1} e^{x'_{1,j}, ..., x'_{q,j}} 
= e^{\text{sum}(x'_1\odot\cdots\odot x'_q)}
\end{equation}
which is an example of an infinite dimensional model.\bigbreak

As we mentioned before, this model is applicable not only with a parametric schema but also to infinite-dimensional models (non-parametric). In particular we can generalize \autoref{tensor_primal} so to include $w\in\ell^p(\mathbb{N})$ a sequence in \lp space (see \defref{lp_space}) indexed by $\mathbb{N}$. The objective function translates into

\begin{equation}\label{tensor_infinite_1}
\min_{w\in\ell^p(\mathbb{N})}
{\gamma\sum^n_{i=1} \left( y_i - \ang{\Phi(x_i)}{w} \right)^2
+ \frac{1}{p}\|w\|^p_p}
\end{equation}

with $p=q/(q-1)$ and $q>2$ even integer, regularization parameter $\gamma>0$, feature map $\Phi : \mathcal{X}\rightarrow \ell^q(\mathbb{N})$. This problem can be solved through the dual function in \autoref{dual_tensor_kernel}, which is expressed in terms of the tensor kernel and so can be solved even though the primal is infinite-dimensional (\autoref{tensor_infinite_1} in this case). This point is particularly important since it allows to make use of non-parametric models.\\

We can further generalize the problem to general loss functions, ending up with 
\begin{equation}\label{tensor_infinite_2}
\min_{w\in\ell^p(\mathbb{N})}
{\gamma\sum^n_{i=1} L(y_i, \ang{\Phi(x_i)}{w}) + \frac{1}{p}\|w\|^p_p}
\end{equation}

where $p>1$, $\gamma>0$, the feature map is defined as $\Phi : \mathcal{X}\rightarrow \ell^q(\mathbb{N})$, and $L:\mathcal{Y}\times \r \rightarrow\r$ is a loss function convex in the second variable. In order to explicitly write the dual function for \autoref{tensor_infinite_2} we should first define the \textit{linear feature operator}

\begin{equation}
\Phi_n: \ell^p(\mathbb{N})\rightarrow \r^n,\quad \Phi_nw 
= \left(\ang{\Phi(x_i)}{w} \right),\quad for\ i=\{1, ..., n\}
\end{equation}

and its adjoint
\begin{equation}
\Phi_n^{*}:\r^n\rightarrow\ell^q(\mathbb{N}),\quad \Phi_n^*\alpha 
= \sum^n_{i=1} \alpha_i\Phi(x_i)
\end{equation}

so now we can write down the dual representation for \autoref{tensor_infinite_2}, as proved in Theorem 3.1 of \citep{salzo2017}:
\begin{equation}
\min_{\alpha\in\r^n}
{\frac{1}{q}\|\Phi_n^*\alpha\|^q_q
+ \gamma\sum^n_{i=1}L^*\left(y_i, -\frac{\alpha_i}{\gamma}\right)}
\end{equation}
where $L^*(y_i, \cdot)$ is the \textit{Fenchel conjugate of} $L(y_i, \cdot)$. Furthermore, the theorem states that the primal problem $F$ has a unique solution, the dual problem $\Lambda$ has solutions and $\min\ F = -\min\ \Lambda$ (strong duality).

\newpage

\newpage
\newpage
\section{\secfive}\label{sec_five}
In this section we are going to discuss the improvement brought to this field by proposing a new layout for storing the Tensor Kernel. We begin by first showing how it was implemented, starting by the $2$-dimensional case, switching to the $3$-dimensional one, empirically proving the reduction made possible, so then extending it to a higher dimension. Afterwards we show experiments on memory usage (in \autoref{memory_gain}), on execution times both on real world dataset (in \autoref{sec_wpbc}) and on other datasets (in \autoref{sec_dexter} and \autoref{memory_synthetic}). In \autoref{sec_wpbc} we briefly summarise the method we are comparing to.

\bigbreak
As we said in the previous chapter, we need to build a tensor in order to store the entries of the kernel function. In this case, given the fact that the kernel function has more than 2 entries, we cannot make use of the Gram matrix, as it was the case for classical kernel functions. So a higher-order structure is needed to store the information, that is a tensor. \bigbreak

The problem with tensors is that they require a huge amount of memory. Think for example, that storing the fourth order tensor kernel for $n=100$ points, requires $100^4 = 100$ million numbers in memory. \\
This aspect has restricted the usage of tensor kernels to few points. \bigbreak

Kernel functions have the property of being symmetric and tensor kernels preserve that property. Symmetry for higher order structures is defined as
\begin{equation}
\forall x_1, ..., x_q \in \r^d,\ and\ every\ permutation\ \sigma\ of\ \{1, ..., q\},\ 
k(x_{\sigma(1)}, ..., x_{\sigma(q)}) = k(x_1, ..., x_q)
\end{equation}
for a $q$-order tensor kernel. In our experiments we dealt with the case of $q=4$, that is $4^{th}$ order tensor kernels. So, given 4 points $x_1, x_2, x_3, x_4$ the corresponding kernel value for each possible permutation of these 4 points, are equal. Meaning that $k(x_1, x_2, x_3, x_4) = k(x_2, x_1, x_3, x_4) = ...$ and so on.\\
This piece of information makes it clear that most of the entries of the tensor contain repeated values. Hence, we can avoid storing the same value for all the possible permutations for each set of points. We should find a way to retrieve the corresponding tensor kernel value, given a set of points, no matter the permutation.\bigbreak

This is done by simply storing all the non-repeated elements linearly (in an array) and retrieving a formula that takes in input the 4 indices corresponding to the 4 points given in input and returns the corresponding entry of the array (regardless of the permutation). For example, given points $x_1, x_2, x_{23}, x_{42}$, we should retrieve some formula that receives in input the indices $(1, 2, 23, 42)$ and returns the corresponding entry of the array of tensor kernel values corresponding to $k(x_1, x_2, x_{23}, x_{42})$. Note that this formula returns the same value regardless of the order in which indices are given.\\

\begin{figure}[!ht]
\centering
\includegraphics[width=0.95\textwidth]{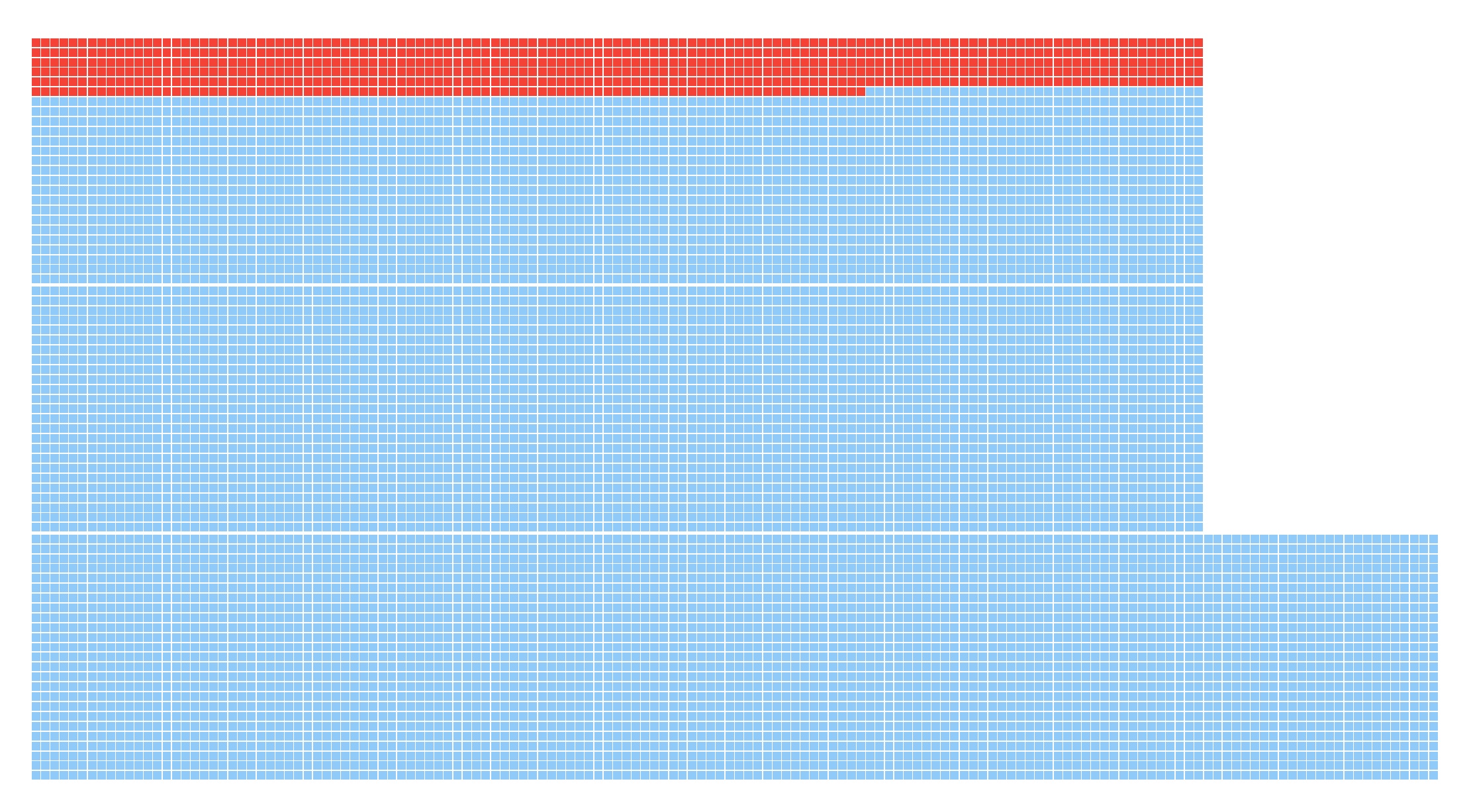}
\caption{Number of elements stored in a tensor kernel of $4^{th}$ order, for $n=10$ points. In red it is shown the effective non-repeated elements. Namely the tensor has 10.000 entries, while there are only 715 non-repeated ones (denoted in red).}
\label{memory_n10}
\end{figure}

To give a graphical insight of the amount of memory saved by making use of this trick, look at \autoref{memory_n10}.

\newpage
In order to give an idea on how to solve this problem, start by considering the $2$-dimensional case, where we deal with symmetric matrices.

\begin{figure}[!ht]
\centering
\includegraphics[width=0.4\textwidth]{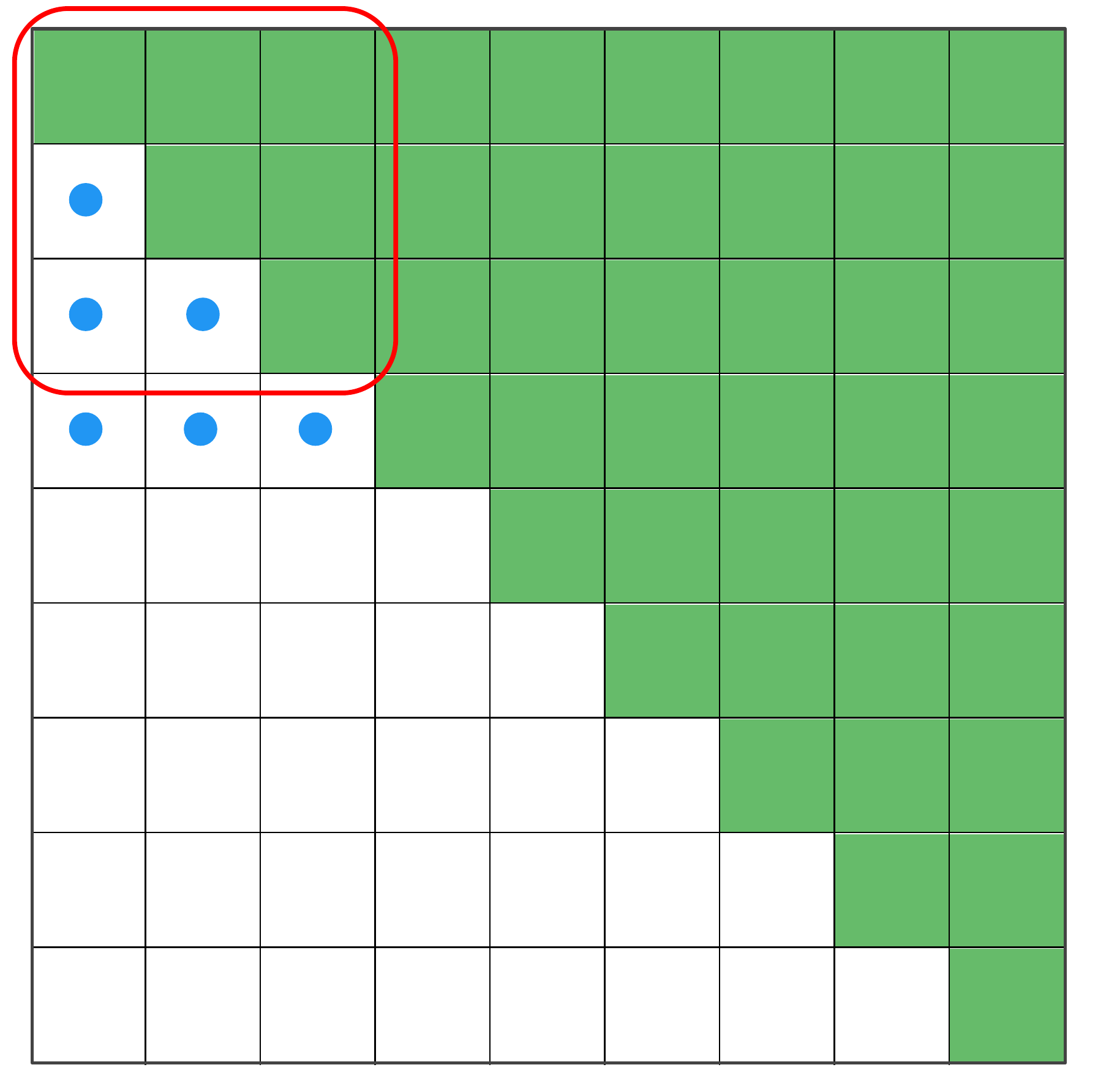}
\caption{A symmetric matrix where the upper triangular part is coloured in green, denoting the unique elements. In red the submatrix S ($3\times 3$) needed for \autoref{skip_2d}. In blue points that have to be skipped in \autoref{index_2d}.}
\label{matrix_sym}
\end{figure}

Given a matrix $A = (a_{ij}) \in \r^{n\times n}$, the symmetry property translates into $a_{ij} = a_{ji}$, which means that the upper-triangular part of the matrix contains all the useful elements, while the rest of the matrix has repeated elements only. Graphically shown in \autoref{matrix_sym}. 

We would like to store only the elements showed in green. So we construct a vector big enough to store all of them, in particular we have 
\begin{equation}
b_{matrix} = \frac{n*(n+1)}{2}
\end{equation}
non-repeated (unique) elements (those displayed in green)(here we use letter $b$ since all other common letters are already in use). This would be the case for the Gram matrix of well-known two entry kernel functions, applied on $n$ points. As we said, we store the green entries only, in a vector, preserving the order present in \autoref{matrix_sym}, where we consider first the columns (from left to right) and then rows (from top to bottom).\bigbreak

In this case, to access a particular element, given the two indices $(r, c)$ of the matrix entry we are interested in, we compute the corresponding entry in the vectorial storage of the matrix, with the following rule:
\begin{equation}\label{index_2d}
v_{2d}(r, c) = (r*n) - skip_{2d}(r) +c
\end{equation}
where $(r*n)+c$ accounts for the position as if the whole matrix was stored into a vector, but we have to calculate also the amount of repeated points we did not store, hence we detract $skip_{2d}(r)$ which is defined as
\begin{equation}\label{skip_2d}
skip_{2d}(r) = \frac{r*(r+1)}{2}
\end{equation}

To clarify this calculations, take in example the case in which we want to retrieve position $(r=3, c=3)$ of the $(9\times9)$ matrix of \autoref{matrix_sym} (indexing starting from 0). Formula $(r*n)+c$ gives as a result $30$ (skip 3 full rows, take fourth element of fourth row), as if the entire matrix was stored. This is not our case, so we have to remove those points incorrectly considered, that are points denoted in blue in \autoref{matrix_sym}. To calculate the number of such points we can consider the submatrix S $(3\times 3)$ and its number of upper-triangular entries ($skip_{2d}(r=3)$), which is exactly the amount we want to skip.
\bigbreak

\begin{figure}[!ht]
\centering
\includegraphics[width=0.9\textwidth]{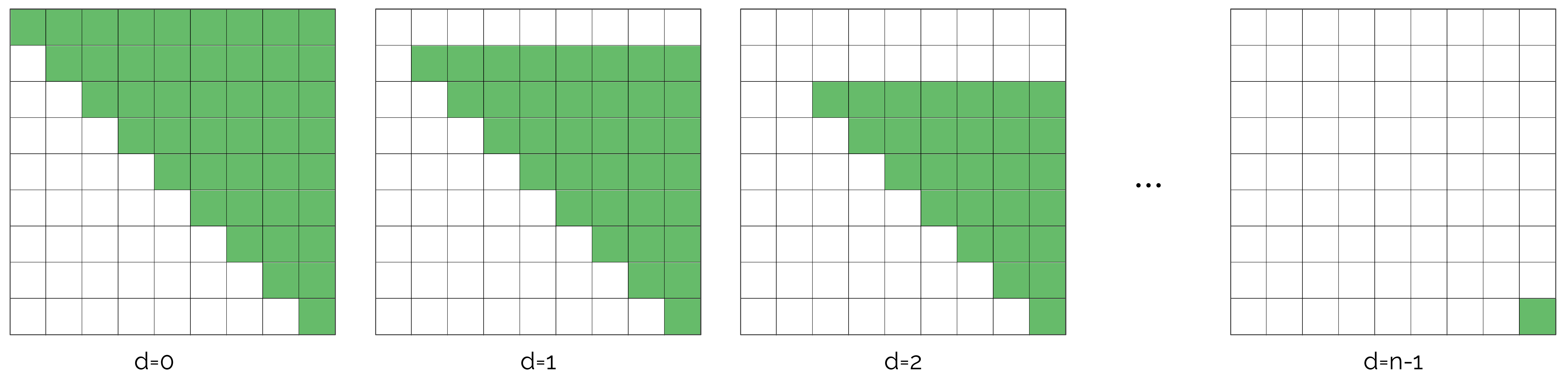}
\caption{Symmetric cube unrolled on the third dimension ($d$). Unique entries are denoted in green, all the others are repeated (because of symmetry). }
\label{cube_sym}
\end{figure}

This same procedure can be extended to the $3$-dimensional case, in which the amount of repeated elements grows, so the actual number of unique entries in a symmetric cube becomes
\begin{equation}
b_{cube} = \frac{n*(n+1)*(n+2)}{6}
\end{equation}

to visually perceive the growth in repeated elements in the cubic case, look at \autoref{cube_sym}. For graphical clarity the cube has been unrolled in the third dimension $d$ (written under each layer).

As you can note, the last layer of the cube ($d=n-1$) has only one unique element and $n^2-1$ repeated entries. In the first layer we need approximately half of the elements, while for each successive layer we need increasingly less points, making it particularly wasteful to store all of these elements.
To give a numeric perception look at \autoref{t_size}.

\begin{table}[!ht]
\centering
\begin{tabular}{|l||c|c|c||c|c|c||c|c|c|}
\hline
$n$
& $n^2$ & $b_{mat}$ & $\%_{reduce}$ 
& $n^3$ & $b_{cube}$ & $\%_{reduce}$ 
& $n^4$ & $b_{tensor4}$ & $\%_{reduce}$ \\ \hline
10 & 100&55&45.00\% & 1000&220&78.00\% & 10000&715&92.85\% \\ \hline
20 & 400&210&47.50\% & 8000&1540&80.75\% & 160000&8855&94.47\% \\ \hline
30 & 900&465&48.33\% & 27000&4960&81.63\% & 810000&40920&94.95\% \\ \hline
40 & 1600&820&48.75\% & 64000&11480&82.06\% & 2.56 M&123410&95.18\% \\ \hline
50 & 2500&1275&49.00\% & 125000&22100&82.32\% & 6.25 M&292825&95.31\% \\ \hline
100 & 10000&5050&49.50\% & 1 M&171700&82.83\% & 100 M&4421275&95.58\% \\ \hline
\end{tabular}
\caption{For each number of points $n$, this table shows the number of entries for a full matrix($n^2$), full cube($n^3$) and a full fourth order tensor($n^4$). Next to each of these is showed the actual number of non-repeated entries for a symmetric matrix($b_{matrix}$), symmetric cube($b_{cube}$) and a symmetric tensor($b_{tensor}$). Next to each of these is showed the percentage of repeated entries. }
\label{t_size}
\end{table}

As you can see in the $2$-dimensional case (first block of 3 columns ($n^2$, $b_{mat}$, $\%_{reduce}$), storing a matrix requires approximately half of the space if the symmetry property is present (look at column $\%_{reduce}$, which expresses the proportion of repeated elements w.r.t. the entire structure). Switching to the second block of 3 columns, we have the case for $3$-dimensional structures (cubes), where we expect the number of repeated elements to be more than half, as we noticed in \autoref{cube_sym}. Indeed here $\%_{reduce}$ is approximately equal to $81\%$, meaning that we can store only $19\%$ of the entire cube and the rest can be retrieved thanks to the symmetry, thus achieving an $81\%$ of reduction in memory. Below is explained how.\bigbreak

Following the same reasoning of the matrix case, we can build up a vectorial layout to store only the unique entries. 
In order to retrieve the corresponding vectorial index, given cubic indices $(r,c,d)$ we have to compute
\begin{equation}\label{v3d}
v_{3d}(r,c,d) = (d*n^2) - skip_{3d}(d) + v_{2d}(r,c)
\end{equation}
where
\begin{equation}\label{skip_3d}
skip_{3d}(d) = n^2 - (n-d)^2 + skip_{2d}(n-d-1) + skip_{3d}(d-1)
\end{equation}
The reasoning is similar to that of $v_{2d}$. Here we first compute the third dimension (layer $d$) as if we were considering the entire cube ($d*n^2$), then we account for those entries which are not stored since they are repeated elements ($skip_{3d}(d)$) and finally, once we are in the proper layer $d$ (which can be seen as a matrix), we simply move to the corresponding entry with $v_{2d}(r,c)$. \\

$skip_{3d}$ has been written in a recursive way for simplicity. Note that since we are dealing with relatively small $n$ (in the order of hundreds) and it has only 1 input variable, we can think of storing a table which entries are $skip_{3d}(i)$ for all $i\in\{0, ..., n-1\}$. This would require $\mathcal{O}(n)$ memory and can be computed only once in the beginning of the execution, and later accessed to avoid the computation of this recursion, making the index retrieval operation ($v_{3d}(r,c,d)$) executable in constant time $\mathcal{O}(1)$. Note that we can further improve this step by considering the whole recursion process in inverted order, that is start from $d=0$, where there is no recursion because of the base case, and continue with $d+1$ in which we can avoid calling the recursion since we already have the corresponding entry of the skip table for $d=0$, hence making it executable in constant time. Go all the way up until $skip_{3d}(d=n-1)$ which is going to need $skip_{3d}(d=n-2)$ which has already been computed and inserted into the skip table. In other words, filling the skip table containing all entries of $skip_{3d}(i)$ requires only $\mathcal{O}(n)$ time.\bigbreak

Going up in the number of dimensions leads to no difference, in the sense that the overall procedure is going to be similar. So now we consider the case of a $4$-dimensional structure. Higher dimensional structures can be referred to as \textit{tensors}, while in our particular case of an $(n\times n\times n\times n)$ structure it is called a \textit{tesseract} ($4$-dimensional analogue of a cube). The word \textit{tesseract} was coined and first used in 1888 by \textit{Charles Howard Hinton} in his book \textit{A new era of thought} \citep{hinton1888}. It comes from Greek words for "four rays", referring to the four lines outgoing from each vertex. \\
Given a tensor $K$ of $q^{th}$-order, symmetry is defined as
\begin{equation}
K_{\sigma(1), ..., \sigma(q)} = K_{1, ..., q}
\ for\ every\ permutation\ \sigma\ of\ \{1, ..., q\}.
\end{equation}

If we take in example the $4^{th}$ order tensor, the number of non-repeated elements is given by
\begin{equation}
b_{tensor4} = \frac{n*(n+1)*(n+2)*(n+3)}{24}
\end{equation}
although this number is still in the order of $\mathcal{O}(n^4)$, it provides a considerable amount of reduction in space. Looking at \autoref{t_size} you can notice in the last column that the amount of repeated elements composes approximately $95\%$ of the tensor. Graphically visualized in \autoref{memory_n10} also, achieving such reduction would allow for processing higher numbers on applications where a tensor is required. More information on memory requirement on \autoref{memory_gain}. We keep the $4^{th}$ order example from now on. \bigbreak

Just as before, the idea is to store the data in a linear fashion (in a vector), in a particular order, and have a function that transforms the tensorial indices into the corresponding vectorial index where the desired data entry resides. So we store the data by preserving the order we used so far, that is we consider them first by column (from left to right), then we consider rows (top-down), then the third dimension (here we referred with $d$, so from $d=0$ to $d=n-1$), so the fourth dimension $t$, again from $0$ to $n-1$.\\
One simple way to consider this order is by means of 4 nested loops constructed as follows:

\begin{lstlisting}
 for(r=0 to n-1):
 	for(c=r to n-1):
 		for(d=c to n-1):
 			for(t=d to n-1):
 				...
\end{lstlisting}

This simple way of looping indices $(r,c,d,t)$ effectively considers only the $b_{tensor4}$ non-repeated elements of the tensor.\\
Note that for higher order tensors, the nested loops structure may be replaced by a recursive call for each dimension, carrying out a list of indices in the end of the recursion tree, where each index is greater or equal than the previous one. \bigbreak

Similarly, we write down the formula that converts indices $(r,c,d,t)$ into its corresponding vectorial entry index:
\begin{equation}\label{v4d}
v_{4d}(r,c,d,t) = (t*n^4) - skip_{4d}(t) + v_{3d}(r,c,d)
\end{equation}
where we preserved the same exact structure as in \autoref{v3d}, with the only difference of considering one more order of dimensions. Here the function $skip_{4d}(t)$ still takes a single input and is defined as
\begin{equation}
skip_{4d}(t) = (t*n^2) + skip_{3d}\rule{1ex}{.4pt}inv(t)
+ skip_{4d}(t-1)
\end{equation}
where in this case we have to consider the inverted function for $skip_{3d}$ instead, which is defined as follows

\begin{equation}
skip_{3d}\rule{1ex}{.4pt}inv(d) = n^2 - (n-d)^2 + skip_{2d}(n-d-1) + skip_{3d}\rule{1ex}{.4pt}inv(d+1)
\end{equation}
with the base case being met for $(d=n)$, in which case the returned value is $0$. Note that the only difference from \autoref{skip_3d} stands in the recursive call. Note also that the call to $skip_{2d}$ is executed in constant time $\mathcal{O}(1)$.\\ 
The same reasoning we brought out previously for $skip_{3d}$ also holds in this inverted case, that is we can store a table containing all values of $skip_{3d}\rule{1ex}{.4pt}inv(i)$ for all $i\in\{0, ..., n-1\}$. This computation may be dealt with by an inverted fashion, as explained before (to be clear, in this case would be the inversion of the inversion). Having such table would result in function $skip_{3d}\rule{1ex}{.4pt}inv(d)$ being executed in constant time $\mathcal{O}(1)$. Again we point out that $n$ is a really small number, thus storing a table of size $\mathcal{O}(n)$ is irrelevant with respect to the memory we are using for the tensor.\\

We can similarly build up a table holding values of $skip_{4d}(i)$ for all $i\in\{0, ..., n-1\}$ by making the same considerations already considered twice. We build it in an inverted order, so to avoid useless recursive calls. \\
Few things to keep in mind: First, after building the table for $skip_{3d}\rule{1ex}{.4pt}inv$ and finish using it for the other tables, there is no need to keep it in memory, since we are going to use only the table for $skip_{3d}$ for further references. Secondly, after building the above 2 skip tables, the call to $v_{4d}$ becomes executable is constant time $\mathcal{O}(1)$ since in \autoref{v4d} $skip_{4d}(t)$ becomes a table reference and the call to $v_{3d}(r,c,d)$ is executed in $\mathcal{O}(1)$ as well. That is because unrolling $v_{2d}$ call yields 
\begin{equation}
v_{2d}(r,c) = (r*n)-\frac{r*(r+1)}{2}+c
\end{equation}
hence the call to $v_{3d}$ becomes 
\begin{equation}
v_{3d}(r,c,d) = (d*n^2)-table\rule{1ex}{.4pt}skip_{3d}[d] + (r*n)-\frac{r*(r+1)}{2}+c
\end{equation}
which is executed in constant time $\mathcal{O}(1)$.\bigbreak

In our experiments, the access to the tensor kernel values never actually makes use of $v_{4d}$ since it simply reads all of the tensor entries in a linear fashion. This was made possible by introducing some little programming tricks. That is to say that the computation is further improved by the fact that we actually store tensor entries linearly.\bigbreak

All the experiments to follow are executed on the local machine with the following configuration:
\begin{itemize}
\item CPU Intel Core i5-5300U 2x CPU 2.3-2.9 GHz (2 logical cores per physical)
\subitem L1d cache: 32K, L1i cache: 32K
\subitem L2 cache: 256K
\subitem L3 cache: 3072K 
\item RAM 8 GB DDR3L (Hynix HMT41GS6BFR8A-PB)
\item SSD 256 GB Opal2
\item Ubuntu 16.04 LTS
\end{itemize}

Code can be found at \url{https://github.com/feliksh/TensorKernel}.

\newpage
\subsection{Memory Gain}\label{memory_gain}
Working with $4^{th}$ order tensors is memory expensive. To get an idea about it, think that for $n=200$ you need an $1600M$ entry tensor. Fortunately we can reduce this amount of memory by $95\%$ down to approximately $68M$.\\
In terms of effective memory, in our experiments we worked with \textit{double} types in a \textit{C++} environment. In most architectures, 8 bytes are used to store a double type. So to compute the quantity of bytes required to store a $4^{th}$ order tensor made of double entries
\begin{equation}\label{old_mem}
mem(n) = n^4 * 8 B
\end{equation}
gives us the result. In order to convert it into GigaBytes we have to divide the result of \autoref{old_mem} by $2^{30}$.\bigbreak

Instead, if we use the proposed layout to store the symmetric tensor, the following 
\begin{equation}
new\rule{1ex}{.4pt}mem(n) = \frac{n(n+1)(n+2)(n+3)}{24} * 8B
\end{equation}
gives us the number of bytes required to store the structure.\\

\begin{figure}[!ht]
\centering
\includegraphics[width=0.95\textwidth]{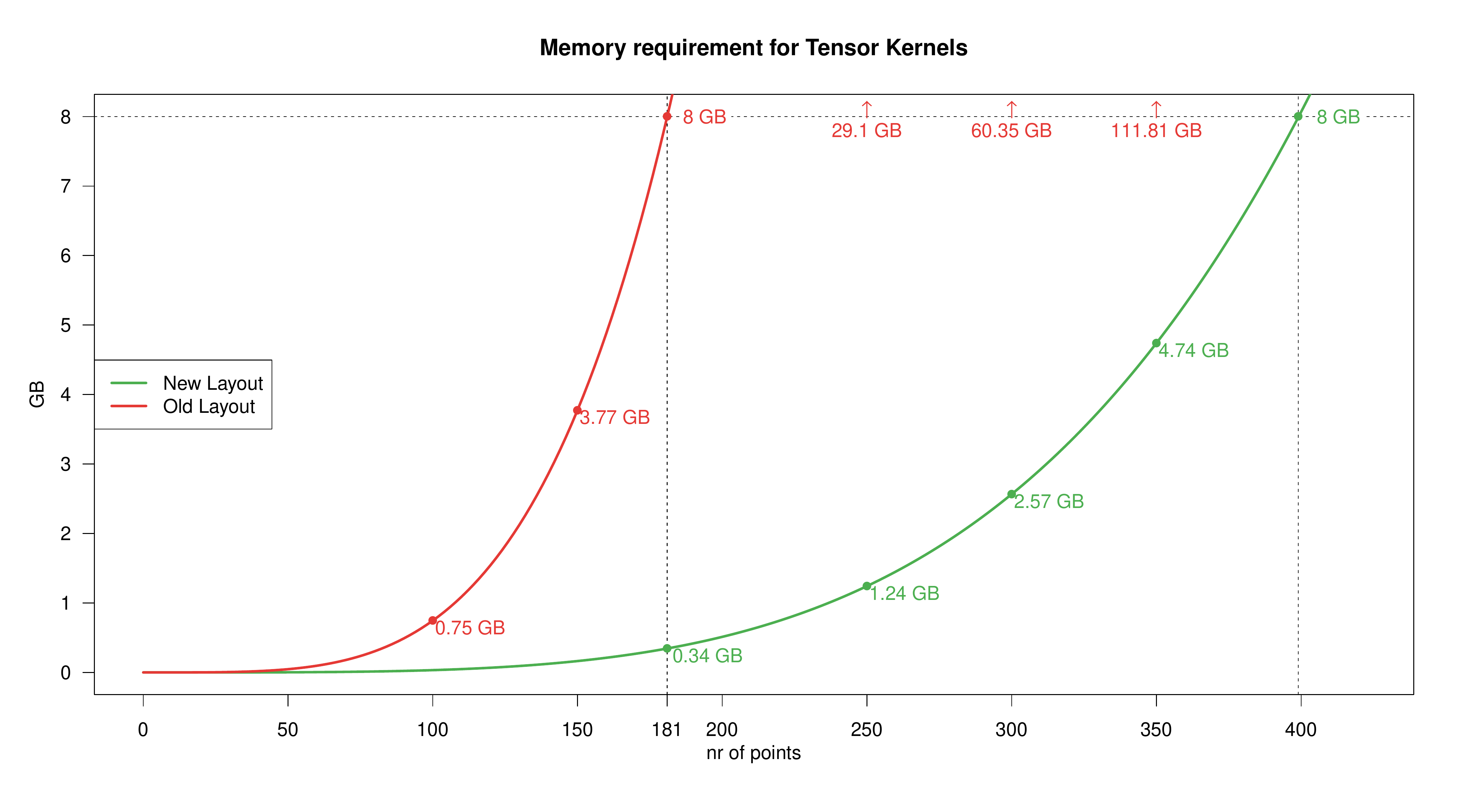}
\caption{Memory comparison between naive and proposed layout to store $4^{th}$ order symmetric tensor. Entries are assumed to be 8 Bytes double type.}
\label{memory_plot}
\end{figure}

Putting this two quantities in comparison, we have the situation depicted in \autoref{memory_plot}.

As you can see the memory difference is considerable. In a machine having a typical RAM of 8 GB of size, if we could completely fill it up with nothing else than the tensor (although we know it is not possible), with the old layout only 181 points would fit in, compared to the 399 points permitted by the new layout. \\
As you can notice, the old layout becomes particularly expensive in terms of memory starting from small values, \textit{e.g.} for $n=130$ it requires $2.13$ GB, adding $20$ more points would rise it up by $1.64$ GB (requiring $3.77$ GB for $n=150$). \\
On the other side, the new layout is less restrictive, \textit{e.g.} for $n=130$ it requires only $90$ MB (compared to the $2.13$ GB), while for $n=150$ only $163$ MB (compared to the $3.77$ GB). Thanks to the proposed layout, a higher number of points can be used \textit{e.g.} for $n=250$ we need $1.24$ GB of space (compared to $29.1$ GB with the old layout), reaching the limit on the example of the RAM when $n=399$, which requires $7.99$ GB instead of the $188.8$ GB of the old layout, resulting in a substantial difference of more than $180$ GB.\bigbreak

Bear in mind that this difference affects also the number of points we have to consider in order to compute the tensor kernel, thus the execution time. This aspect is part of the following experiments.

\subsection{Comparing on Wpbc dataset}\label{sec_wpbc}
For this analysis a real world dataset has been used. We considered the Breast Cancer Wisconsin (Diagnostic) dataset. Information on this dataset can be found from \url{http://pages.cs.wisc.edu/~olvi/uwmp/cancer.html}, you can retrieve it inside that page, in particular in \url{ftp://ftp.cs.wisc.edu/math-prog/cpo-dataset/machine-learn/cancer/WPBC/}. It is present also in UCI repository at \url{ftp://ftp.cs.wisc.edu/math-prog/cpo-dataset/machine-learn/cancer/WPBC/}.\\

This dataset is composed of 194 examples (4 were removed since labels are missing) with 32 variables. The goal is to predict whether a case is benign or malignant, so it is a classification problem but it is easy to cast it into a regression framework for our algorithm. \bigbreak

The dataset was divided into a training set of 60 points, a validation set of 60 points and the remaining 74 as the test set.\\
\begin{figure}[!ht]
\centering
\includegraphics[width=0.7\textwidth]{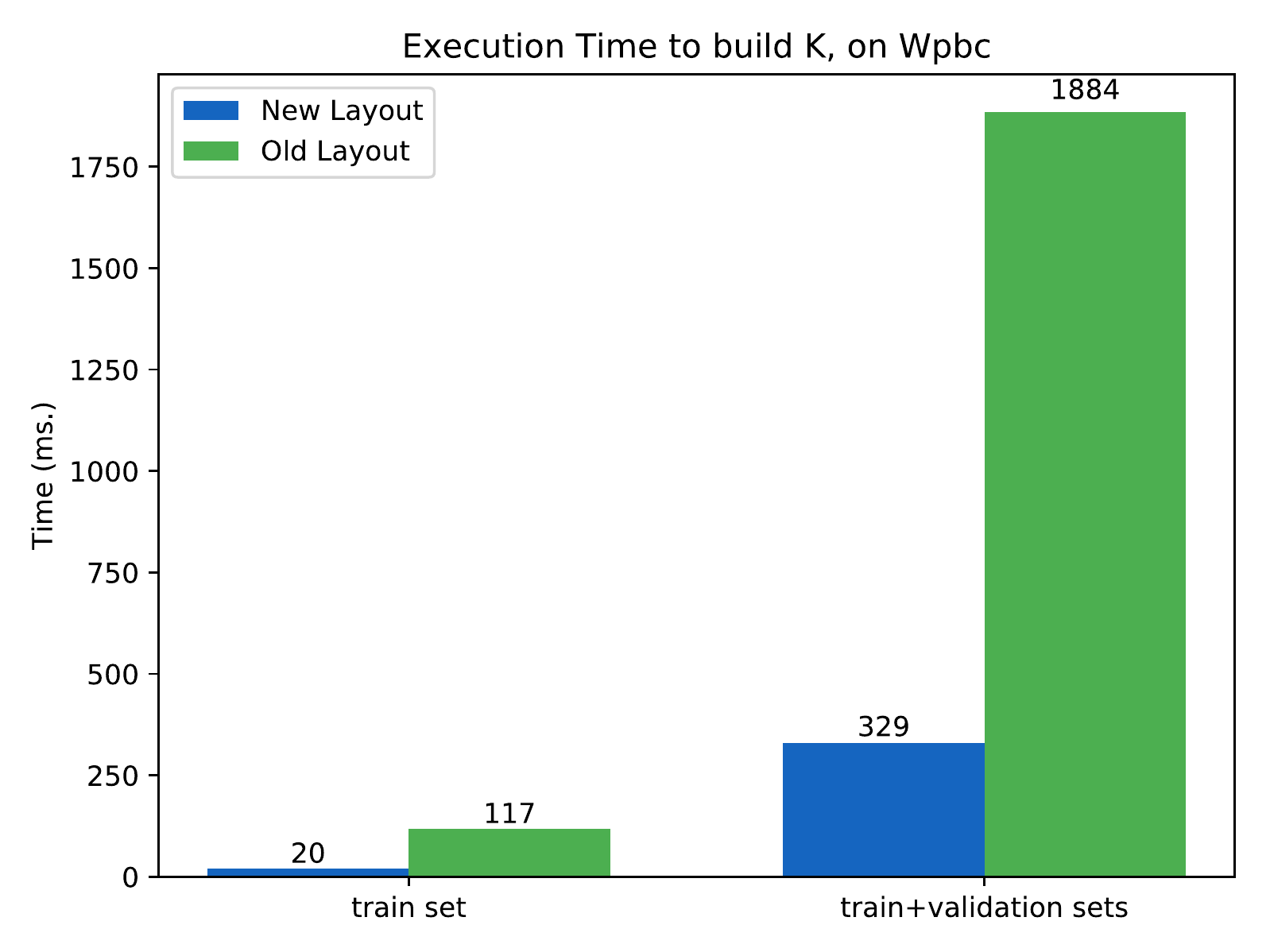}
\caption{Running time on Wpbc dataset, to build the tensor kernel $K$. On the left considering only the training set ($n=60$), on the right considering both training and validation sets ($n=60+60=120$).}
\label{compare_wpbc}
\end{figure}

The experiments carried out on this dataset showed the difference in execution time to compute the tensor kernel $K$. Results are graphically shown in \autoref{compare_wpbc}, where on the left we consider the tensor build on $n=60$ training set points, while on the right both training and validation sets are considered, $n=120$. \\

The difference present in \autoref{compare_wpbc} is due to the fact that by considering the new layout, we are drastically decreasing the number of entries of the tensor kernel to be computed. In particular, for $n=60$ we have to compute $n^4 = 12.96 M$ entries for the old layout, compared to the $595.665$ for the new layout. This difference leads to a decrease in the execution time. \\
The computation of the tensor kernel using the old layout is carried out in an optimized way, as described in \citep{salzo2017}. That is, they stored the tensor in a matrix fashion, since the $4^{th}$ order tensor can be viewed as an $n^2\times n^2$ matrix. The optimization procedure was modified accordingly, resulting in an optimized algorithm. In particular, the computation of the tensor matrix is efficient as well. It is computed as follows

\begin{lstlisting}
 Given data as a matrix `${n\times d}$` 
 Let x1 be a matrix `${n^2 \times d}$` 
 k=0
 `for`(i=0 to n-1)
 	`for`(j=i to n-1)
 		x1[k] = data[i]*data[j]
 		k = k+1
 matrix K = x1 * x1`$^T$`
\end{lstlisting}
as you can see from the code above, the computation time is mainly dominated by dot products and a transposition, which are operations particularly efficient and can be highly optimized by the compiler. This is the reason for which the difference in \autoref{compare_wpbc} is not as big as we would expect, given the difference in the number of entries. Even with this optimization, the algorithm based on the new layout outperforms the other anyway, resulting in a speedup of \colblue{5.85} for the training set only and \colblue{5.72} for both sets (speedup computed as $117ms./20ms.$ and $1884ms./329ms.$ respectively). \bigbreak

Note that the optimized code discussed above depends heavily on the size $d$ of data points we are considering. Indeed this difference is going to be noticeable in experiments in which $d>\!>n$ (which is the case for feature selection scenarios in general).

\subsection{Comparing on Dexter}\label{sec_dexter}
This experiment has been carried out in a dataset called Dexter. It was part of the 2003 NIPS feature selection challenge (more information in \citep{guyon2003} and at \url{http://clopinet.com/isabelle/Projects/NIPS2003/}).\bigbreak

This dataset is composed of 300 examples for the training set, 300 for the validation set and 2000 for the test set (which is not available since it was used only for the competition, so it was never published). It is a classification dataset and the classes are perfectly balanced (150-150 for training and 150-150 for validation). Each point lies in a $20000$-dimensional space with $9947$ variables being real and the remaining $10053$ being added randomly, thus not affecting the response. The dataset is sparse, in the sense that only $0.5\%$ of entries are non zero.\bigbreak

Since the test set is not available we decided to re-arrange the splits into 200 for training, 200 for validation and 200 for testing. Random sampling was carried out to pick such splits, even though initial datas already have random order.\bigbreak

In \autoref{compare_dexter} we report the execution time for the training set on Dexter dataset using the linear tensor kernel with $\gamma=0.12$, mean-std standardized datas.

\begin{table}[!ht]
\centering
\begin{tabular}{|l|c|c|}
\hline
Method & Time Build K & Time Optimization \\ \hline
New Layout & 1746 s. & 968 s. \\ \hline
Old Layout & - & -\\
\hline
\end{tabular}
\caption{Execution time on Dexter dataset, on 200 training examples. The execution using old layout could not finish since it causes an execution error related to memory. Linear tensor kernel with $\gamma=0.12$.}
\label{compare_dexter}
\end{table}

As you can see from the table, to create the tensor kernel we spend $1746 s.$ of execution time. This is due to the fact that such kernel has more than $68$ M entries and for each entry a pairwise multiplication between $4$ vectors of $20000$ dimensions is computed. The rest of the execution is due to the optimization algorithm and it requires $968 s.$ to execute.\bigbreak

On the other side, the execution using the old layout could not be reported since the experiments were carried out on $8$ GB of RAM, while the structure alone (the tensor matrix as explained in \autoref{sec_wpbc}) would require $11.92$ GB of space. So the process finished with exit code 9: OutOfMemory. Meaning that utilizing the old layout, it is not possible to handle such numbers on the current hardware setup.

\subsection{Comparing on synthetic data}\label{memory_synthetic}
The following experiment has been conducted on a synthetic dataset. As such we could manually choose $n,d$ and the desired sparsity $s$.\\

The dataset was constructed by first generating a matrix $X$ of size $(n\times d)$ with each entry picked from a standard normal distribution ($x_{i,j}\sim \mathcal{N}(0,1)$). A sparse vector $w_*$ was created such that all entries are equal to $0$ except for $s$ randomly picked ones. Those entries are first assigned a sign, which in our case we picked as the sign of a standard normal distribution draw ($sgn(h),\ h\sim\mathcal{N}(0,1)$) and  subsequently multiplied by $(1-0.3u)$, with $u\sim U(0,1)$ drawn according to a standard uniform distribution. A noise vector $\varepsilon$ was created simply as a vector in $\r^d$ where $\varepsilon_i \sim \mathcal{N}(0, 1)$.\\
The response variable is computed according to the formula
\begin{equation}
y = Xw + \sigma\varepsilon
\end{equation}
with $\sigma$ being the noise parameter, in our experiments set to $0.05$.\bigbreak

\begin{figure}[!ht]
\centering
\includegraphics[width=0.7\textwidth]{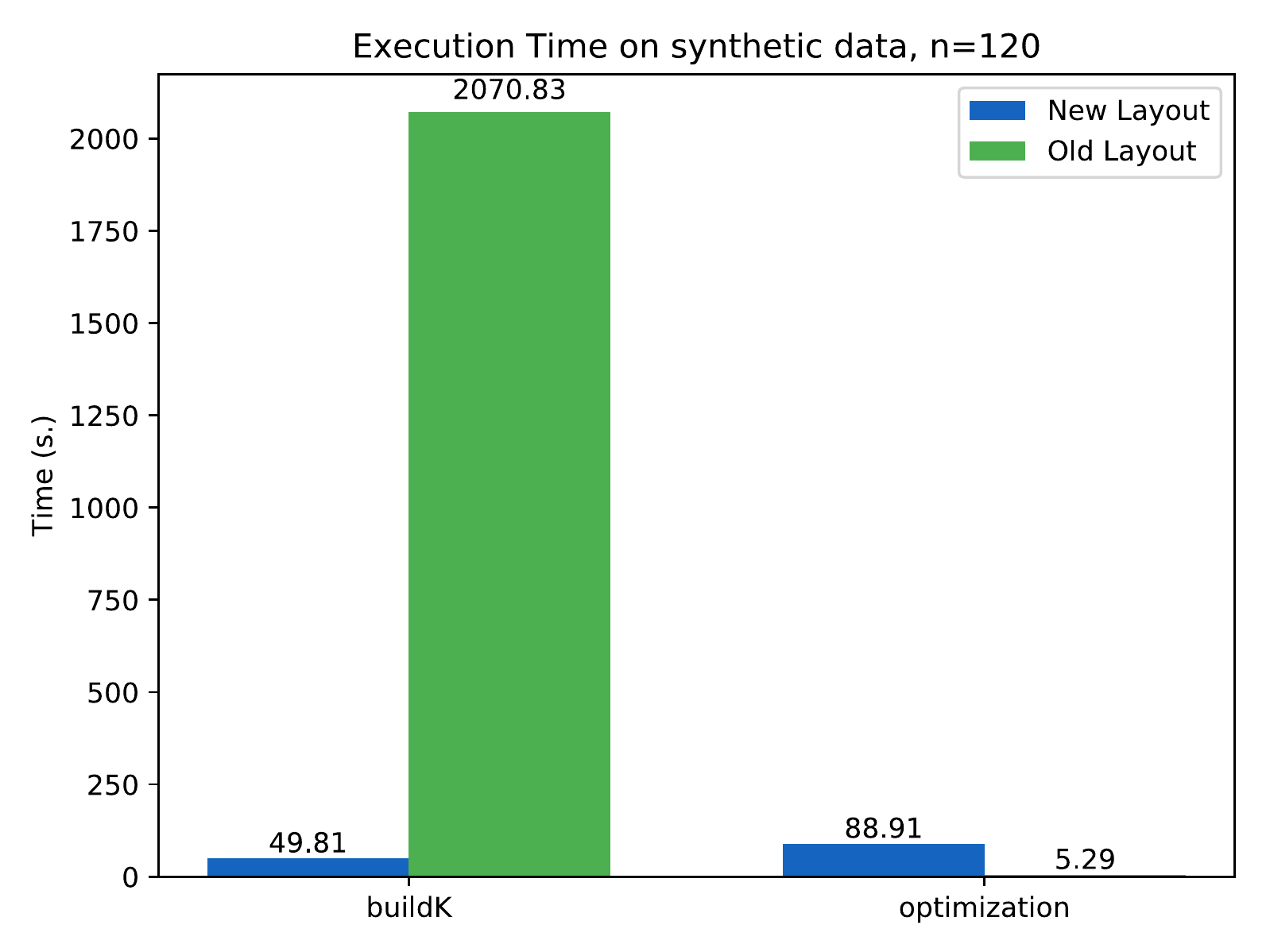}
\caption{Execution time for $n=120$ on synthetic data, with $d=5000,\ s=7,\ \gamma=0.86$ and 40 optimization iterations and linear tensor kernel.}
\label{compare_n120}
\end{figure}

We consider 2 experiments, the first one with $n=120$ points constructed with the formula explained above, where we used a dimensionality $d=5000$ with only $s=7$ relevant features.\\
By employing the linear tensor kernel, execution times are exposed in \autoref{compare_n120}.\bigbreak

There is a substantial difference in the times required to build the tensor kernel $K$ as you can see from the left part of the plot in \autoref{compare_n120}. In particular, by making use of the proposed layout we achieve a speedup of \colblue{41.57}. The reason for this difference, even by using the optimization for the old layout as discussed in \autoref{sec_wpbc}, is due to the fact that it is influenced by both $n$ and $d$. In this case we are considering $d>\!>n$, differently from the experiment in \autoref{sec_wpbc}, meaning that the algorithm used for the old layout has to compute inner products between bigger vectors, resulting in a performance decay.\bigbreak

On the other hand instead, using a matrix structure to store the tensor kernel considerably improves the execution time for the optimization algorithm. Indeed, as you can see from the right side of \autoref{compare_n120}, employing the old layout results in a faster optimization execution. In particular it results in a speedup of \colgreen{16.80}. This is given by the fact that the optimization algorithm has been modified accordingly in order to speed up the execution, which was the main purpose of the work done in \citep{salzo2017}. Such algorithm is mainly composed of matrix and vector multiplications, which are particularly easy for the compiler to optimize, thus providing a considerable improvement in time.\\
However, this improvement is not enough, compared to the time needed to build the tensor kernel. Indeed the overall times are $138.72 s.$ with the new layout compared to the $2076.12 s.$ required for using the old layout. So it results in an overall \colblue{14.96} of speedup by using the new layout.

Moreover, as already proved before, an important reason for preferring the proposed layout is that of providing the chance of working with higher numbers. Indeed the following experiment carried out for $n=250$ shows that the old layout cannot cope with it. The experiment depicted in \autoref{compare_n250} was performed with $d=2000,\ s=9\ \gamma=0.86$ and 40 optimization iterations.\\

\begin{table}[!ht]
\centering
\begin{tabular}{|l|c|c|}
\hline
Method & Time Build K & Time Optimization \\ \hline
New Layout & 955 s. & 2753 s. \\ \hline
Old Layout & - & -\\
\hline
\end{tabular}
\caption{Execution time for $n=250$ on synthetic data, with $d=2000,\ s=9,\ \gamma=0.86$, 40 optimization iterations and linear tensor kernel. }
\label{compare_n250}
\end{table}

Note that 40 optimization iterations are more than enough for a stable result for the optimization algorithm. So we can think of reducing such number in order to decrease the time required for optimization since it composes a big part of the overall time. It should result in no performance decay (in terms of feature selection) since in the first iterations usually relevant features are already discovered and successive iterations serve as weight stabilization. By exploiting an adaptive threshold on weights we can expect the same performance with a lower number of iterations.

\newpage
\newpage
\section{\secsix}\label{\secsix}
In this section we are going to introduce another method to deal with higher numbers, that is the \textit{Nystr\"om} like strategy. Next we are going to present experimental results
on effectively large numbers, otherwise intractable for the hardware in use (in \autoref{nystrom_synthetic}), evaluating both accuracy and feature selection.
After that we carry out qualitative experiments on the real world dataset (in \autoref{nystrom_wpbc}) mentioned before and on the synthetic one to further analyse extracted features (in \autoref{nystrom_manual}). We conclude by discussing a crossed experiment carried out by utilising both proposed improvements.\bigbreak

Inspired by the work of \citep{rudi2015} we decided to consider subsampling methods, also known as \nystrom approaches, for our experimentation. In \citep{rudi2015} they effectively managed to reduce the memory and time requirement by making use of this subsampling approach. In particular, they tested the interplay between the regularization and the subsampling parameter in order to achieve regularization and a better performance without loosing precision. More details to follow.\bigbreak

\subsection{Managing big numbers}\label{nystrom_synthetic}
This experiment was carried out on the synthetic dataset as described in \autoref{memory_synthetic}, but in this case considering a large case scenario in which there are $n=7000$ points on the dataset, divided into $4000$ examples for the training set, $1000$ for validation and $2000$ for testing. We considered $d=5000$ dimensional points with $s=17$ relevant features. \bigbreak

As mentioned before we experiment on the interplay between the regularization parameter $\gamma$ and the subsampling parameter $m$ which determines the number of points used for training the model. That is, once we choose a parameter $m$, we randomly sample 
\begin{equation}
\{\tilde{x}_1, \tilde{x}_2, \dotsc, \tilde{x}_m\}
\end{equation}
from the training set and use those points to train our model. We expect the results to be only slightly varying on small changes of the 2 considered parameters.\bigbreak

\begin{figure}[!ht]
\centering
\includegraphics[trim=0.5cm 0 3cm 0,clip,width=0.97\textwidth]{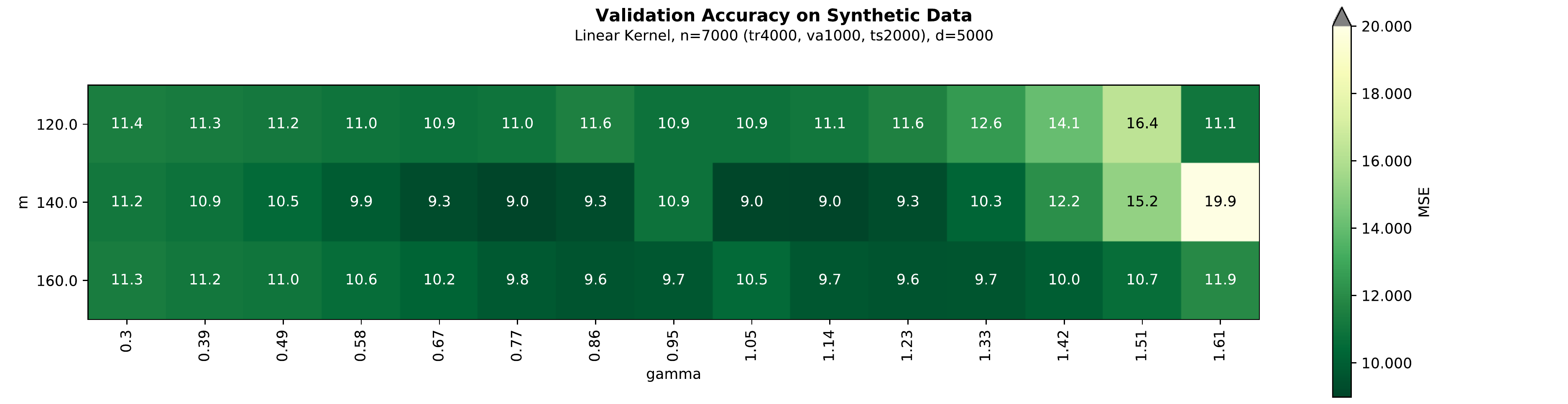}
\caption{Validation error (MSE) on synthetic data with $n=7000, d=5000, s=17$. Linear tensor kernel, with subsampling parameter $m$.}
\label{heat_synthetic}
\end{figure}

In \autoref{heat_synthetic} you can see the result of using the training set with the subsampling technique just discussed, on varying values of $m$ and $\gamma$. To measure the performance of the regression model we used the Mean Squared Error (MSE), defined as
\begin{equation}
MSE = \sum_{i=1}^{m} \frac{\left(f(x_i) - y_i \right)^2}{m}
\end{equation}

\begin{figure}[!ht]
\centering
\includegraphics[width=0.9\textwidth]{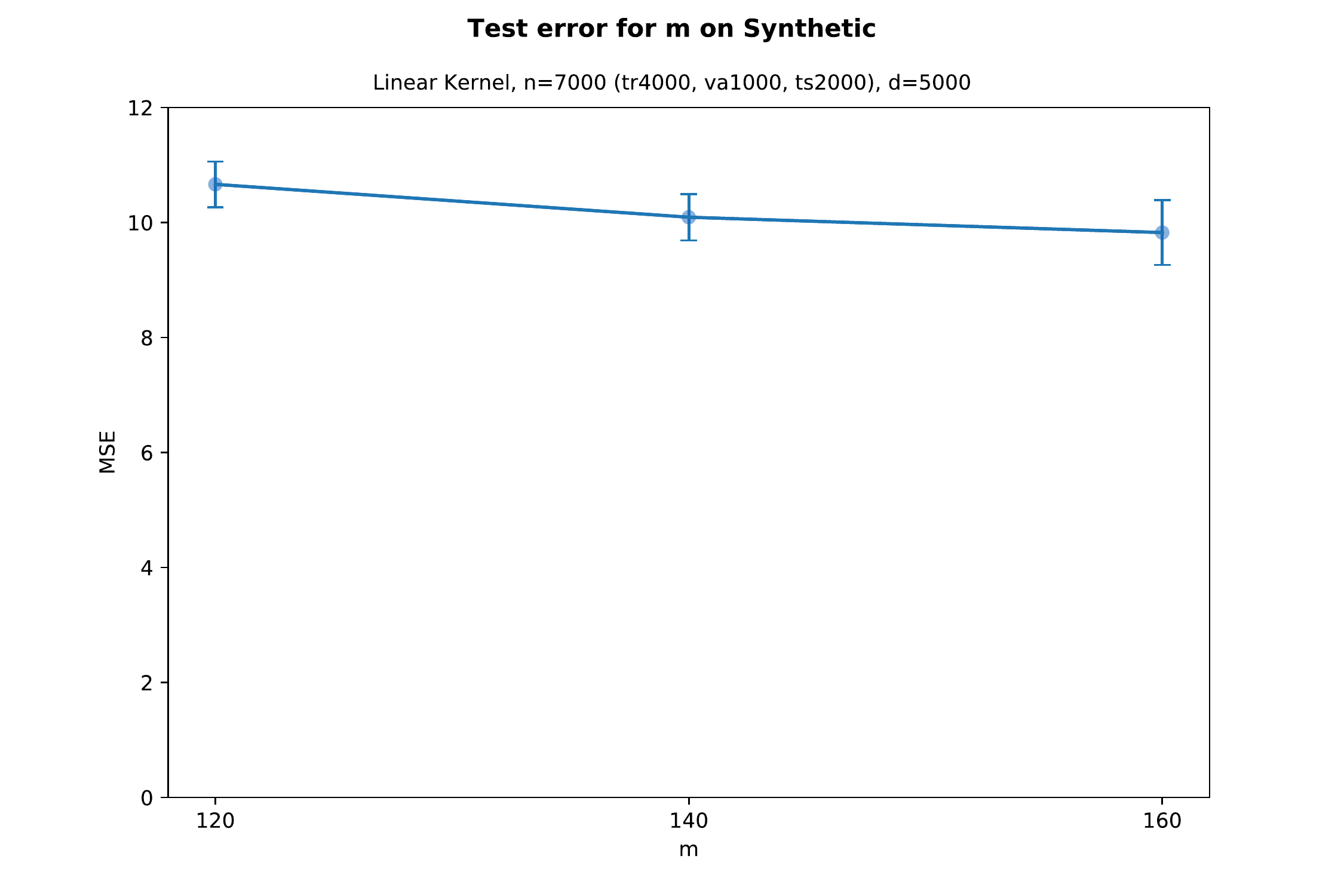}
\caption{Test error (MSE) on synthetic data with $n=7000, d=5000, s=17$. Linear tensor kernel, with subsampling parameter $m$.}
\label{test_synthetic}
\end{figure}

The procedure is defined as follows:\\
First we randomly divide the entire synthetic dataset into training-validation-test. Then we randomly sample $m$ points out of the training set and use them to train our model and evaluate it on the validation set. In \autoref{heat_synthetic} we report the values computed on the validation set in order to show the behaviour of the algorithm on varying values of the parameters. Then we should pick the best pair $(m,\gamma)$ based on results of the validation set and measure the performance on the test set. Instead, in order to give an experimental view on the behaviour of different values of $m$ we chose to show in \autoref{test_synthetic} the result on the test set for each choice of $m$ (by taking the best $\gamma$ for that particular $m$).\newpage

As you can see from the heatmap in \autoref{heat_synthetic} by varying the regularization parameter we achieve the expected behaviour of having a bad performance for small values of $\gamma$ (meaning high regularization since we consider $1/2\gamma$ in the dual problem \autoref{dual_tensor_kernel}) since it pushes learned weights to be extremely close to zero providing little estimation. On the other side, for growing values of $\gamma$ (meaning smaller regularization) it reaches a good balanced point (in \autoref{heat_synthetic} approximately in range $\gamma\in\{0.58, 1.33\}$), and successively reduces its performance for $\gamma > 1.50$ since it promotes overfitting situations which results in poor values of MSE on the validation set, as shown in the rightmost part of the heatmap.

From \autoref{test_synthetic} you can see the MSE measured on the test set. As you can notice the improvement for picking higher values of $m$ is relatively small (lower values of MSE are better). In order to have a more stable result, we carried out 10 evaluations on the test set. In particular, for each choice of $(m,\gamma)$ we sampled $m$ points from the training set, trained the model using $\gamma$ as a regularization parameter and then evaluated the performance on the test set. This process was repeated 10 times, picking 10 different samples. Overall, the reported variation is relatively small in all 3 considered cases, as you can notice in \autoref{test_synthetic}.\\

In what follows we show the performance of the task of predicting the relevant features for the 3 models trained on the test set.
Since we created the data, we know also the true sparse vector (the relevant features), so in the following figures we show the true weights in blue and the estimated ones in red.
 Parameters: $n=7000, d=5000, s=17,\ linear\ kernel$.

\begin{figure}[!ht]
\centering
\includegraphics[trim=1cm 0.5cm 1.5cm 1cm,clip,width=0.78\textwidth]{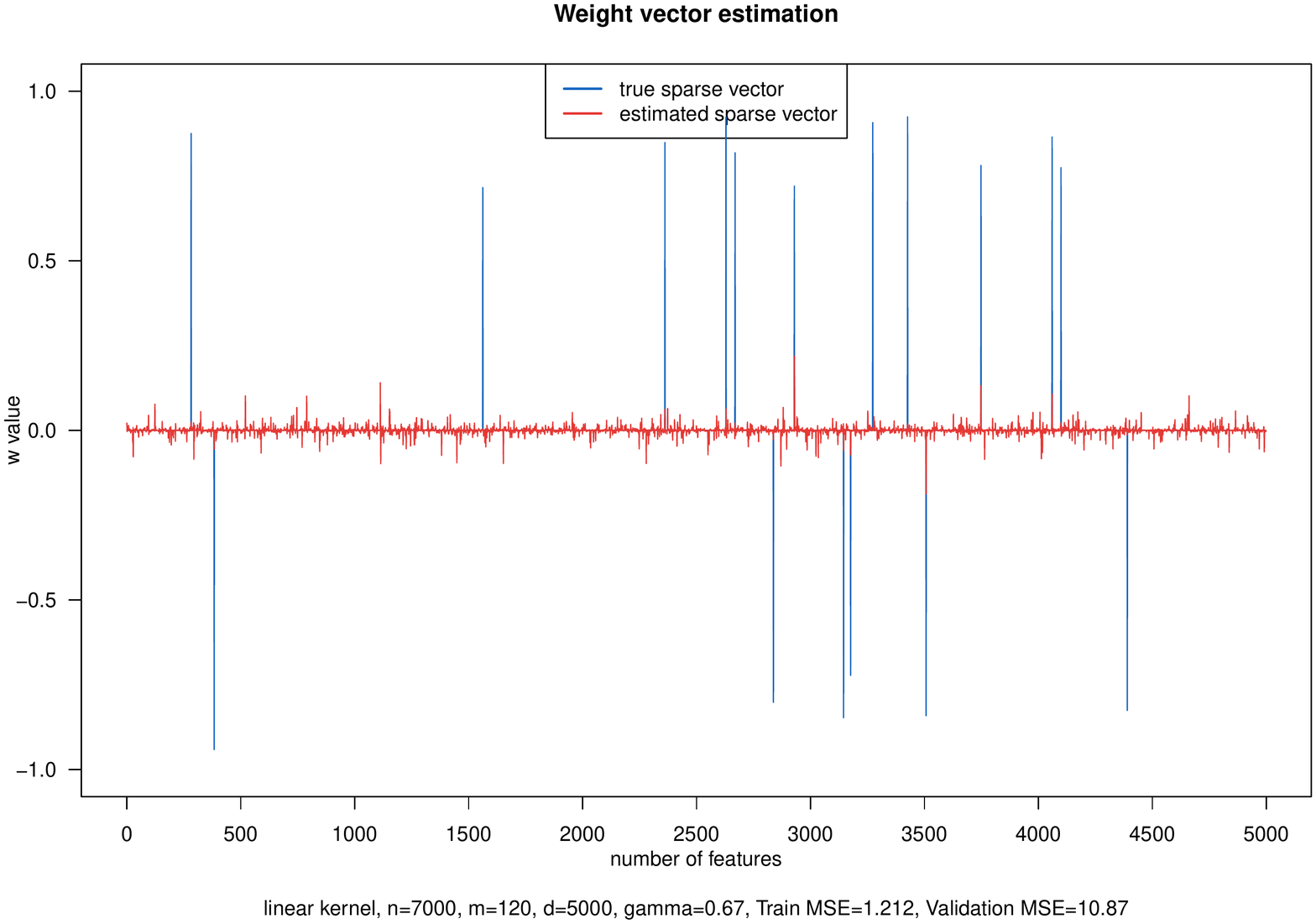}
\caption{Feature selection result for $m=120$ on the synthetic data experiment.}
\label{test_synthetic_m120}
\end{figure}

\begin{figure}[!ht]
\centering
\includegraphics[trim=1cm 0.5cm 1.5cm 1cm,clip,width=0.78\textwidth]{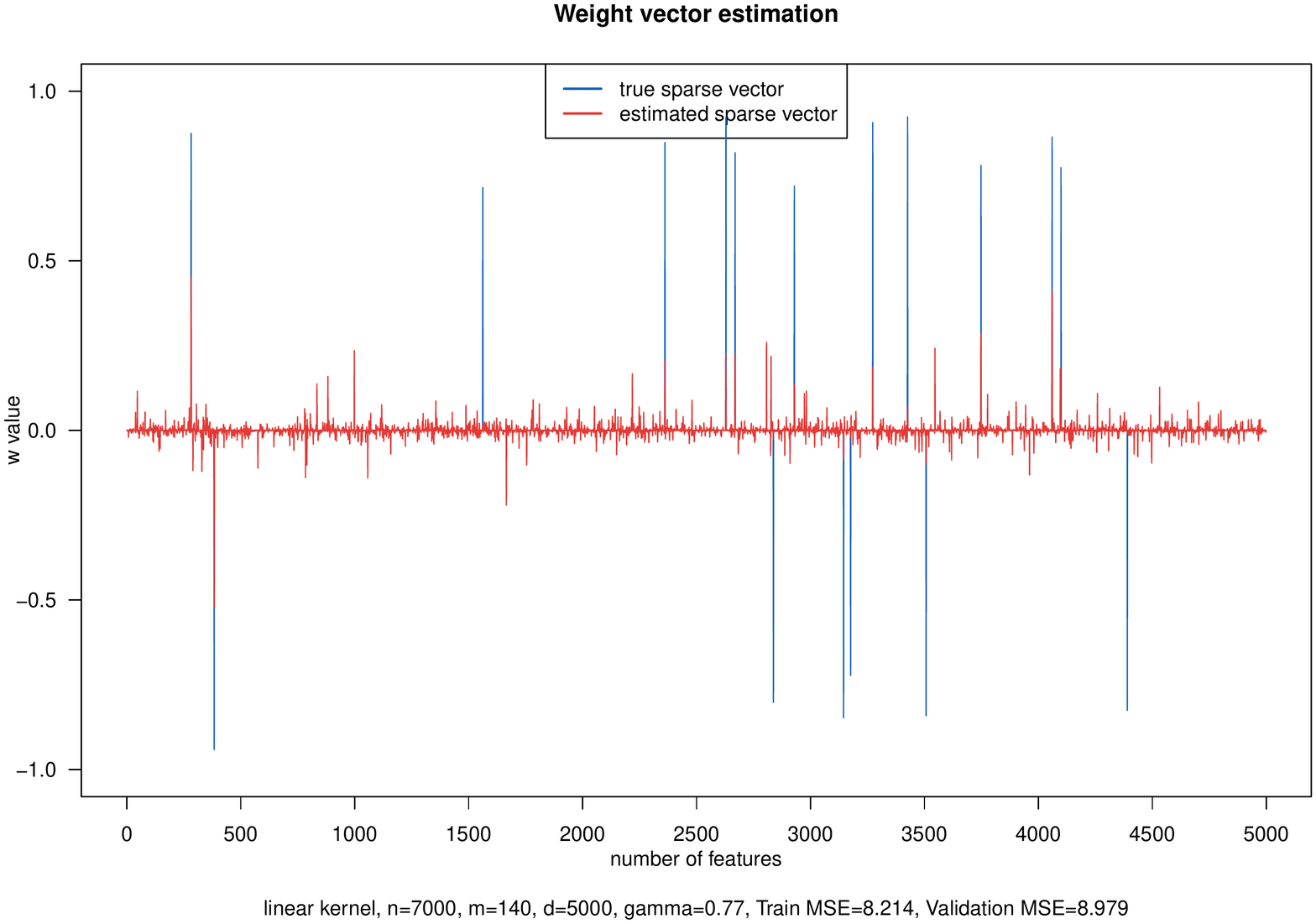}
\caption{Feature selection result for $m=140$ on the synthetic data experiment.}
\label{test_synthetic_m140}
\end{figure}

\begin{figure}[!ht]
\centering
\includegraphics[trim=1cm 0.5cm 1.5cm 1cm,clip,width=0.78\textwidth]{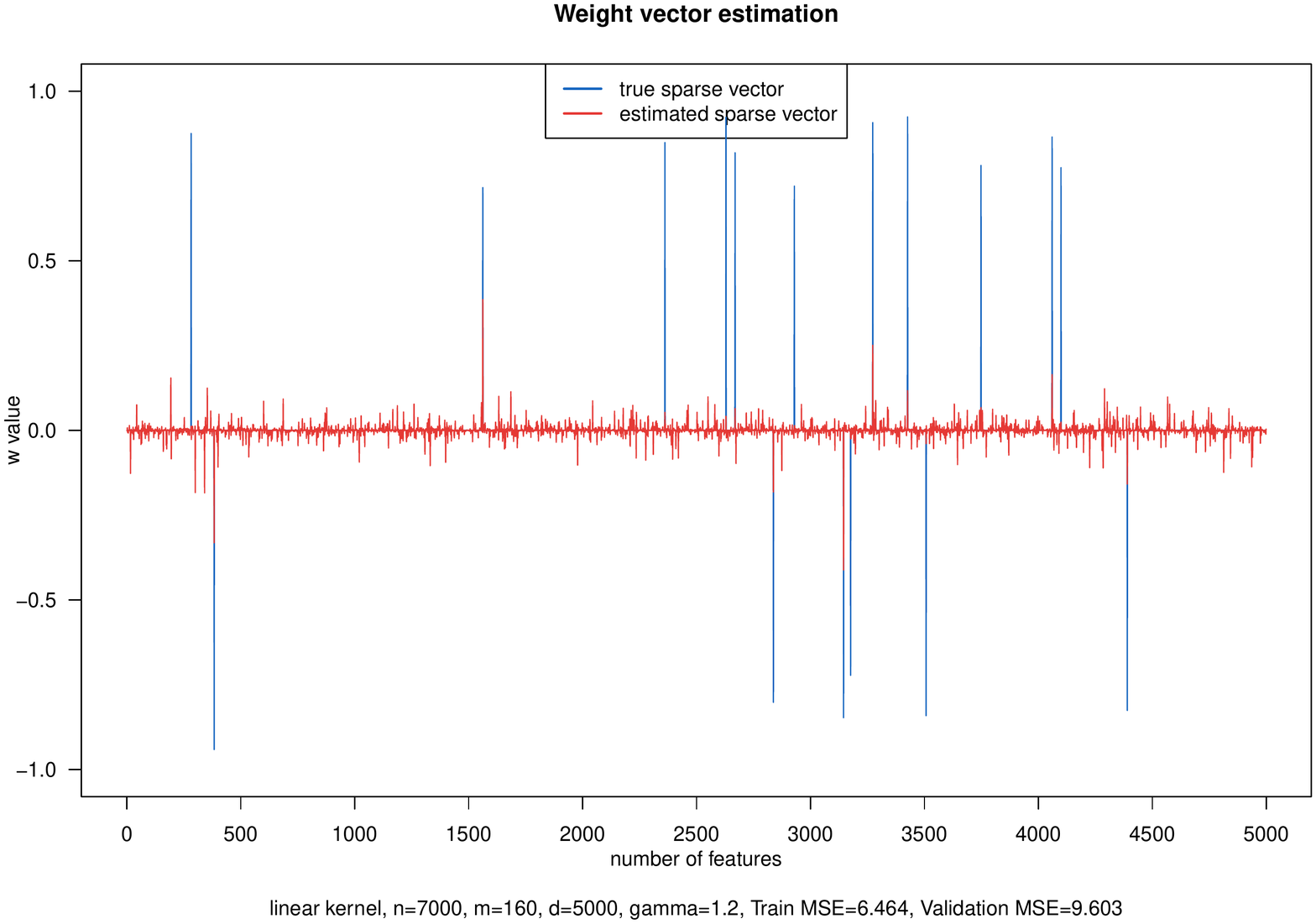}
\caption{Feature selection result for $m=160$ on the synthetic data experiment.}
\label{test_synthetic_m160}
\end{figure}

As you can see from \autoref{test_synthetic_m120} for $m=120$, \autoref{test_synthetic_m140} for $m=140$ and \autoref{test_synthetic_m160} for $m=160$, the overall prediction of the true features varies in the 3 cases. More or less all 3 models capture most of the $s=17$ relevant features, but some of them have more noise (on irrelevant features) and some have more accentuated correct weights. In general the prediction capability depends on the choice of $\gamma$. For $m=140$ and $m=160$ the situation seems to be visually similar, while $m=120$ seems to be the more distant case. That is because of the small variability of $m=120$ on the validation error, as noticeable in \autoref{heat_synthetic}, which caused the algorithm to pick as the best $\gamma$ a value which actually had a really small value of train MSE (as you can see from the subtitle in \autoref{test_synthetic_m120}), differently from the other 2 models. A more fine-grained search through the $\gamma$ values would have resulted in a better situation. Nevertheless, its result shows a good capability to retrieve the true weights, just as the other 2 models.

\newpage
\subsection{Results on Wpbc}\label{nystrom_wpbc}
\begin{figure}[!ht]
\centering
\includegraphics[trim=0cm 1cm 1cm 0,clip,width=0.99\textwidth]{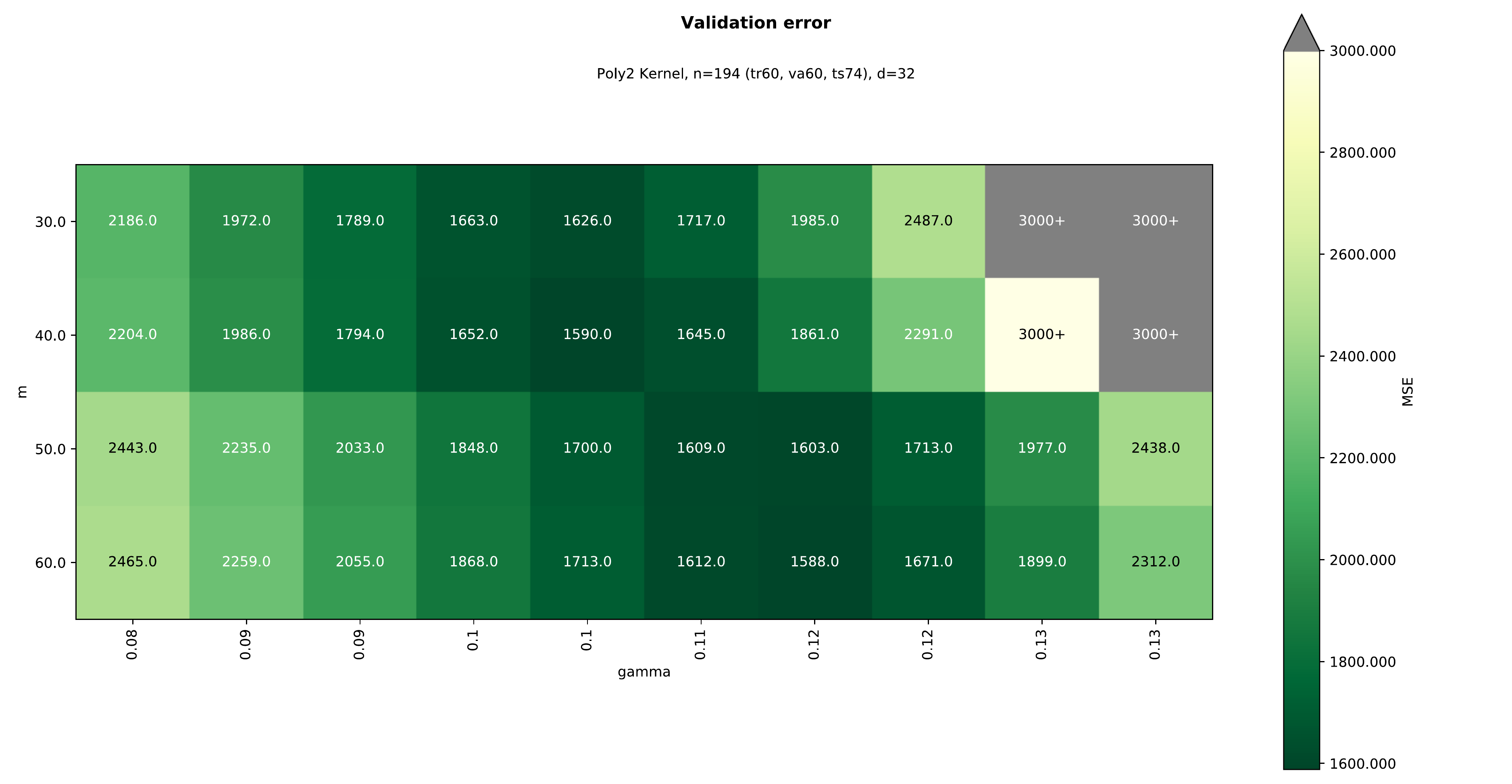}
\caption{Validation error (MSE) on Wpbc dataset, model trained on $m$ points sampled from training set ($60$ points) and evaluated on validation set ($60$ points).}
\label{heat_wpbc}
\end{figure}

This experiment has been carried out on the real world dataset on breast cancer, introduced in \autoref{sec_wpbc}. In this experiment we used the polynomial tensor kernel of degree 2, defined as in \autoref{poly_tensor}, and the results are presented in \autoref{heat_wpbc}.\\
Experimental setting remains the same as in the previous section, parameters $m$ and $\gamma$ are investigated through a gridsearch, points are sampled from the training set and validated on the validation set. Given the small training set (only $60$ points), there is not much room for $m$ variation, indeed the reduction is small, we considered $m=\{30, 40, 50, 60\}$, where $60$ corresponds to no subsampling.\bigbreak
The heatmap shows a relatively small change for varying $m$ values (interesting to note that for $m=40$ we achieved a better result than for $m=50$), confirming that subsampling approach seems a viable choice. On the other side, as expected, the variation on the regularization parameter brings a considerable change on both extremes, as already discussed. 
These results are in line with the ones presented in the supplementary work of \citep{salzo2017}, retrievable at \url{http://proceedings.mlr.press/v84/salzo18a/salzo18a-supp.pdf}, where they applied the tensor kernel approach on the same dataset. The results are in line with their outcome. \\
Looking again at \autoref{heat_wpbc} we note that the difference between the best runs on $m=40$ and $m=60$ is equal to only $2$, making it clear that switching to a subsample scenario does not involve a substantial loss on the accuracy of the model. 

\begin{figure}[!ht]
\centering
\includegraphics[width=0.99\textwidth]{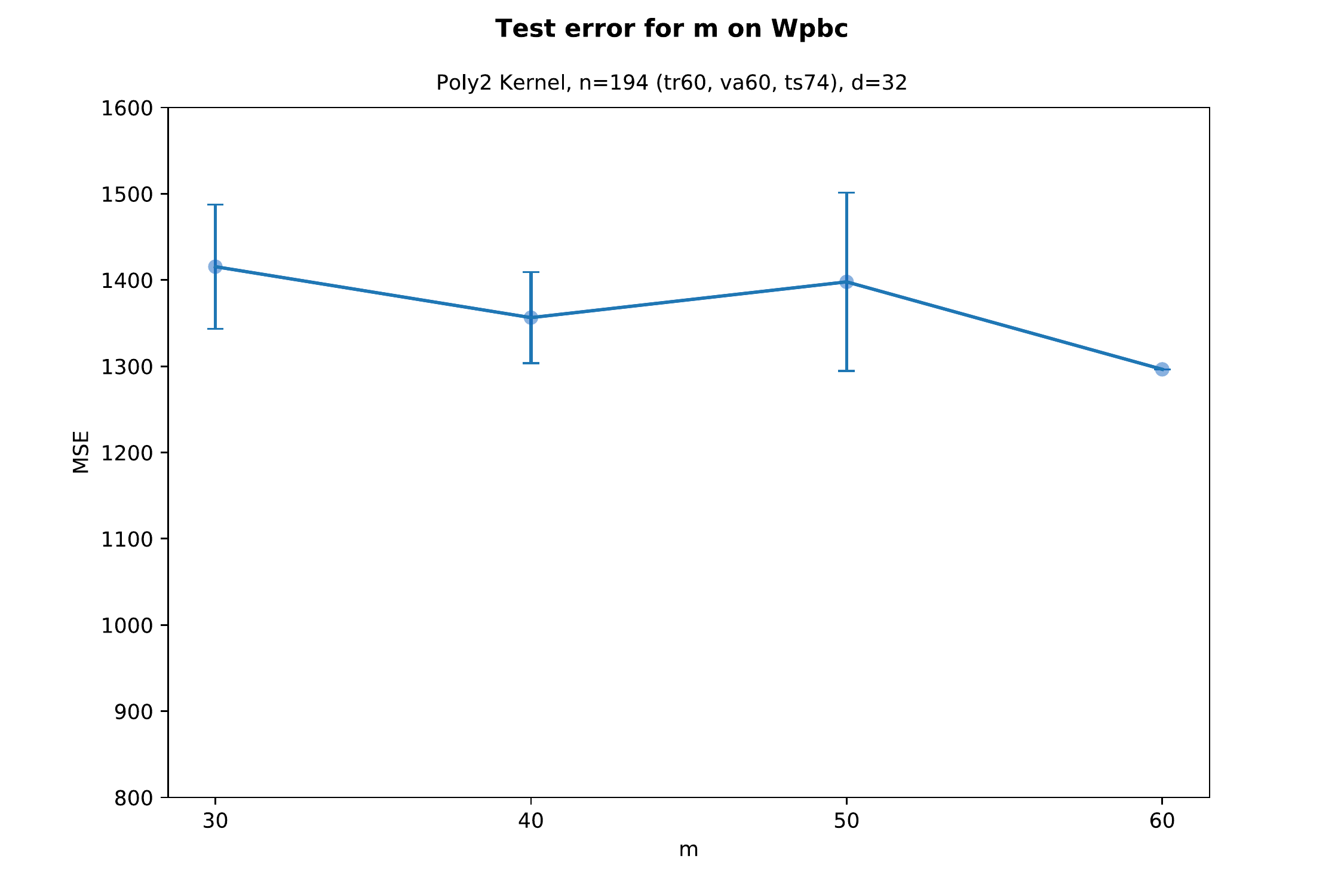}
\caption{Test error (MSE) evaluated on Wpbc dataset. Polynomial kernel of degree 2, $\gamma$ for each $m$ is selected through the validation set.}
\label{test_wpbc}
\end{figure}

Since we want to analyse the \nystrom approach, in \autoref{test_wpbc} we present the results on the test set for each value of $m$, where we chose the best corresponding $\gamma$. Again this is carried out on $10$ runs, where in each run we operate a new sampling of $m$ points and train the model with $\gamma$ already chosen. As you can see, this result is in line with the previous analysis, suggesting that employing \nystrom method is a viable choice. Note that in \autoref{test_wpbc} the run for $m=60$ has no variation since it includes the entire training set, thus the 10 executed runs all output the same value, hence $std=0$. As you can see there is a relatively small difference when we consider subsampling ($m$), and the loss with respect to the MSE obtained with $m=n$ is negligible.

\newpage
\subsection{Analysing feature selection capability}\label{nystrom_manual}
\begin{figure}[!ht]
\centering
\includegraphics[trim=1cm 0.5cm 1cm 1.5cm,clip,width=0.9\textwidth]{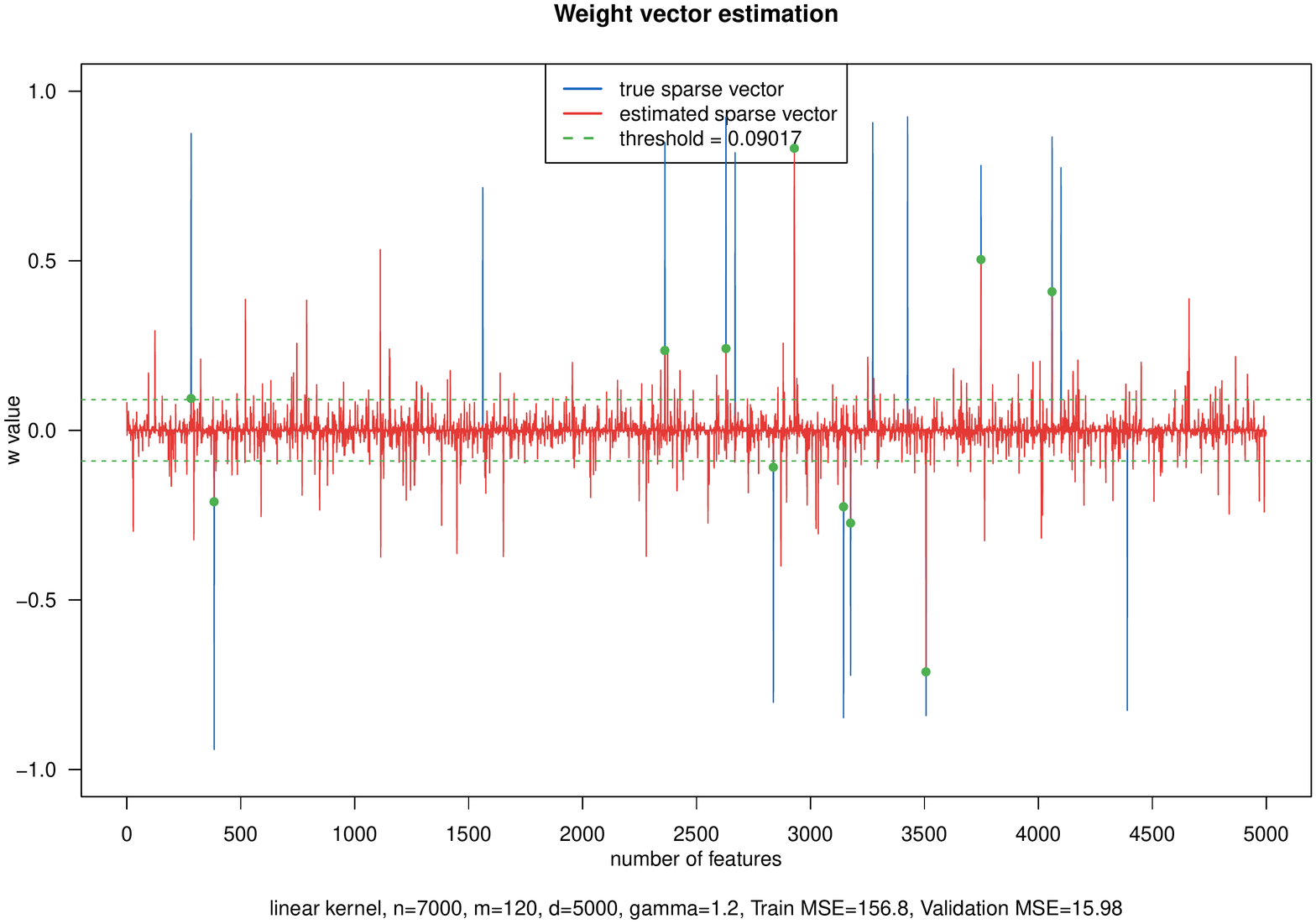}
\caption{Resulting weight plot for $m=120$. Features over threshold: $11 / 223$. Correctly estimated features over the threshold are pointed with a green point.}
\label{threshold_120}
\end{figure}

Considering again the synthetic datas as in the setting exposed in \autoref{nystrom_synthetic}, we would like to analyse the feature extraction capability of the algorithm in more detail. \bigbreak
In what follows we manually chose the $\gamma$ parameter which better emphasizes the behaviour of the extracted features weights for each value of $m$.
In \autoref{threshold_120} you can see the resulting weights vector $w$, after the model has been trained on a sample of $m=120$ points out of the $4000$ of the training set. Recall that we built the synthetic dataset with $s=17$ true relevant features. Recall also that we are not applying $L_1$ regularization but $L_{4/3}$ instead, which is an approximation of the sparsity providing method discussed in \autoref{l1_section}, thus we have to discard some of the noise created as a result of this. For that purpose we applied an adaptive thresholding on the weights simply determined as twice the value of the standard deviation $2*std$. The actual value of the applied threshold is written in the legend of following plots. In \autoref{threshold_120}, out of $d=5000$ features, $11$ of the relevant features have been captured crossing the threshold, together with $223$ non relevant ones.

\begin{figure}[!ht]
\centering
\includegraphics[trim=1cm 0.5cm 1cm 1.5cm,clip,width=0.9\textwidth]{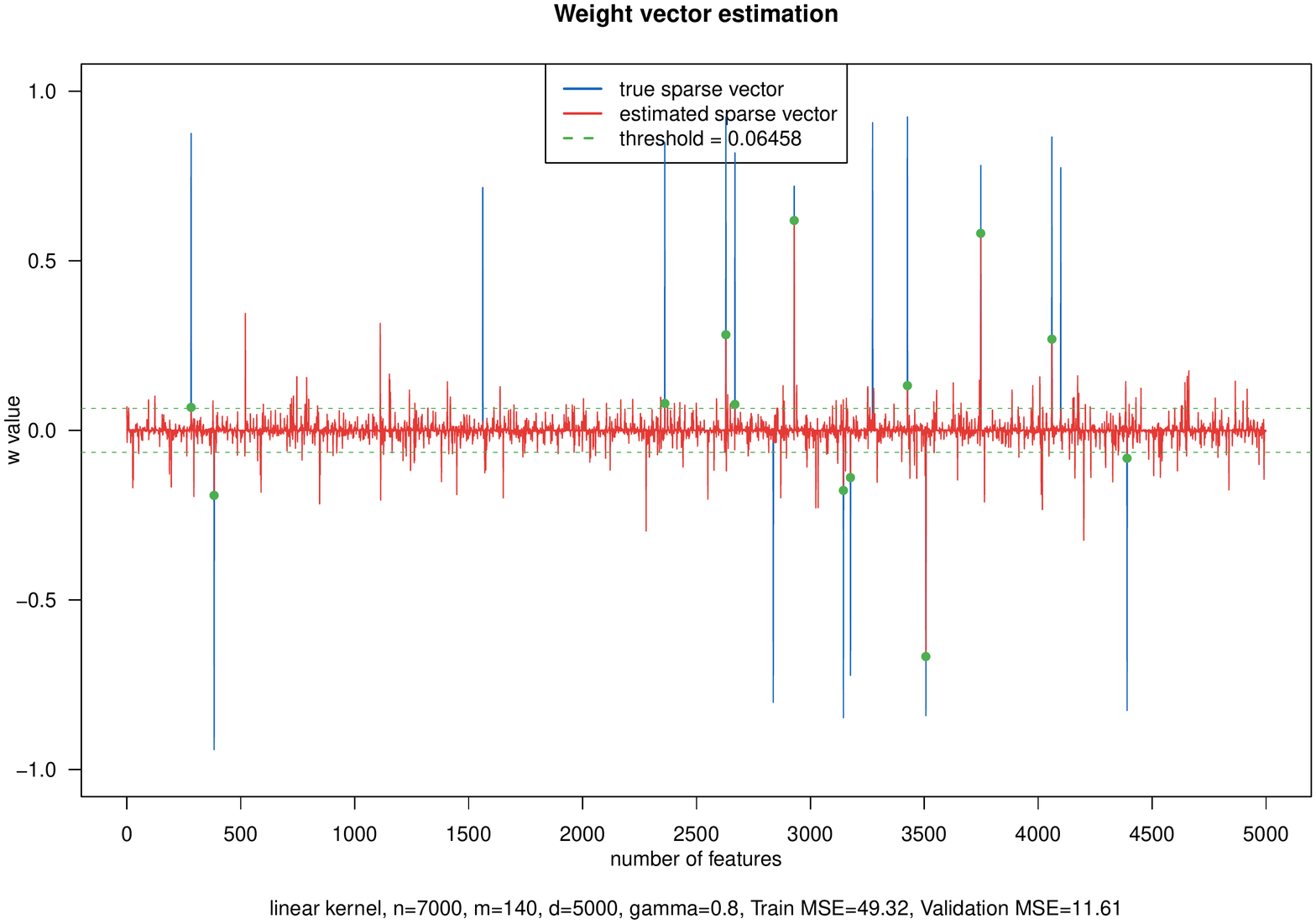}
\caption{Resulting weight plot for $m=140$. Features over threshold: 13 / 202. Correctly estimated features over the threshold are pointed with a green point.}
\label{threshold_140}
\end{figure}

On the next experiment instead we report the result for the run with $m=140$, where we picked $\gamma=0.8$ as the more visually informative case (see \autoref{threshold_140}). In this scenario we have $13$ correctly estimated features (out of the $s=17$) over the threshold, and $202$ non relevant ones. You can also notice how the noise (weights associated to all the other features) is somehow reduced and visually more shallow, while weights in correspondence to true relevant features are more accentuated.

\begin{figure}
\centering
\includegraphics[trim=1cm 0.5cm 1cm 1.5cm,clip,width=0.9\textwidth]{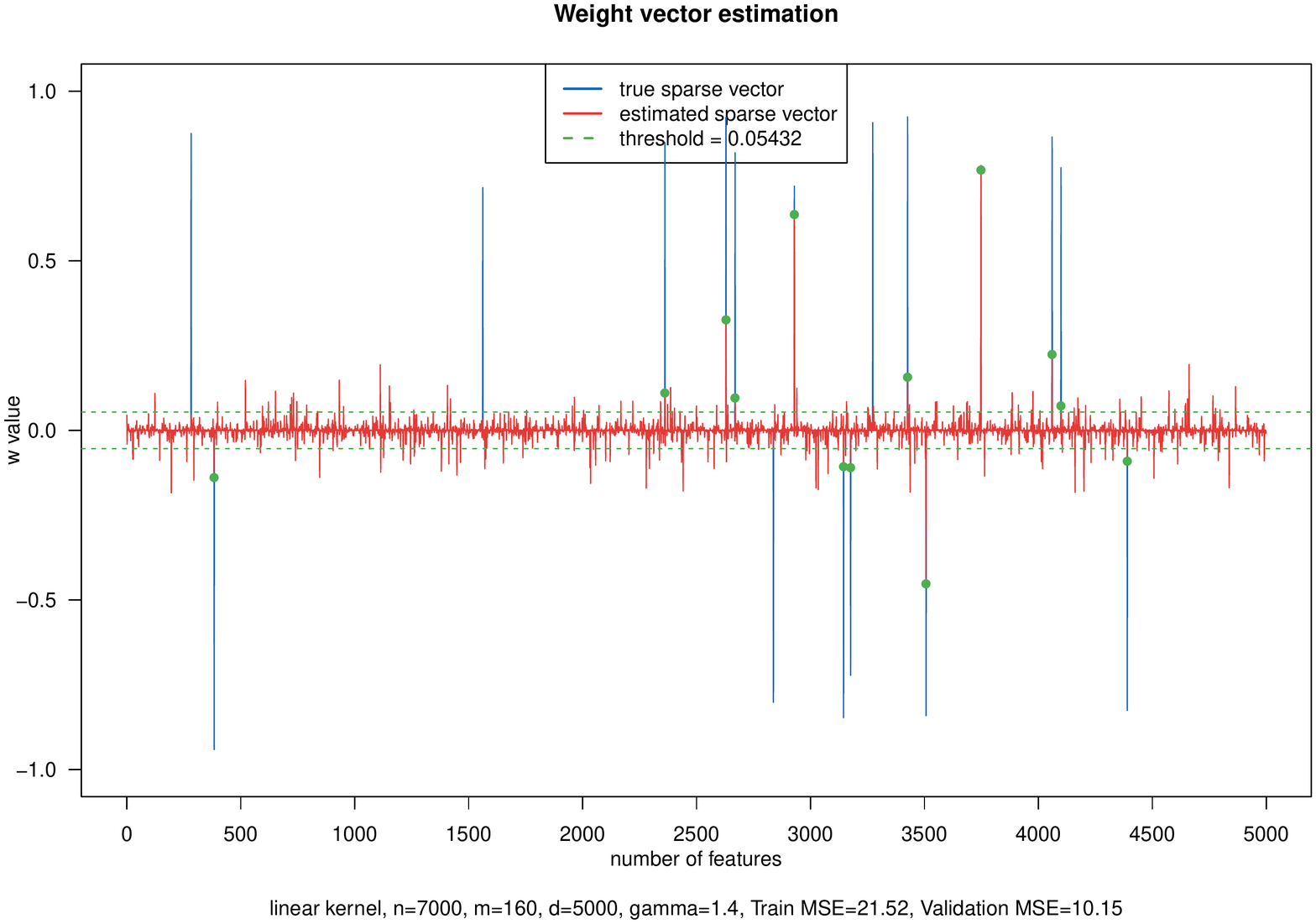}
\caption{Resulting weight plot for $m=160$. Features over threshold: 13 / 184. Correctly estimated features over the threshold are pointed with a green point.}
\label{threshold_160}
\end{figure}

\newpage
The last considered plot is the one showed in \autoref{threshold_160}. The behaviour is the same as the one described for $m=160$, with more shallow noise and more emphasis on relevant features (pointed out with a green point). In this case we still capture $13$ of the relevant features as before, but only $184$ irrelevant ones. Thus providing a small improvement from the previous scenario.

\newpage
\subsection{Comparison employing both improvements}
The last experiment has been carried out on  Dexter dataset (described in \autoref{sec_dexter}). We analysed the execution time by employing both the considered improvements, \nystrom subsampling approach and the proposed new layout (as described in \autoref{sec_five}). Subsampling has been operated on $n=200$ training set points. Datas are mean-std standardized. In this case $40$ optimization iterations have been considered.\bigbreak

As you can see from \autoref{nystrom_memory}, the left column (Time New) indicating execution times employing the proposed layout is constantly favorable for each choice of $m$. In particular, by picking $m=20$ subsampled points, we achieve a speedup of \colblue{24.2} by making use of the new layout instead of the previous one. This behaviour is incrementally suggesting the improvement over the old layout, indeed for successive values of $m$ we achieve better values of speedup. Ending up with an incredible speedup of \colblue{40.2} for $m=80$. Note how the speedup improves for each increment of $m$, giving empirical evidence that there is further improvement for higher values of $m$, suggesting once more that employing the proposed layout brings a considerable enhancement.
\label{nystrom_memory}
\begin{table}[!ht]
\centering
\begin{tabular}{|c|r|r|c|}
\hline
m & Time New & Time Old & Speedup \\ \hline
20 & 0.228 s. & 5.522 s. & 24.2		\\ \hline
30 & 1.052 s. & 29.383 s. & 27.9	\\ \hline
40 & 3.357 s. & 103.857 s. & 30.9	\\ \hline
50 & 8.273 s. & 259.327 s. & 31.3	\\ \hline
60 & 17.509 s. & 570.761 s. & 32.6	\\ \hline
70 & 33.236 s. & 1192.254 s. & 35.8	\\ \hline
80 & 58.801 s. & 2368.586 s. & 40.2	\\ \hline
\end{tabular}
\caption{Execution times comparing the old to the new proposed layout, evaluated on Dexter dataset, using \nystrom subsampling with $m$ points. For the optimization problem we considered 40 iterations, $\gamma=1.0$, features are mean-std standardized.}
\end{table}

\newpage
\nocite{*}
\addcontentsline{toc}{section}{References}
\bibliographystyle{te}
\bibliography{sec/biblio}

\begin{thebibliography}{17}
\newcommand{\enquote}[1]{``#1''}
\providecommand{\natexlab}[1]{#1}
\providecommand{\url}[1]{\texttt{#1}}
\providecommand{\urlprefix}{URL }
\providecommand{\bibAnnoteFile}[1]{%
  \IfFileExists{#1}{\begin{quotation}\noindent\textsc{Key:} #1\\
  \textsc{Annotation:}\ \input{#1}\end{quotation}}{}}
\providecommand{\bibAnnote}[2]{%
  \begin{quotation}\noindent\textsc{Key:} #1\\
  \textsc{Annotation:}\ #2\end{quotation}}

\bibitem[{Berlinet and Thomas-Agnan(2011)}]{berlinet2011}
Berlinet, Alain and Christine Thomas-Agnan (2011), \emph{Reproducing kernel
  Hilbert spaces in probability and statistics}. Springer Science \& Business
  Media.
\bibAnnoteFile{berlinet2011}

\bibitem[{Cover and Hart(1967)}]{cover1967}
Cover, Thomas and Peter Hart (1967), \enquote{Nearest neighbor pattern
  classification.} \emph{IEEE transactions on information theory}, 13, 21--27.
\bibAnnoteFile{cover1967}

\bibitem[{Guyon(2003)}]{guyon2003}
Guyon, Isabelle (2003), \enquote{Design of experiments of the nips 2003
  variable selection benchmark.} In \emph{NIPS 2003 workshop on feature
  extraction and feature selection}, volume 253.
\bibAnnoteFile{guyon2003}

\bibitem[{Hinton(1888)}]{hinton1888}
Hinton, Charles~Howard (1888), \emph{A new era of thought}. S. Sonnenschein \&
  Company.
\bibAnnoteFile{hinton1888}

\bibitem[{Kimeldorf and Wahba(1971)}]{kimeldorf1971}
Kimeldorf, George and Grace Wahba (1971), \enquote{Some results on
  tchebycheffian spline functions.} \emph{Journal of mathematical analysis and
  applications}, 33, 82--95.
\bibAnnoteFile{kimeldorf1971}

\bibitem[{Koltchinskii(2009)}]{koltchinskii2009}
Koltchinskii, Vladimir (2009), \enquote{Sparsity in penalized empirical risk
  minimization.} In \emph{Annales de l'IHP Probabilit{\'e}s et statistiques},
  volume~45, 7--57.
\bibAnnoteFile{koltchinskii2009}

\bibitem[{Rennie(2005)}]{rennie2005smooth}
Rennie, Jason~DM (2005), \enquote{Smooth hinge classification.}
\bibAnnoteFile{rennie2005smooth}

\bibitem[{Rudi et~al.(2015)Rudi, Camoriano, and Rosasco}]{rudi2015}
Rudi, Alessandro, Raffaello Camoriano, and Lorenzo Rosasco (2015),
  \enquote{Less is more: Nystr{\"o}m computational regularization.} In
  \emph{Advances in Neural Information Processing Systems}, 1657--1665.
\bibAnnoteFile{rudi2015}

\bibitem[{Salzo and Suykens(2016)}]{salzo2016}
Salzo, Saverio and Johan~AK Suykens (2016), \enquote{Generalized support vector
  regression: duality and tensor-kernel representation.} \emph{arXiv preprint
  arXiv:1603.05876}.
\bibAnnoteFile{salzo2016}

\bibitem[{Salzo et~al.(2017)Salzo, Suykens, and Rosasco}]{salzo2017}
Salzo, Saverio, Johan~AK Suykens, and Lorenzo Rosasco (2017), \enquote{Solving
  $\ell^{p}$-norm regularization with tensor kernels.} \emph{arXiv preprint
  arXiv:1707.05609}.
\bibAnnoteFile{salzo2017}

\bibitem[{Smola and Sch{\"o}lkopf(1998)}]{smola1998}
Smola, Alex~J and Bernhard Sch{\"o}lkopf (1998), \emph{Learning with kernels},
  volume~4. Citeseer.
\bibAnnoteFile{smola1998}

\bibitem[{Steinwart and Christmann(2008)}]{steinwart2008}
Steinwart, Ingo and Andreas Christmann (2008), \emph{Support vector machines}.
  Springer Science \& Business Media.
\bibAnnoteFile{steinwart2008}

\bibitem[{Stone(1977)}]{stone1977}
Stone, Charles~J. (1977), \enquote{Consistent nonparametric regression.}
  \emph{Ann. Statist.}, 5, 595--620,
  \urlprefix\url{https://doi.org/10.1214/aos/1176343886}.
\bibAnnoteFile{stone1977}

\bibitem[{Von~Luxburg and Sch{\"o}lkopf(2011)}]{luxburg2011}
Von~Luxburg, Ulrike and Bernhard Sch{\"o}lkopf (2011), \enquote{Statistical
  learning theory: Models, concepts, and results.} In \emph{Handbook of the
  History of Logic}, volume~10, 651--706, Elsevier.
\bibAnnoteFile{luxburg2011}

\bibitem[{Zhang et~al.(2009)Zhang, Xu, and Zhang}]{zhang2009}
Zhang, Haizhang, Yuesheng Xu, and Jun Zhang (2009), \enquote{Reproducing kernel
  banach spaces for machine learning.} \emph{Journal of Machine Learning
  Research}, 10, 2741--2775.
\bibAnnoteFile{zhang2009}

\bibitem[{Zhang(2004)}]{zhang2004solving}
Zhang, Tong (2004), \enquote{Solving large scale linear prediction problems
  using stochastic gradient descent algorithms.} In \emph{Proceedings of the
  twenty-first international conference on Machine learning}, 116.
\bibAnnoteFile{zhang2004solving}

\bibitem[{Zou and Hastie(2005)}]{elasticnet}
Zou, Hui and Trevor Hastie (2005), \enquote{Regularization and variable
  selection via the elastic net.} \emph{Journal of the Royal Statistical
  Society, Series B}, 67, 301--320.
\bibAnnoteFile{elasticnet}

\end{thebibliography}

\section*{}
\addcontentsline{toc}{section}{Acknowledgements}

\newenvironment{changemargin}[2]{%
  \begin{list}{}{%
    \setlength{\topsep}{0pt}%
    \setlength{\leftmargin}{#1}%
    \setlength{\rightmargin}{#2}%
    \setlength{\listparindent}{\parindent}%
    \setlength{\itemindent}{\parindent}%
    \setlength{\parsep}{\parskip}%
  }%
  \item[]}{\end{list}}
  
\begin{changemargin}{1cm}{1cm}

\begin{center}
\huge
\textbf{Acknowledgements}\bigbreak
\end{center}

I would like to thank my advisor of thesis, Saverio, who helped me a lot in each step of this thesis. I want to express my gratitute to prof. Pelillo who keeps providing me with amazing opportunities since my bachelor, I totally appreciate everything, thank you! In particular thanks for this collaboration with IIT of Genoa, thanks to Massimiliano. Thanks to the representative group for providing a friendly working environment (in particular thanks to prof. Raffaetà for being a lovely person).\medbreak
My gratitute goes also to my family, that provided me with the best conditions and environment that allowed me to study for all of this time. I would like to give a big thanks to my ethiopian buddies (manu) Leulee, Yoshua and Yonathan who introduced me to the humanities library in Venice and to late night studying and shared their hard working habits with me. 
Thanks for all the time spent together, thanks for the time in via roma apartment (thanks Carmelo), Hiwot Betam des Titalech (with arms wide open)! 
A warm thought goes to my library friends, Mattia (because every great friendship begins with a lighter), Giorgio (thanks for all the great music and great time), Marco (thanks for writing my name on public bathrooms around the world) and Giacomos (little Giacomo: thanks for the laughs, big Giacomo: thanks for the amazing discussions while having dinner). Thanks for the philosophy I didn't ask for, but greatly appreciated. 
Thanks for filling my head full of philosophical reasoning after long days of studying. 
Thanks for all the amazing time spent in Campo s. Margherita (probably this line includes all the people cited here). 
Thanks to Laura and Flavia for sharing an amazing late night library time, followed by a well-deserved cold beer. In particular to Flavia for the beer part, while Laura was a great canteen companion :). 
So thanks to the canteen for all the delicious food and the amazingly funny time spent on those tables. 
Thanks for laughing tears while eating, thanks to all the people who ate with me. 
Thanks to Sebastiano that introduced me to the fantastic world of research. Thanks for spending the christmas break for working on that paper. 
I feel honored to be the first student having you in the graduation commission.
Thanks to all the PhDs of the group, in particular to Ismail for all the time spent having fun and sharing your deep learning wisdom. Thanks to Federico for graduating together with me, for the second time, in particular thanks for last minute commentary about last minute changes (just as this line :).
Thanks to Shadow and Greg who spent the first summer in that library with me, working on projects and creating socks coloured heatmaps. \medbreak
Thanks to the stunning city of Venice for providing thoughtful walks and inspiring views at every corner. It was a complete honor to be part of Ca' Foscari, to live in Venice and meet all these fantastic people inside and outside of university. Thanks to all the hundreds and thousands of people from all around the world visiting Venice who stopped by to share some happiness with me. Thanks for teaching me how to say:\bigbreak

\includegraphics[trim=17cm 22.3cm 21cm 7cm,clip,width=0.2\textwidth]{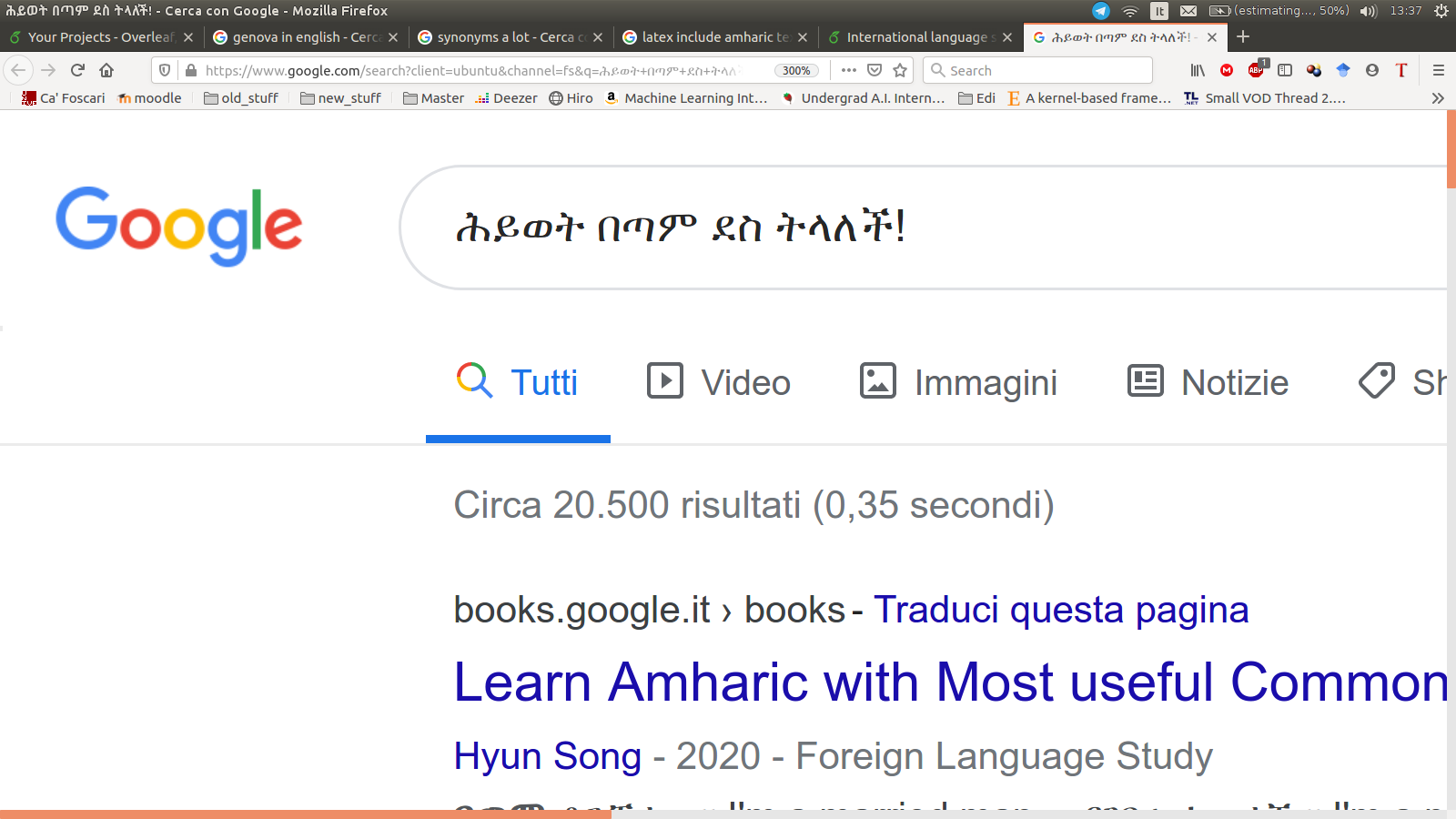}
 - Shenghuo feichang piaoliang - Jizn prekrasnaya - El\"am\"a on todella kaunista - $\check{Z}$ivot je lijep - Az élet nagyon szép - A vide é muito bonita - Jeeveethey hongak lassamai - Livet \"ar sa snygg - Dzive ir loti skaista - Zendegi zibast\\
Jeta \"esht\"e shum\"e e bukur!

\vspace{1cm}
\end{changemargin}
\newpage

\end{document}